\documentclass{article} 
\usepackage[final]{colm2026_conference}

\usepackage{microtype}
\usepackage{hyperref}
\usepackage{url}
\usepackage{booktabs}

\usepackage{graphicx}
\usepackage{enumitem}
\usepackage{tcolorbox}
\usepackage{amsmath}
\usepackage{subcaption}
\usepackage{multirow}
\usepackage{pifont}
\usepackage{minitoc}
\usepackage[toc,page,header]{appendix}
\usepackage{arydshln}
\usepackage{adjustbox}
\usepackage{algorithm}
\usepackage{algpseudocode}

\usepackage{lineno}

\newcommand\model[1]{{\footnotesize \texttt{#1}}}

\newcommand\deepseekLlama{\model{DeepSeek-R1-Distill-Llama-8B}}
\newcommand\deepseekQwen{\model{DeepSeek-R1-Distill-Qwen-14B}}
\newcommand\QwQLarge{\model{QwQ-32B}}
\newcommand\GPTmini{\model{GPT-4o-mini}}

\newcommand{\Puser}{{\footnotesize $\mathcal{P^{\text{(0)}}_{\text{user}}}$}}

\newcommand{\Psystemzero}{{\footnotesize $\mathcal{P^{\text{(0)}}_{\text{system}}}$}}

\newcommand{\Psystemone}{{\footnotesize $\mathcal{P^{\text{(1)}}_{\text{system}}}$}}

\newcommand{\Psystemthree}{{\footnotesize $\mathcal{P^{\text{(3)}}_{\text{system}}}$}}

\newcommand{\cmark}{\ding{51}} 

\definecolor{darkblue}{rgb}{0, 0, 0.5}
\hypersetup{colorlinks=true, citecolor=darkblue, linkcolor=darkblue, urlcolor=darkblue}

\title{TRACES: Tagging Reasoning Steps for Adaptive Cost-Efficient Early-Stopping}

\author{Yannis Belkhiter\\
IBM Research Europe\\
Trinity College Dublin\\
\texttt{yannis.belkhiter@ibm.com}\\
\And
Seshu Tirupathi\\
IBM Research Europe\\
Dublin, Ireland\\
\texttt{seshutir@ie.ibm.com}\\
\And
Giulio Zizzo\\
IBM Research Europe\\
Dublin, Ireland\\
\texttt{giulio.zizzo2@ibm.com}\\
\AND
\\
\\
\\
\And
John D. Kelleher\\
ADAPT Research Centre\\
Trinity College Dublin\\
\texttt{john.kelleher@tcd.ie}\\
}

\begin{document}

\ifcolmsubmission
\linenumbers
\fi

\newcommand{\ybnote}[1]{\ifshownotes{\color{blue}\noindent #1}}

\maketitle

\begin{abstract}
The field of Language Reasoning Models (LRMs) has advanced rapidly, enabling longer, and more accurate reasoning. However, a growing body of studies show that LRMs are still inefficient, over-generating verification and reflection steps. Additionally, the high-level role of each reasoning step and how different step types contribute to the generation of correct answers, is largely underexplored. To address this challenge, we introduce TRACES (\textbf{T}agging of the \textbf{R}easoning steps enabling \textbf{A}daptive \textbf{C}ost-\textbf{E}fficient early-\textbf{S}topping), a lightweight framework that tags reasoning steps in real-time, and enable adaptive, cost‑efficient early stopping of large‑language‑model inferences. 
By monitoring reasoning behaviors during inferences, we find that LRMs tend to shift their reasoning behavior after reaching a correct answer. 
We demonstrate that the monitoring of the specific type of steps can produce effective interpretable early stopping criteria. 
Evaluated on three mathematical reasoning benchmarks (MATH500, GSM8K, AIME) and two knowledge and reasoning benchmarks (MMLU and GPQA), TRACES achieve 20-50\% token reduction while maintaining comparable accuracy to standard generation. On harder tasks (BeyondAIME, IMO AnswerBench), the accuracy-efficiency trade-off is steeper, requiring a more conservative early-stopping threshold.
This work offers a novel way to study and control the generation's behaviors of LRMs.
\end{abstract}

\section{Introduction}

\begin{figure}[t]
    \centering
    \includegraphics[width=0.85\linewidth]{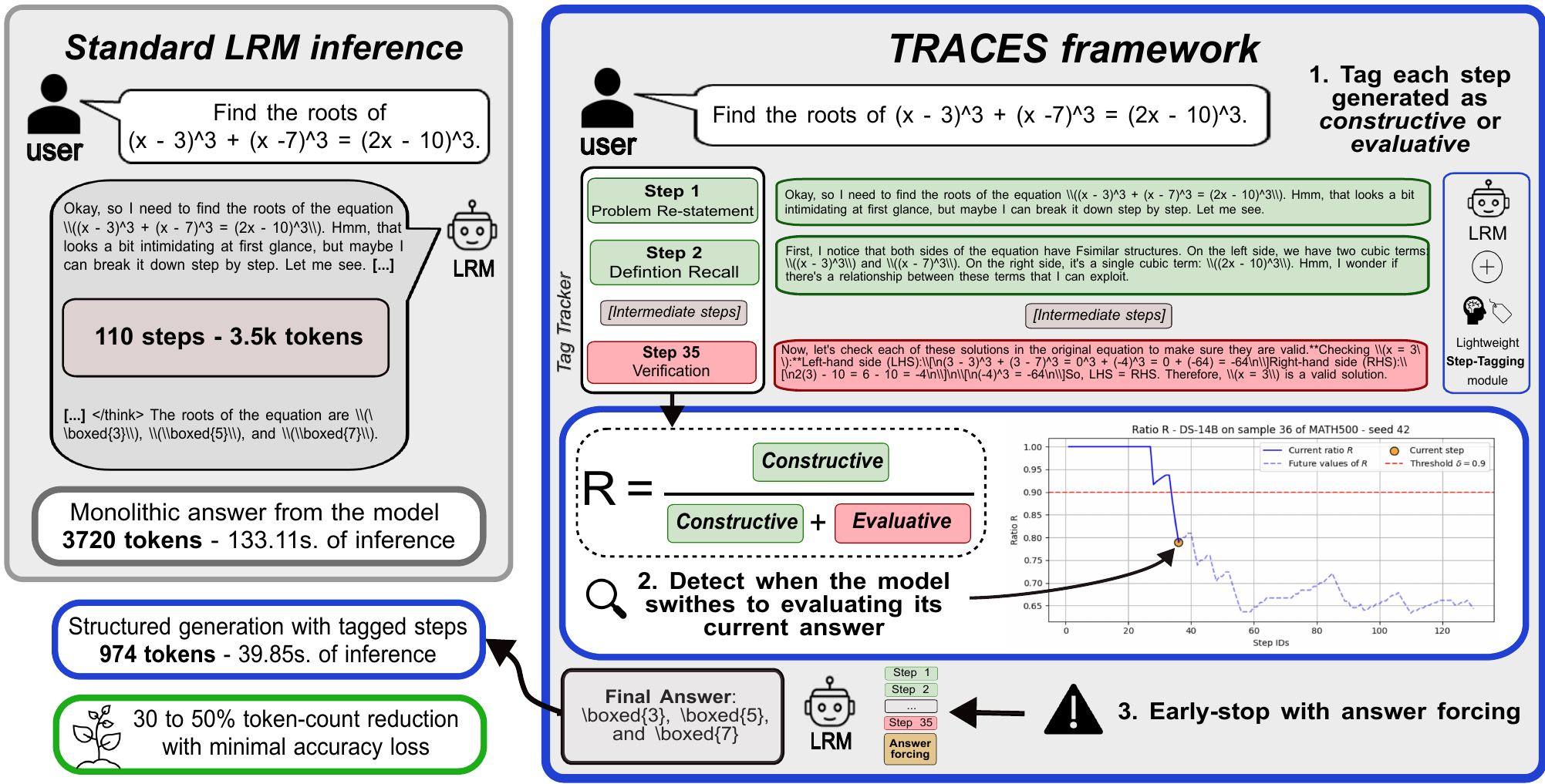}
    \vspace{-0.35cm}
    \caption{\small TRACES: a framework for monitoring and early-stopping the generation of LRMs - example on sample 36 from MATH500 test with \deepseekQwen{} - seed $42$}
    \label{fig:step-tagging}
    \vspace{-0.45cm}
\end{figure}

For the past few years, the field of Language Reasoning Models (LRMs) has experienced significant growth in terms of capabilities. Initiated by the work on model prompting such as Chain-of-Thought \citep{wei2023chainofthoughtpromptingelicitsreasoning} and Self-Consistency \citep{wang2023selfconsistencyimproveschainthought}, Inference Time Scaling has emerged as a popular field with the goal of making models more accurate at reasoning. At the same time, fundamental work on 
Training Time Scaling has led to the release of strong \emph{System-2} reasoning models.\\
\phantom{m}
However, recent studies have shown that LRMs need to generate a very large number of tokens\,---\,several thousands\,---\,in order to generate accurate answers for challenging questions \citep{qu2025surveyefficientreasoninglarge, chen2025think23overthinkingo1like, sui2025stopoverthinkingsurveyefficient}. This behavior makes reasoning models extremely inefficient - scaling in both compute resources and inference time. 
Although recent works have suggested solutions to this problem, the existing literature lacks work understanding the role of reasoning steps towards generation of correct answers \citep{yu2026explainablechainofthoughtreasoningempirical}. Furthermore, efficient inference-time scaling methods are either static - fixed budgets/prompt compression, lacking of interpretability - or dynamic, but they do not exploit the semantic content of reasoning steps to guide early stopping. As a result, most existing approaches to reduce the verbosity of LRMs overlook the possibility of monitoring the reasoning as it is being generated, and using this information to improve LRMs efficiency. 
\phantom{m}
To address this challenge, this paper aims to offer a new perspective on the efficiency of LRMs by focusing on online monitoring of models. Our contributions are as follows:

\begin{itemize}[leftmargin=*, itemsep=0pt, topsep=0pt]
    \item \textbf{Step-Tagging module:} We introduce the \emph{Step-Tagging} module (see Figure \ref{fig:step-tagging}), an online lightweight classifier capable of identifying the nature of each step that LRMs are generating. To encompass the various possible reasoning behaviors, we propose \emph{ReasonType}, a fine-grained taxonomy of reasoning steps enabling structured identification of steps. 

    \item \textbf{TRACES, an interpretable early-stopping framework:} 
    We found that LRM's step-types shift after generating a correct answer, moving from the \emph{construction} of the answer to its \emph{evaluation}. Based on these observations, we built a black-box early-stopping framework (TRACES) that dynamically stops token generation when such transition phase is detected. Tested on 6 open-source LRMs across 7 reasoning datasets, 
    TRACES reduced token-count by 20-50\% while maintaining accuracy on standard benchmarks.
\end{itemize}
%

\section{Related Work} \label{sec:related-work}

To render models less verbose and more efficient, Train and Test Time Scaling approaches have been explored \citep{qu2025surveyefficientreasoninglarge,li202512surveyreasoning,chen2025reasoningerasurveylong} and recent work has explored monitoring the generation of LRMs \citep{lee2025evaluatingstepbystepreasoningtraces,luo2025deconstructinglongchainofthoughtstructured}. 

\noindent \textbf{Efficient Reasoning through Training.} Using SFT approaches, work such as \citet{xia2025tokenskipcontrollablechainofthoughtcompression} explored fine-tuning models on compressed reasoning traces to limit the verbosity of LRM generation. Other papers have suggested various RL algorithms designed to make models more efficient. For instance, \citet{luo2025o1prunerlengthharmonizingfinetuningo1like, kimiteam2025kimik15scalingreinforcement, yu2025dapoopensourcellmreinforcement} showed that including a length component in the reward function leads to more efficient inference. 

\noindent \textbf{Efficient Reasoning during Inference.} Researchers have also explored Inference Time Scaling technique to increase the efficiency of models \citep{qu2025surveyefficientreasoninglarge}. \emph{Model Switch} uses a router module to select small or large models for inference depending on the complexity of the problem \citep{ong2025routellmlearningroutellms}. Similarly, \emph{System Switch} looked at dynamically selecting inference settings based on the problem \citep{aytes2025sketchofthoughtefficientllmreasoning}. \emph{Length Budgeting} aims to reduce the budget allocated to the generation of answers. Works such as \citet{lee2025llmscompresschainofthoughttoken, han2025tokenbudgetawarellmreasoning, xu2025chaindraftthinkingfaster} showed that careful prompt engineering can lead to more efficient generation compared to standard inference. In addition, \citet{pu2025thoughtterminatorbenchmarkingcalibratingmitigating} demonstrated that calibration experiments can be performed to estimate the optimal number of tokens to solve particular problems. However, these techniques often lack \textbf{interpretability} since they rely on either prompting or token-count estimation. Furthermore, these techniques define constraints \textbf{before inference} and so remain \textbf{static} and \textbf{do not adapt to the evolving reasoning traces} of the model. As a result, they cannot exploit the information contained in the model's own generated reasoning during inference.

\noindent \textbf{Monitoring LRM generation.} We observe an emerging theme of research on monitoring LRM generation at a step level. Specifically, \citet{lee2025evaluatingstepbystepreasoningtraces} proposes a taxonomy of reasoning traces evaluators. However, the authors acknowledged that existing monitoring approaches are not adapted to complex reasoning traces. Similarly, \citet{golovneva2023roscoesuitemetricsscoring} prompts a model to generate step-by-step reasoning, and defines an error-type taxonomy to evaluate reasoning steps. Nevertheless, this method is post-hoc and focuses on measuring the quality of reasoning. 
Furthermore, \citet{luo2025deconstructinglongchainofthoughtstructured} decomposes long reasoning traces by prompting an LLM to parse reasoning steps and assigns to each step one high-level class - from a taxonomy of four step-types. However, this approach is applied for training purposes and does not monitor reasoning during inference. Closer to our work, LCoT2Tree converts long CoT into hierarchical tree structures to analyse the reasoning patterns of LRMs \citep{jiang2025makesgoodreasoningchain}. The framework classifies thoughts into types of reasoning, but their framework is applied after the generation, and is not applied dynamically. Taking a white-box approach, \citet{venhoff2025basemodelsknowreason} derive a taxonomy of reasoning behaviors through trained Sparse Auto-encoders on step activations, clustering them to identify common behaviors. While their approach results in a rich taxonomy, their technique requires full-access to the model's internals. \citet{yu2026explainablechainofthoughtreasoningempirical} also note that \emph{``the high-level semantic roles of reasoning steps and their transitions remain underexplored."}. As a result, existing works often overlook the question of how to \textbf{dynamically monitor} LRMs reasoning during \textbf{single inferences} from \textbf{text alone}. Furthermore, existing \textbf{monitoring methods are not applied for early-stopping purposes}.


\textbf{Monitoring for dynamic efficient inference.} While existing early-stopping methods above are defined before inference, recent work has also explored early-stopping dynamically. \citet{yang2025dynamicearlyexitreasoning} proposes dynamic early stopping criteria based on the model's confidence (DEER). Similarly, \citet{wang2025entropylangletextttthinkrangle} suggests an early-stopping criteria based on the entropy of the tokens generated after forcing the generation of end-of-thinking tokens (EAT). However, both DEER and EAT rely on model's internals, require \textbf{white-box} LRMs for analysis by nature, and does not \textbf{exploit the the text content} of the reasoning trace itself.

\textbf{Contributions.} Our TRACES framework addresses the following gaps:

\begin{itemize}[leftmargin=*, itemsep=0pt, topsep=0pt]

    \item \textbf{Online, interpretable step classification.} Existing efficient inference-time methods treat LRMs as a black-box and apply fixed or dynamic constraints based on signals that carry no interpretation on what the model is doing at each step. We address this by proposing an online tagging framework that classifies the nature of each steps that is generated. 

    \item \textbf{Black-box inference monitoring.} Existing step-level monitoring methods are either applied after generation, or requires access to the model's internals. We address this with an online black-box monitoring module that requires only text generated of LRMs.

    \item \textbf{Content-aware dynamic early-stopping.} No prior work bridges online step-type monitoring to dynamic early-stopping. We address this by building a novel dynamic early-stopping controller driven by the live sequence of classified step types.

\end{itemize}


\section{Monitoring the reasoning process of LRMs} \label{sec:monitoring-lrms}

The concept of a reasoning step is central in evaluating and improving the generation of LRMs. However, defining a reasoning step is a non-trivial problem. As highlighted by \citet{yao2023treethoughtsdeliberateproblem, lee2025evaluatingstepbystepreasoningtraces, cao2025stepguidedreasoningimproving}, various definitions exists depending on the models, the problem, and the research goals. In this section, we survey existing approaches and select the one most commonly adopted in the literature.

%
%

\textbf{Step-by-step generation.} Rather than viewing the model's output as a monolithic text sequence, recent work has shifted towards decomposing generation of LRMs into discrete steps (see Appendix \ref{app:step-def}). This decomposition enables finer-grained analysis of model behavior and facilitates targeted interventions. Building upon the auto-regressive generation of LLMs \citep{schuurmans2024autoregressivelargelanguagemodels}, we first formalize the notion of \emph{stepwise generation}. 

Let $y = y_{1:n} \in V^n$ be the output token sequence generated by the model over the vocabulary $V$. We define a reasoning delimiter token $\alpha \in V$.  Let $R = \{r_0 = 1, \dots, r_i, \dots, r_{T} = n \}$ denote the indices in $y$ corresponding to the occurrence of $\alpha$ in $y$. $r_0$ and $r_{T}$ correspond to the first and last indexes of $y_{1:n}$. Based on these indices, we define a sequence - of length $T$ - of reasoning steps formed by $y_{1:n}$ with the delimiter $\alpha$:
\begin{equation}\label{eq:step-equation}
S = \{s_1, \ldots, s_i, \ldots, s_{T}\}, \text{ such as } s_i = y_{r_{i-1}:r_{i}}
\end{equation}
where each step $s_i$ corresponds to a sub-part of the full output $y$. From the literature, in Appendix \ref{app:step-def}, multiple delimiters have been defined. In this paper, we segmented the reasoning traces based on the delimiter $\alpha = \texttt{"\textbackslash n\textbackslash n"}$ \citep{zhang2025reasoningmodelsknowtheyre}. Building on this segmentation, we introduce \emph{Step-Tagging}: a lightweight module capable of identifying, and tagging reasoning steps in real-time during inference.


\textbf{Step-Tagging module.} Our definition of a reasoning step enables users to segment reasoning steps within model outputs. However, this definition alone does not allow the user to annotate the segmented steps with reasoning types. Such annotation would enable users to track logical transitions within model outputs. To do this, we must first define a tag dictionary $\mathcal{T}_{\text{tags}}$ (i.e., a label space of reasoning step tags) that covers the types of reasoning steps generated by models. Given a sequence of reasoning steps \( S = \{s_1, s_2, \ldots, s_T\} \), we wish to label each step \( s_i \) with a tag \( \tau_i \in \mathcal{\mathcal{T}_{\text{tags}}} \). Formally, we are looking to construct a step-tagging function $\phi$ such as:
\begin{equation}\label{eq:step_tagging}
    \forall \hspace{0.1cm} i \in [1,T], \phi(s_i) = \tau_i
\end{equation}
where $s_i \in S$ is a reasoning step from the full output sequence $y$, $\phi$ is the step-tagging function, and $\tau_i \in \mathcal{T}_{tags}$ is the reasoning tag associated to the step $s_i$.

\textbf{Taxonomy of the type of steps.} To enable fine-grained monitoring of reasoning behavior, we need to know the different types of reasoning steps that are typically generated by LRMs (i.e., we need to define $\mathcal{T}_{\text{tags}}$). To do so, we created a taxonomy based on the outputs of both \deepseekLlama{} \citep{deepseekai2025deepseekr1incentivizingreasoningcapability} and \QwQLarge{} \citep{qwq32b} models.

\begin{figure}[h]
    \centering
    \includegraphics[width=0.7\linewidth]{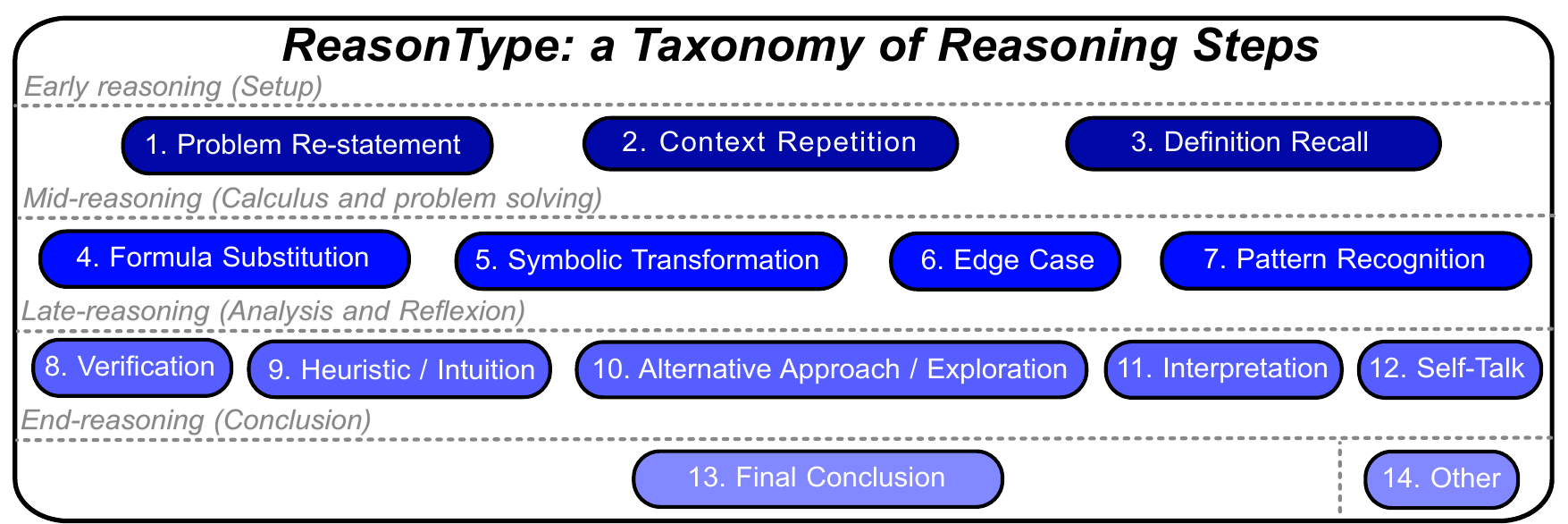}
    \caption{ReasonType - A taxonomy of reasoning step types as per \texttt{gpt-4o-mini}}
    \label{fig:taxonomy-step-type}
\end{figure}

Inspired by prior work on model behavior analysis \citep{galichin2025icoveredbaseshere, kuznetsov2025featurelevelinsightsartificialtext}, we first created a prompt to identify distinct types of reasoning steps in the traces. We then sampled $100$ reasoning traces from the MATH500 train dataset (covering 20 samples per difficulty level for each model) and using our prompt submitted the traces to \GPTmini{} \citep{openai2024gpt4technicalreport}. The prompt resulted in a series of different step-types. We merged overlapping categories, to construct a taxonomy that reflects the temporal and reasoning progression of model's traces. We refer to this taxonomy as \emph{ReasonType} (Figure \ref{fig:taxonomy-step-type}) encompassing $13$ categories, including early-stage behaviors such as \emph{Definition Recall}, and later reasoning stages like \emph{Verification} and \emph{Exploration}. Our process is detailed in Appendix \ref{app:construction-taxonomy}. To validate our taxonomy, we conducted a series of ablation studies in Appendices \ref{sec:appendix-reliability-annotation}, \ref{sec:appendix-validation-taxonomy-shuffled}, and \ref{sec:appendix-generalization-taxonomy}, assessing the reliability, robustness, and generalization of our taxonomy, respectively. 

\section{Experimental Setting} \label{sec:experimental-section}

Our paper contains two objectives. First, our goal is to prove that LRMs change their reasoning behavior after knowing the answer for the first time (\textbf{Claim 1}). Furthermore, we show that the we can monitor reasoning phase transition robustly using lightweight classifiers, and that such monitoring can be used to make the generation of LRMs more efficient (\textbf{Claim 2}). In this section, we will first motivate our selection of models, datasets, metrics, and inference settings followed by our baseline implementation.

\textbf{Model selection.} To apply our framework a user must have access to the fine-grained reasoning traces of LRMs. However, many high-performing closed-source models (such as, o3 and Claude 3.7) do not expose raw reasoning traces. Instead, these models output summaries of thinking tokens generated, which can bias the estimation of their efficiency compared to open-source models. In contrast, open-source models like DeepSeek-R1 and QwQ consistently provide reasoning traces. For this reason, we focus our analysis exclusively on \deepseekLlama{}, \deepseekQwen{} and \QwQLarge{}, which offer the granularity needed to monitor the reasoning process. This choice is motivated by their variety in term of size, full open-source availability, and diversity in providers. 

\textbf{Datasets.} To assess our approach, we selected five state-of-the-art reasoning datasets. Table \ref{tab:reasoning-datasets} in Appendix \ref{app:appendix-metrics} presents the datasets that we selected. First, we evaluated our models on mathematical tasks. We selected 3 datasets, with growing level of difficulty. Indeed, \emph{GSM8K}, \emph{MATH500}, and \emph{AIME} are issued from high-school, mathematic competition, and Olympiads, respectively. More complex tasks requires the model to generate more tokens, allowing us to test our framework upon different circumstances. Second, we evaluated the applicability of our framework on different domains. \emph{GPQA-Diamond} evaluates the model reasoning ability on Physics or Biology PhD-level questions, while \emph{MMLU-Pro} evaluate larger range of domains. We evaluated the correctness of the answers using Math-Verify (mathematic tasks) or MCQ-Prompting (knowledge and reasoning benchmarks). Appendix \ref{app:appendix-metrics} details our evaluation set-up. Our analysis in Section \ref{sec:influence-step-type} and the training of our step-tagger is conducted on the \emph{training splits}. We then evaluated TRACES in Section \ref{sec:st-es-evaluation} on the held-out \emph{testing splits}.







\noindent \textbf{Inference setting.} To monitor the steps and intervene in the generation process, we suggest a new definition of the generation process of LRMs. We assume that each model generates one token at a time, and we split the steps dynamically. However, for the purposes of our experiments instead of re-designing the generation process, we performed standard inference and applied our \emph{Step-Tagging} and \emph{TRACES} algorithms \emph{offline}. To ensure the robustness and reproducibility of our approach, we generated five outputs per test sample using fixed random seeds (namely $40$, $41$, $42$, $43$, and $44$), with deterministic decoding. \\
\phantom{m}
The five-seed evaluation was conducted on GSM8K and MATH500 across all three model sizes (8B, 14B, 32B). Due to computational constraints, for AIME, GPQA and MMLU, we report results on a single seed (42) using the 14B model. The low variance observed across seeds on GSM8K and MATH500 (see Table \ref{tab:variance_results} in Appendix \ref{sec:appendix-st-es-performance-math500-gsm8k}) justifies the usage of a single seed, and is consistent with common practices \citep{deepseekai2025deepseekr1incentivizingreasoningcapability}. 

\noindent \textbf{Baselines.} To assess the effectiveness of our approach, we define two baselines:

\begin{itemize}[leftmargin=*, itemsep=0pt, topsep=0pt]
    \item \textbf{Ideal Early stopping - $\mathcal{IES}$:} We observe a growing understanding that, up to a token-budget, thinking longer may be leading to worse results \citep{hassid2026dontoverthinkitpreferring,muennighoff2025s1simpletesttimescaling}. \citet{yang2025dynamicearlyexitreasoning} observes that certain models achieved correct answers very early in the generation, and defines the \emph{Pearl Reasoning} - i.e. the minimal number of steps needed by a model to get a correct answer. Inspired by this work, we define the \emph{Ideal Early Stopping}, which prompts the model after every step generated to retrieve its current best answer (i.e. answer forcing). By doing so, we identify the first step where the model reaches the correct answer - if any (see Algorithm \ref{alg:ies_algorithm} in Appendix \ref{sec:appendix-ideal-early-stopping} for implementation). 

    \item \textbf{Prompt-guided efficiency - $\mathcal{P_{\text{guided}}}$:} We also observe that LRMs are sensitive to the input prompt \citep{lee2025llmscompresschainofthoughttoken}. In this case, we compare our framework with user-prompt and system-prompt variants, with Zero-Shot and Few-Shot prompts that aim to reduce the reasoning computation while retaining accuracy. We explicitly instruct the models to not generate verbose output, or over-verification steps. We select 4 variants, namely: zero-shot user and system prompt, and few-shot system prompt with $1$ and $3$ examples: \Puser{}, \Psystemzero{}, \Psystemone{}, \Psystemthree{}, respectively (see Figure \ref{fig:prompt_baselines} in Appendix \ref{sec:appendix-prompt-guided-baselines}). 

\end{itemize}

Our Step-Tagging method enables us to track the semantic progression of the reasoning process. By definition, our $\mathcal{IES}$ baseline allows us to identify the first step in the reasoning trace at which the model produces the correct answer. This provides a reference point for analyzing how the reasoning evolves with respect to answer correctness. In the following section, we observe that LRMs generate different type of reasoning steps before and after reaching the correct answer in their reasoning trace. 

\section{Influence of the step-type on the model's efficiency} \label{sec:influence-step-type}

\textbf{Ideal Early-Stopping.} Let $S_{IES} \in S = \{s_1, \dots, s_T\}$ be the first step where the model expressed the correct answer. If the trace is incorrect, $S_{IES} = s_T$. To begin, we computed the $S_{IES}$ on the training datasets. To better observe reasoning patterns leading to correct answers, the experiments in this section are pruned on \emph{correct samples}. Figure \ref{fig:IES-train} in Appendix \ref{sec:appendix-ideal-early-stopping} presents the position of $S_{IES}$ relatively to the number of steps. The $\mathcal{IES}$ algorithm show the inefficiency of the models on reasoning datasets, with answer reached within 33\% of the full reasoning steps on average. It confirms that frequently, the model knows the correct answer very early in its reasoning trace, but still continues to generate (see \citet{yang2025dynamicearlyexitreasoning}).

\begin{figure}[h]
    \centering
    \vspace{-0.2cm}
    \includegraphics[width=0.9\linewidth]{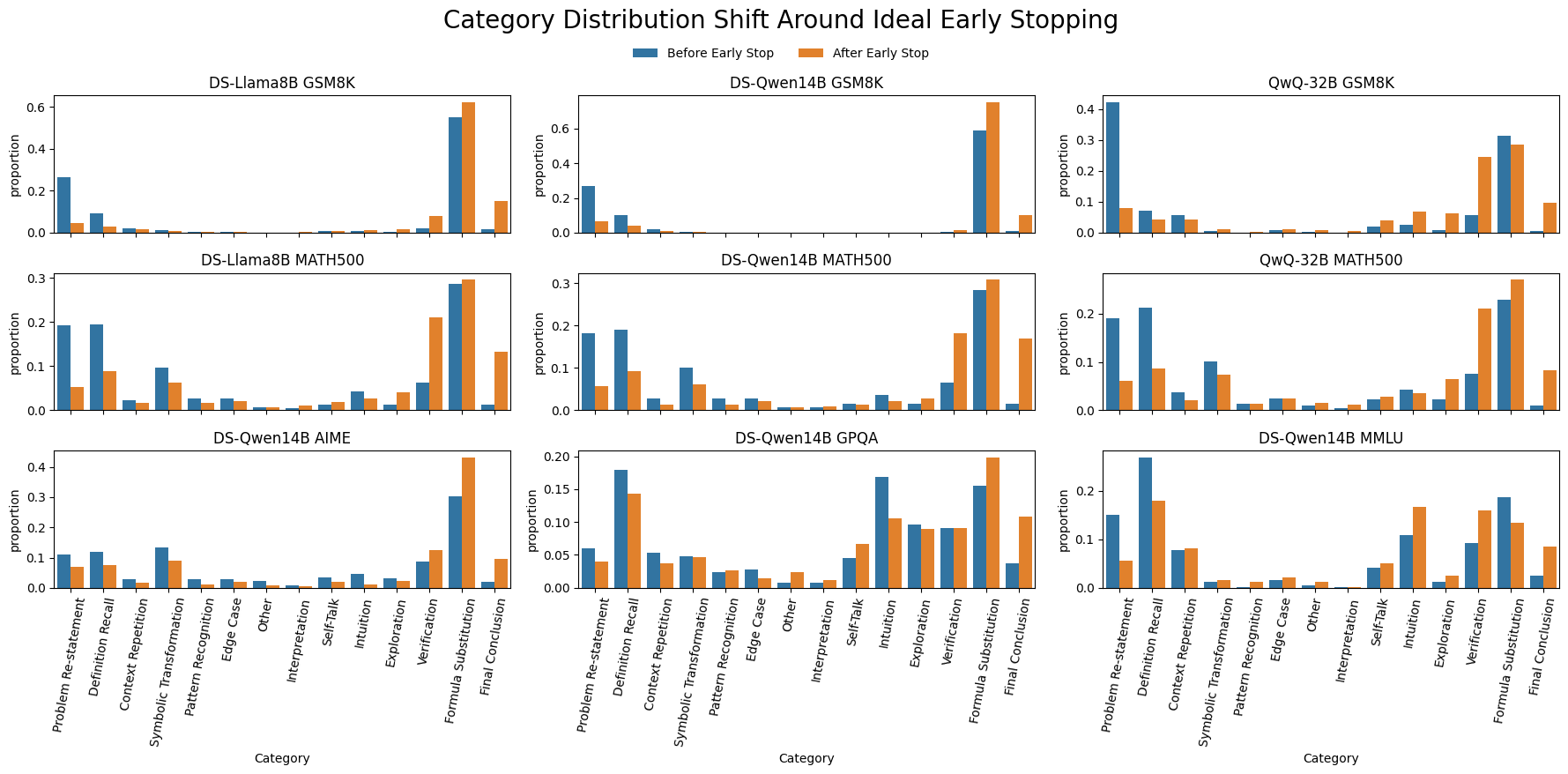}
    \vspace{-0.3cm}
    \caption{\small Step-Type distribution before and after the Ideal Early Stopping step $S_{IES}$ - Categories are ordered by decreasing value of distribution difference before and after $S_{IES}$, from DS-8B GSM8K}
    \label{fig:distribution_before_after}
    \vspace{-0.2cm}
\end{figure}

\textbf{Distribution shift of the step-types after Pearl Reasoning.} On the strength of this observation, we analyse the model behavior after reaching the correct answer. To do so, we labeled the reasoning traces, and separated the steps happening before and after the $S_{IES}$ step. Specifically, we define $S_{\text{before}} = \{s_1, \dots, s_{IES}\}$ and $S_{\text{after}} = \{s_{IES+1}, \dots, s_T\}$, where $S = \{s_1, \dots, s_t\} = S_{\text{before}} \bigcup S_{\text{after}}$. Figure \ref{fig:distribution_before_after} shows the step-type frequency distribution of $S_{\text{before}}$ and $S_{\text{after}}$ across samples, in blue and orange, respectively. 

In Figure \ref{fig:distribution_before_after}, we observe a common pattern across most models and datasets. Step-types such as \emph{Problem-Restatement} or \emph{Definition Recall} represent a high proportion of step-types before the model generates a correct answer for the first time (bars in orange), and their proportion drop after (bars in blue). By the nature of the labels, we refer to these step-types as \emph{Constructive}. Conversely, other step-types such as \emph{Verification} or \emph{Final Conclusion} demonstrate opposite behavior: those steps are not present before the $S_{IES}$ step, but are of high proportion after. We refer to these labels as \emph{Evaluative}.

\textbf{From distribution shift to detecting reasoning phase transition.} Building on this observation, we are looking to observe the distribution shift more closely through the reasoning process. Let $\tau_{\text{constructive}} = \{\tau_{1}, \dots, \tau_{n}\}$ and $\tau_{\text{evaluative}} = \{\tau_{1}, \dots, \tau_{m}\}$ be two classes of tags, including step-types typically occurring before and after the $S_{IES}$ step in the reasoning flow, respectively. Equation \ref{eq:es-criteria} defines an early-stopping criteria such as:
\begin{equation} \label{eq:es-criteria}
    R_i = \frac{\sum_{j=1}^{{i}} \mathbf{1}[\tau_j \in \tau_{\text{constructive}}]}{\sum_{j=1}^{{i}} \mathbf{1}[\tau_j \in \tau_{\text{constructive}}] + \mathbf{1}[\tau_j \in \tau_{\text{evaluative}}]}
\end{equation}
For a running sequence of steps $S_{\text{running}} = \{s_1, \dots, s_i\}$, $R_i$ represents the cumulative ratio of the \emph{constructive} steps compared to the sum of \emph{constructive} and \emph{evaluative} steps. From Figure \ref{fig:distribution_before_after}, we expect $R_i$ to suddenly decrease when the reasoning switch from \emph{constructive} to \emph{evaluative} state. Indeed, $\tau_{\text{constructive}}$ is rare after $S_{IES}$, and $\tau_{\text{evaluative}}$ is frequent. From Figure \ref{fig:distribution_before_after}, we observed that $\tau_{\text{constructive}}$ and $\tau_{\text{evaluative}}$ are common across models and datasets. For the rest of the paper, we fix $\tau_{\text{constructive}} = \{\tau_{\text{Problem-Restatement}}, \tau_{\text{Definition Recall}}\}$ and $\tau_{\text{evaluative}} = \{\tau_{\text{Verification}}, \tau_{\text{Final Conclusion}}\}$ across selected models and datasets.

\textbf{Quantifying the reasoning phase transition.} Figure \ref{fig:reasoning_transition} shows the average ratio $R$ across samples, for the selected configurations. Each datasets leads to different numbers of steps and ideal early-stopping indices. For this reason, we normalized the reasoning process. We set the x-axis as $\frac{|S_i|-|S_{IES}|}{|S|}$, where $|S_i|$ is the step index, $|S_{IES}|$ the index of the IES, and $|S|$ the overall number of step. Negative and positive values represent the reasoning flow before and after the model gets the answer, respectively, with $x=0$ highlighting the $S_{IES}$ steps. In Figure \ref{fig:reasoning_transition}, the ratio $R$ is high and relatively constant for most configurations before $S_{IES}$.

\begin{minipage}{0.4\linewidth}
    This confirms that before $S_{IES}$, the model tends to generate steps responsible for shaping its answer, and rarely generates steps verifying its current solution. Right after $S_{IES}$, we observe a sudden drop in $R$ values. It shows that after $S_{IES}$, the model transition from constructing its answer, to evaluating and verifying it. 
\end{minipage}
\hfill
\begin{minipage}{0.6\linewidth}
    \centering
    \vspace{-0.3cm}
    \includegraphics[width=0.9\linewidth]{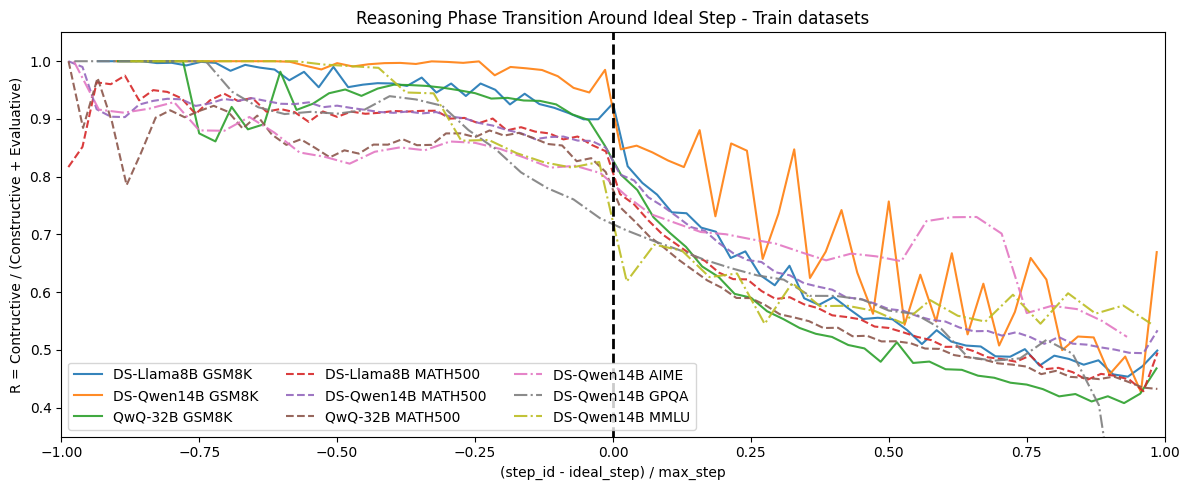}
    \vspace{-0.3cm}
    \captionof{figure}{Ratio $R$ - Reasoning Phase Transition}
    \label{fig:reasoning_transition}
\end{minipage}

We interpret this result as a shift in the reasoning mode of LRMs. Before getting to the correct answer, the model generates specific step-types helping to shape and solve the problem ($\tau_{\text{constr.}}$). Once it got the answer, the model shift from constructing its answer to evaluating it ($\tau_{\text{eval.}}$). Figures \ref{fig:answer-switch-rate} and \ref{fig:answer-correctness-rate} in Appendix \ref{sec:appendix-ideal-early-stopping} further support our interpretation. Our results across the training datasets indicate that the ratio $R$ constitutes a good signal for tracking the reasoning phase transition, and estimating the $S_{IES}$ step. While the analysis presented in this section is conducted on the training datasets, the following section will validate this signal on the held-out test reasoning datasets.




\section{ST-ES: an interpretable early-stopping condition} \label{sec:st-es-evaluation}

Building on our observations from Section \ref{sec:influence-step-type}, we examine the ratio $R$ as a good early-stopping signal. In this section, the central challenge that we address is to determine when to stop the generation of LRMs based on the value of $R$, creating an interpretable stopping criterion. We first suggest a stopping-criteria based on our results from Section \ref{sec:influence-step-type}. We then show that a lightweight sentence classifier can enable online estimation of the ratio $R$. We conclude by demonstrating its effectiveness on the test datasets.

\textbf{Early-stopping criteria.} Assuming that our Step-Tagging module can effectively monitor the steps (Equation \ref{eq:step_tagging}), we can define a constraint $c_{\tau^*}$ on $R$ computed from the monitoring of our model. Each constraint operates online, over a running sequence of reasoning steps $S_{\text{running}} = \{s_1, \dots, s_i\}$, where step $s_j$ has a tag $\tau_j \in \tau^* = \{\tau_{\text{constructive}}, \tau_{\text{evaluative}}, \tau_{\text{others}}\}$:
\begin{equation}\label{eq:early_stopping_step_tag}
    c_{\tau^*}(S_{\text{running}}, \delta) = \mathbf{1}[R(S_{\text{running}}, \tau_{\text{constructive}}, \tau_{\text{evaluation}}) > \delta] 
\end{equation}
$c_{\tau^*}(S_{\text{running}}, \delta)$ is the constraint on the ratio $R_i$ over the step-sequence $S_{\text{running}}$ being generated, given the threshold $\delta$. While the constraint $c_{\tau^*}$ is satisfied, the generation continues. If the constraint is violated for $w$ consecutive steps, the generation stops (see Algorithm \ref{alg:st_es_algorithm} in Appendix \ref{sec:appendix-es_algorithm} for more implementation details). From Figure \ref{fig:reasoning_transition}, we set $\delta \in [0.9, 0.8, 0.7, 0.6, 0.5, 0.4]$, and $w=5$ for all configurations (models and datasets).

\textbf{Answer forcing.} To early-exit the generation of LRMs, we adopted common practices and prompted LRMs right after the stopping signal to retrieve their current best answer with an additional budget of $100$ tokens. We borrowed this approach from \citet{muennighoff2025s1simpletesttimescaling}, who showed that such intervention helped to adapt the computation of models. 





\textbf{Lightweight Step-Tagger for estimating ratio $R$.} Given that our reasoning step taxonomy was created using GPT-4o-mini \citep{openai2024gpt4technicalreport} the most direct way to label a reasoning trace would be to use GPT-4o-mini. However, this GPT-4o-mini annotation is costly, each step requiring around a second to be annotated (see Table \ref{tab:annotation-steps-data} in Appendix \ref{sec:appendix-labeling-process}). Consequently, instead, we use GPT-4o-mini to label a dataset of reasoning traces with the labels from the taxonomy that we use to train a light weight reasoning step classifiers. We construct training datasets by running each LRMs on the training datasets (with a seed of $42$). For each step $s_i$ in generated outputs, we prompted GPT-4o-mini to assign a tag $\tau_i$. Appendix \ref{sec:appendix-reliability-annotation} confirms the reliability of GPT-4o-mini to annotate the reasoning steps.

\textbf{Universal Step-Tagger.} To estimate the ratio $R$ and build our early-stopping criterion defined in Equation \ref{eq:es-criteria} and \ref{eq:early_stopping_step_tag}, respectively, we need a way to identify the $\tau_{\text{constructive}}$ and $\tau_{\text{evaluative}}$ steps. To do so, we trained a BERT classifier on a 3-class classification problem (see Appendix \ref{sec:appendix-step-tagging-training} for details). Importantly, we want our method to generalize across models and datasets. To this extent, we train our classifier on the traces of \deepseekQwen{} across MATH500 and GPQA train split (mathematical and in-domain data). We then evaluate our classifier on traces of the same model on three held-out datasets (GSM8K, AIME, and GPQA), and on the two other models on two datasets (MATH500 and GSM8K). 

\begin{minipage}{0.35\linewidth}
    Figure \ref{fig:step-tagging-performance} present the performance of our step-tagging module on the training reasoning traces. We observe a good performance of the held-out test from the training data, with Macro-F1 ranging from $0.82$ to $0.83$. Importantly, we observe that our module generalizes well on others models and data (Macro-F1 from $0.76$ to $0.88$).
\end{minipage}
\hfill
\begin{minipage}{0.65\linewidth}
    \centering
    \includegraphics[width=0.94\linewidth]{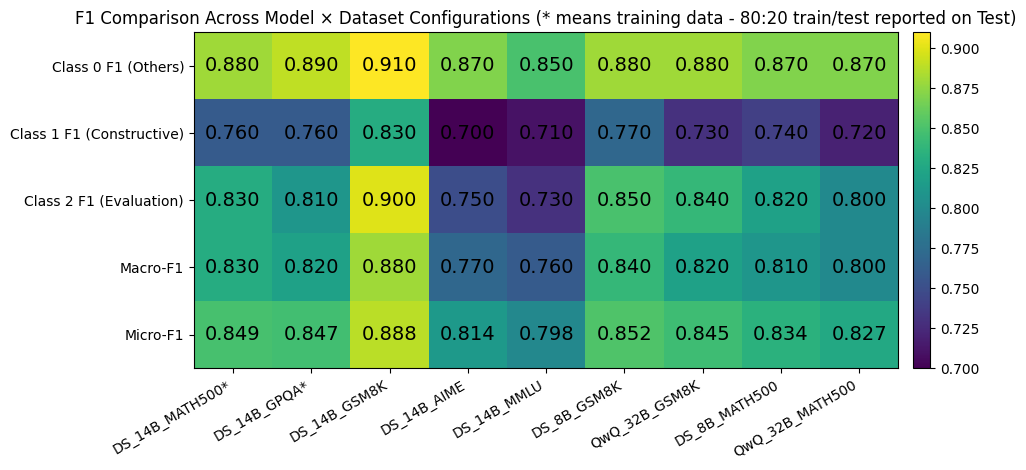}
    \vspace{-0.3cm}
    \captionof{figure}{Performance of the Step-Tagging module} 
    \label{fig:step-tagging-performance}
\end{minipage}

We interpret the strong performance of the classifiers as validating our approach, as well as our reasoning step taxonomy in the sense that it indicates that the step types are distinct. As well, we managed to build universal step-tagging modules -- good generalization on unseen models and datasets -- demonstrating that the training efforts can significantly be reduced. 



\begin{figure}[t]
    \centering
    \begin{subfigure}{0.32\linewidth}
        \centering
        \includegraphics[width=\linewidth]{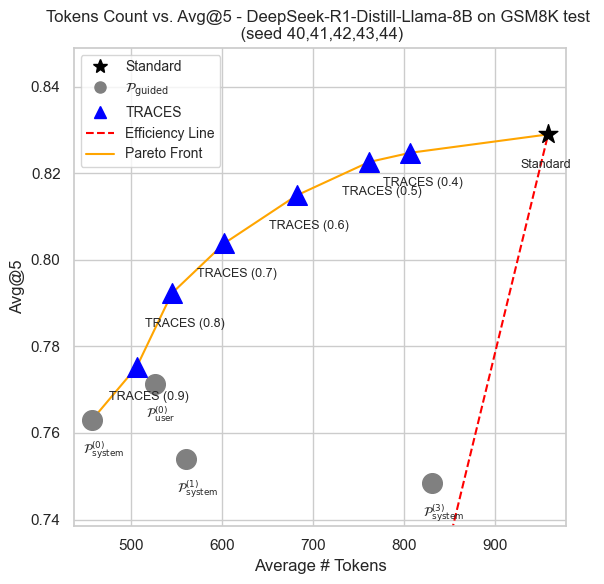}
        \caption{DS-Llama8B on GSM8K}
        \label{fig:DS8B-GSM8K} 
    \end{subfigure}
    \hfill
    \begin{subfigure}{0.32\linewidth}
        \centering
        \includegraphics[width=\linewidth]{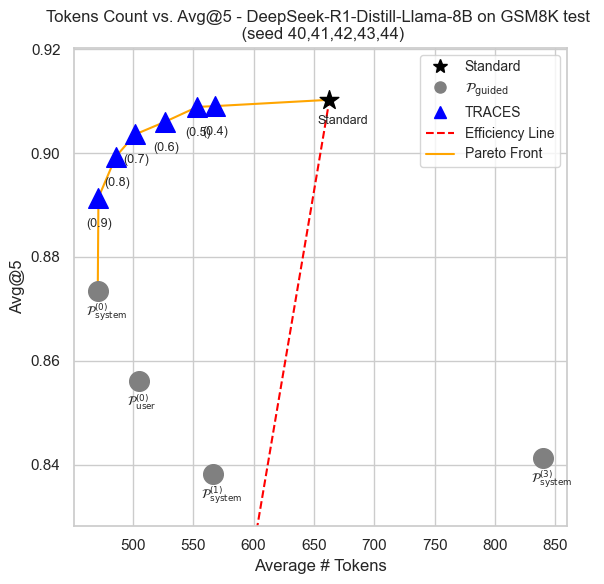}
        \caption{DS-Qwen14B on GSM8K}
        \label{fig:DS14B-GSM8K}
    \end{subfigure}
    \hfill
    \begin{subfigure}{0.32\linewidth}
        \centering
        \includegraphics[width=0.97\linewidth]{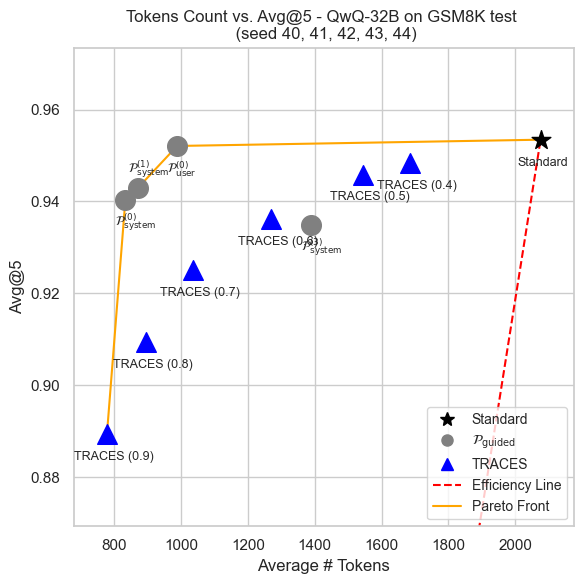}
        \caption{QwQ-32B on GSM8K}
        \label{fig:QwQ32B-GSM8K}
    \end{subfigure}
    \begin{subfigure}{0.32\linewidth}
        \centering
        \includegraphics[width=\linewidth]{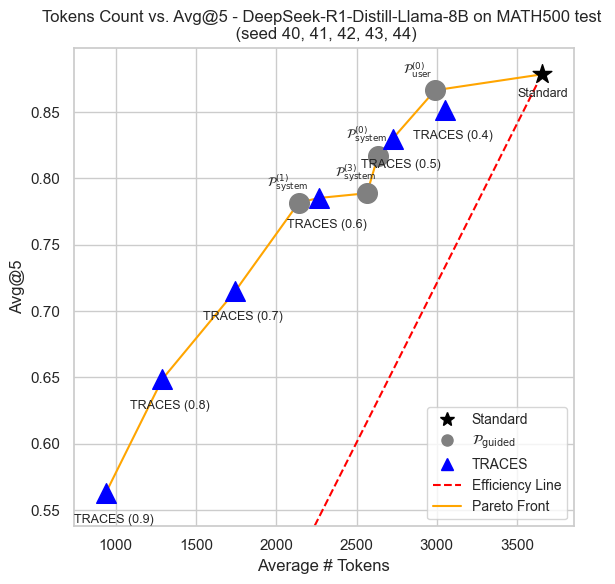}
        \caption{DS-Llama8B on MATH500}
        \label{fig:DS8B-MATH500} 
    \end{subfigure}
    \hfill
    \begin{subfigure}{0.32\linewidth}
        \centering
        \includegraphics[width=\linewidth]{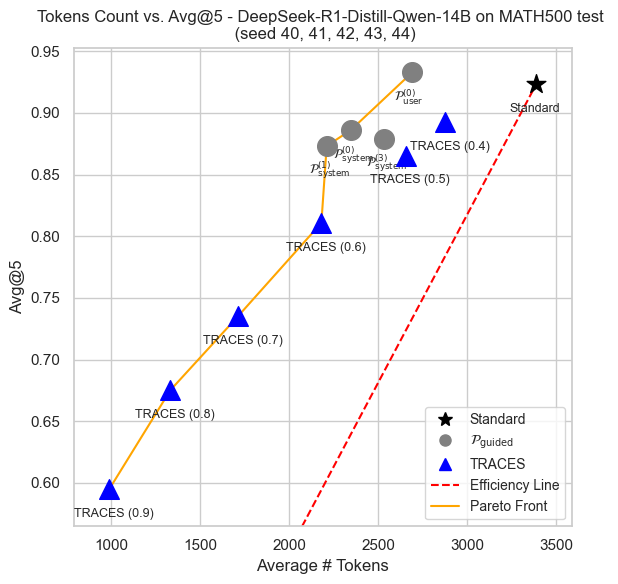}
        \caption{DS-Qwen14B on MATH500}
        \label{fig:DS14B-MATH500}
    \end{subfigure}
    \hfill
    \begin{subfigure}{0.32\linewidth}
        \centering
        \includegraphics[width=0.97\linewidth]{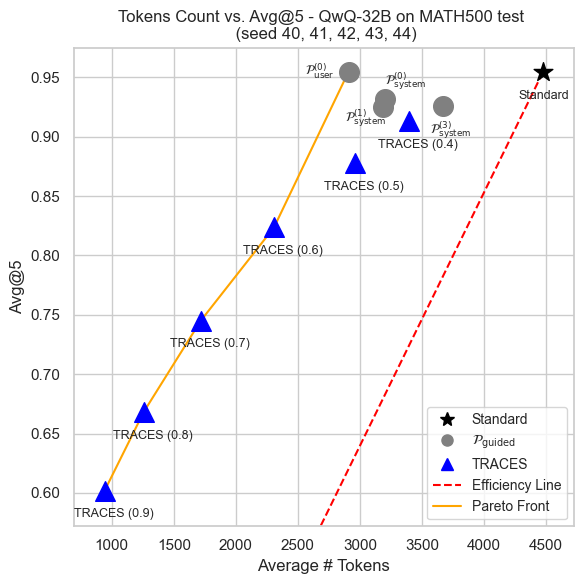}
        \caption{QwQ-32B on MATH500}
        \label{fig:QwQ32B-MATH500}
    \end{subfigure}
    \\
    \begin{subfigure}{0.32\linewidth}
        \centering
        \includegraphics[width=0.95\linewidth]{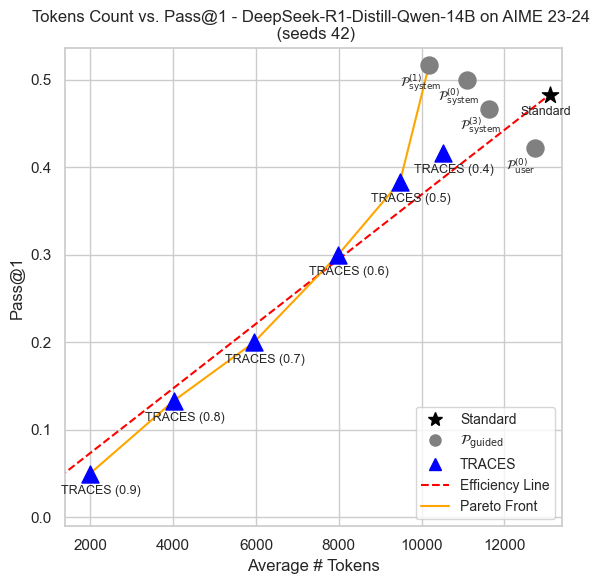}
        \caption{\small DS-Qwen14B on AIME}
        \label{fig:DS14B-AIME} 
    \end{subfigure}
    \hfill
    \begin{subfigure}{0.32\linewidth}
        \centering
        \includegraphics[width=\linewidth]{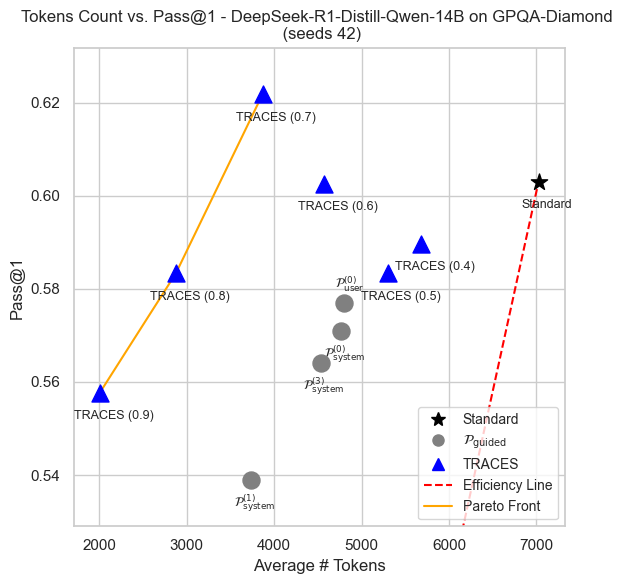}
        \caption{\small DS-Qwen14B on GPQA}
        \label{fig:DS14B-GPQA}
    \end{subfigure}
    \hfill
    \begin{subfigure}{0.32\linewidth}
        \centering
        \includegraphics[width=0.95\linewidth]{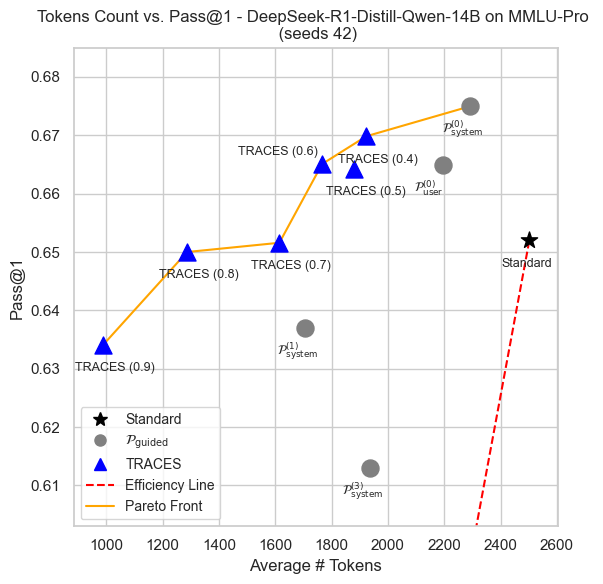}
        \caption{\small DS-Qwen14B on MMLU-Pro}
        \label{fig:DS14B-MMLU}
    \end{subfigure}
    \vspace{-0.2cm}
    \caption{\small Number of Tokens vs. Avg@$5$ and Pass@$1$ - $\mathcal{P_{\text{guided}}}$ Baselines vs. TRACES criteria on test datasets - The efficiency lines in red highlight the configurations that improve the efficiency relative to the standard inference, while the Pareto frontiers in yellow show the most efficient approaches. TRACES achieved up to $20-50$\% token-count saving with minimal accuracy loss.}
    \label{fig:st-es-performance}
    \vspace{-0.5cm}
\end{figure}

\textbf{TRACES framework.} Next, we show in this section that Step-Tagging modules can effectively be used as an early-stopping criteria. Figure \ref{fig:st-es-performance} presents the average token count against the Avg@$5$ for the three LRMs on the MATH500 and GSM8K datasets, as well as the Pass@$1$ of DS-14B on AIME, GPQA and MMLU. Each plot compares the performance trade-offs between the baselines and the ST-ES criteria. Tables \ref{tab:full_results_5_seeds} and \ref{tab:other_dataset} in the Appendix \ref{sec:appendix-st-es-performance} report the quantitative metrics of the baselines and our approach on all configurations.

\noindent \textbf{$\mathcal{P_{\text{guided}}}$ baselines.} We first notice that simple instruction on the models results in strong token-reduction, achieving $20$\% to $60$\% saved tokens across configurations. Specifically, the baselines are giving much better results on \QwQLarge{}, and the system-prompt variants generally lead to more token-reduction for the Deepseek models. While our prompt-guided baselines provide a fair point of comparison under equivalent setting (TRACES and prompt-guided baselines requires no prior knowledge on the budget needed), we further compare our TRACES framework against token-count 
and white-box
early-exit baselines in Appendix \ref{sec:appendix-token-count-baselines} 
and \ref{sec:appendix-white-box-baselines},
respectively, demonstrating competitive performance at matched token-count.

\textbf{Performance of TRACES.} Next, we observe that our TRACES criteria effectively leads to more efficient generation, with all TRACES settings lying on the left side of the Efficiency line compared to the Standard inference for all models. Furthermore, the TRACES criteria appear to outperform most $\mathcal{P_{\text{guided}}}$ baselines for both Deepseek models. Indeed, our TRACES criteria is performing well on the \texttt{DS-Llama8B} model on both MATH500 and GSM8K since almost all TRACES configurations lie on the Pareto front. On GSM8K (Figure \ref{fig:DS8B-GSM8K}), ST-ES with $\delta = 0.8$ achieved approximately the same token reduction as \Puser{} and \Psystemone{} (around $42\%$), while maintaining higher Avg@$5$ ($0.79$ vs. $0.775$ and $0.754$). On MATH500 (Figure \ref{fig:DS8B-MATH500}), TRACES $\delta$=$0.5$ achieves the same token reduction as \Psystemzero{} and \Psystemthree{} (around $30\%$) while maintaining higher Avg@$5$ ($0.83$ vs. $0.82$ and $0.79$, respectively). \texttt{DS-Qwen14B} and \QwQLarge{} also show good performance, obtaining similar token-count gains. 
%
We further evaluate TRACES with three additional models, including DeepSeek-R1-0528-Qwen3-8B, Qwen3-8B, and gpt-oss-120b on MATH500 and GPQA-Diamond, achieving 10 to 40\% token-count reduction with minimal accuracy drop ($\approx -2.4$ to $-6.6$\% see Table \ref{tab:TRACES-other_models}, Appendix \ref{sec:appendix-additional-models}).


\textbf{Generalization to other tasks.} To assess robustness beyond MATH500 and GSM8K dataset, we evaluate our framework to AIME, as well as GPQA-Diamond and MMLU-Pro on DS-14B. Our results indicate that TRACES scales well on these more complex tasks, specifically on GPQA and MMLU, showing that reasoning-step monitoring is not tied to mathematical structure.
%
To further test this claim on harder tasks, we evaluate TRACES on harder and less contaminated benchmarks, namely BeyondAIME and IMO AnswerBench using Qwen3-8B and gpt-oss-120b (see Table \ref{tab:TRACES-other_datasets}, Appendix \ref{sec:appendix-harder-datasets}). On these tasks, TRACES achieves more modest token-count savings (10-37\%), with stronger accuracy drop
(4-14\% and 1.8-5.5\% accuracy drop on BeyondAIME and IMO AnswerBench, respectively), confirming that the efficiency-accuracy trade-off becomes steeper as task difficulty increases.


\textbf{Adaptability of TRACES.} Our findings supports that the value of the threshold can help controlling the computations. While we systematically find across models and standard datasets that $\delta \in [0.4, 0.6]$ and $[0.7, 0.9]$ offers around $20-35\%$ and $40-70\%$ token-count reduction respectively, the performance decreases more sharply on harder reasoning tasks, and accuracy drop at a given $\delta$ is not always minimal. 
%
Our extended results on BeyondAIME and IMO AnswerBench (see Table \ref{tab:TRACES-other_datasets}, Appendix \ref{sec:appendix-harder-datasets}) show that for tasks requiring very large token budgets, more conservative thresholds ($\delta \in [0.2, 0.4]$) are needed to keep accuracy loss minimal, and larger accuracy drop should be expected regardless.
It supports the view that harder reasoning tasks need less restrictive (lower) thresholds, as more evaluative steps are required to maintain accuracy. 
%
Further, it shows that our earlier framing of comparable accuracy applies primarily to standard benchmarks rather than across difficulty levels (especially for extreme tasks).
We extend this discussion in Appendix \ref{sec:appendix-infl-fact-threshold}.

\textbf{Influencing factors of TRACES.} To assess the robustness of our framework, we conducted ablation studies to evaluate the influence of the nature of the classes used by the Ratio $R$, on the step-type taxonomy, on the performance of Step-Taggers, 
and on the cross-model transferability of our classifiers
in Appendices \ref{sec:appendix-infl-fact-classes-ratio}, \ref{sec:appendix-infl-fact-taxonomy}, \ref{sec:appendix-infl-fact-perf-step-tagg}, \ref{sec:appendix-classifiers-new-models} respectively. We summarized our findings in Appendix \ref{sec:summary-TRACES}. 
%
We also report an online implementation of TRACES, evaluating true step-by-step decoding latency separately from our main offline experiments (Appendices \ref{sec:appendix-latency-analysis} and \ref{sec:appendix-online-latency}), as well as the computational overhead and the training-inference cost of our framework in Appendices \ref{sec:appendix-computational-overhead} and \ref{sec:appendix-training-inference}, respectively.

\section{Limitations and Future Work} \label{sec:appendix-limitation_futurr_work}

Our definition of reasoning step is taken from previous work, and relies on heuristics from the model (generation of ``\textbackslash n \textbackslash n" tokens). While adopted by numerous papers in the literature \citep{lightman2023letsverifystepstep,zhang2025reasoningmodelsknowtheyre,Park_2024}, we believe that our step taxonomy can enhance the step definition. Future work should look at leveraging the performance of step-classification to better define reasoning steps.

To train accurate Step-Tagger modules, we suspect that significantly increasing the number of traces could lead to better results. Also, down-sampling could render our training more effective, and increase the Macro-F1. In addition, a better definition of a step could lead to more effective monitoring. For instance, it would be interesting to explore dynamic step segmentation. However, current dynamic definitions are expansive (e.g rendering the method gray/black-box  and/or increasing the latency).

As well, our current TRACES criteria is based on a fixed threshold. Another line of work that could be applied to our setting explored dynamic monitoring of time-series signal. We believe that integrating a dynamic anomaly detection method to our criteria (to better identify when the ratio $R$ drops) could further enhance our early-stopping criteria.

Further, our ReasonType taxonomy and all reasoning-step annotations rely exclusively on GPT-4o-mini, validated against other LLM annotators (GPT-4o, Llama-3-70B, Mistral-8x22B, see Appendix \ref{sec:appendix-reliability-annotation}). Even though the taxonomy is manually curated, we did not conduct a human annotation study to verify the ground-truth quality of these labels. Our inter-model agreement analysis suggests that the annotations are consistent with a stronger model (Fleiss' kappa = 0.780 for GPT-4o-mini inner-model aggregation, Cohen' Kappa = 0.601 for inter-model aggregation against GPT-4o). However, these studies only partially confirm that the annotation process aligns with human interpretations. Future work should conduct a follow-up annotation study with human raters to measure: (1) the inter-annotator agreement among humans, and (2) measure the agreement between human annotators and GPT-4o-mini on the same samples.

\section{Conclusion}

This work offers a novel view on both monitoring and efficiency of LRMs. We propose \emph{ReasonType}, a novel taxonomy of reasoning steps, and demonstrated that users can effectively track the reasoning flow of the generation using our \emph{TRACES} framework. Furthermore, we show that differentiating the step-type in the generation of LRMs is important, and can be used as a reliable and interpretable early-stopping criterion. Through careful monitoring of specific step-types, our framework can enhance the control of the generation of RLMs enabling a significant token-count saving (up to 50\%) 
with minimal accuracy loss on standard benchmarks. On harder benchmarks, savings remain positive (up to 30\%), but come with steeper accuracy trade-off, requiring more conservative early-stopping.

\section*{Acknowledgment}

This work has been partially supported by the 6G-XCEL project (grant agreement 101139194), funded by the EU Horizon Europe program.

This research was partly supported by the ADAPT Research Centre. The ADAPT Centre for Digital Content Technology is funded under the Research Ireland’s Research Centres Programme (Grant 13/RC/2106 II) and is co-funded under the European Regional Development Funds.


\bibliography{colm2026_conference}
\bibliographystyle{colm2026_conference}

\newpage

\appendix
\doparttoc
\faketableofcontents  
\addcontentsline{toc}{section}{Appendix} 
\part{Appendix} 
\parttoc

\newpage

\section{LLM Usage}

We acknowledge the use of Large Language Models for the purpose of our experimentation in our paper. Specifically, as stated in Section \ref{sec:monitoring-lrms}, we relied on \GPTmini{} to set our \emph{ReasonType} taxonomy. This approach is borrowed from work on behavior analysis of LLMs, such as \citet{galichin2025icoveredbaseshere, kuznetsov2025featurelevelinsightsartificialtext}.

\section{Reproducibility statement}

We took several measures to ensure the reproducibility of our experiments, namely:

\begin{itemize}[leftmargin=*, itemsep=0em, topsep=0pt]

    \item \textbf{Online implementation:} To demonstrate the effectiveness of our TRACES framework, we recorded a video demo of its usage using Ollama and Streamlit. We included the video demo in the ZIP folder. We also included the code to run our framework locally, as well as a README.md section helping to set-up the environment.

    \item \textbf{Code availability:} The source code that we developed to conduct our experiments is available in the submission ZIP folder. The source code also include a \model{requirement.txt}, allowing users to create an environment with the correct versions of the libraries we used.

    \item \textbf{Experimental Settings:} We listed in Section \ref{sec:experimental-section} the experimental settings. This includes the datasets used, the models (open-source available on HuggingFace), the parameters of the algorithms, the prompts of the models, and the environment setups (seed and hardware used). We also included scripts to reproduce the experiments we conducted.

\end{itemize}

\newpage

\section{Experimental Setup}

\subsection{Step definitions} \label{app:step-def}

From the literature, we identified four principal methods to segment the output from a model into distinct thoughts:

\begin{itemize}[leftmargin=*, itemsep=0pt, topsep=0pt]
    \item \textbf{Token or sentence level:} N\"aively, thoughts can be decomposed into token \citep{yao2023treethoughtsdeliberateproblem} or sentence level \citep{fu2023complexitybasedpromptingmultistepreasoning}. However, for complex reasoning problems, these definitions are not ideal since reasoning steps are composed of multiple sentences in mathematical reasoning.

    \item \textbf{Paragraph level:} LLMs and LRMs such as Deepseek-R1, QwQ, or GPT are natively generating back-to-line symbols between two thoughts (e.g. \text{.\texttt{\textbf{\textbackslash n\textbackslash n}}}). Since this observation is model agnostic, it has been adopted by several works \citep{cao2025stepguidedreasoningimproving, Park_2024, lightman2023letsverifystepstep}. This definition seems to be the most adopted in the current literature.

    \item \textbf{Dynamic step segmentation:} Another common approach is to prompt the model to force the generation of special tokens to split the thoughts (e.g. \text{\texttt{<next\_step>}}). While some works have used this strategy \citep{zelikman2024quietstarlanguagemodelsteach, sui2025metareasonerdynamicguidanceoptimized, paul2024refinerreasoningfeedbackintermediate}, it suffers from  low reliability and efficiency. Indeed, this approach artificially generates more tokens, and prompt engineering could cause mistakes since models are not pre-trained to perform this sub-task. Furthermore, \cite{liu2025adaptivestepautomaticallydividingreasoning} suggest a dynamic segmentation based on the model's entropy. However, this methods often lead to split steps between equations, and requires the model's internals.

    \item \textbf{Prompting segmentation:} Another approach is to prompt a model to segment its output into steps. \citet{luo2025deconstructinglongchainofthoughtstructured} prompts a model to parse the raw reasoning output into structured steps. \citet{golovneva2023roscoesuitemetricsscoring} prompts the model to generate step-by-step reasoning. Similarly as the previous method, this approach rely on a model to perform the step segmentation.
\end{itemize}

To clearly identify and monitor reasoning steps, the literature does not agree on a consensus regarding the definition of a reasoning step. However, the segmentation using delimiters (e.g. \text{.\texttt{\textbf{\textbackslash n\textbackslash n}}}) appears to be the well adopted by recent research \citep{zhang2025reasoningmodelsknowtheyre, Park_2024, lightman2023letsverifystepstep}. For this reason, we adopted the latter definition for our paper.

\subsection{Datasets and Metrics} \label{app:appendix-metrics}

\begin{table}[h]
    \tiny
    \centering
    \begin{tabular}{ccccccc}
    \toprule
    \multirow{2}{*}{\textbf{Domain}} & \multirow{2}{*}{\textbf{Ref.}} & \multirow{2}{*}{\textbf{Name}} & \multirow{2}{*}{\textbf{Train Size}} & \multirow{2}{*}{\textbf{Test Size}} & \multicolumn{2}{c}{\textbf{Type of evaluation}} \\
    \cmidrule(lr){6-7} 
     &  &  &  &  & Math-Verify & MCQ Prompting \\
    \midrule
       \multirow{3}{*}{Maths} & \cite{cobbe2021gsm8k} & GSM8K & $3,000$ & $1,329$ & \cmark &  \\
        & \cite{hendrycks2021measuringmathematicalproblemsolving} & MATH500 & $1,000$ & $500$ & \cmark &  \\
        & \cite{aime2025dataset} & AIME 22-24 & $30$ & $60$ & \cmark &  \\
    \midrule
        \multirow{2}{*}{In-domain} & \cite{rein2023gpqagraduatelevelgoogleproofqa} & GPQA-Diamond & $40$ & $158$ &  & \cmark \\
        & \cite{wang2024mmluprorobustchallengingmultitask} & MMLU-Pro & $240$ & $1,200$ &  & \cmark \\
    \bottomrule
    \end{tabular}
    \vspace{-0.1cm}
    \caption{Overview of selected reasoning datasets and metrics}
    \label{tab:reasoning-datasets}
    \vspace{-0.2cm}
\end{table}

\noindent \textbf{Evaluating LRMs.} To assess the model's performance on challenging reasoning tasks, the Avg@$k$, Pass@$k$, and Cons@$k$ are common metrics \citep{chen2021evaluatinglargelanguagemodels, chen2025reasoningerasurveylong, yu2025dapoopensourcellmreinforcement}. The Pass@$k$ measures the proportion of the samples where at least one of $k$ attempts leads to the correct answer, while the Cons@$k$ consider a sample correct if all $k$ attempts are correct. Since we are interested about both performance and robustness of our approach, we selected the Avg@$5$, the Pass@$5$ and the Cons@$5$ as the quantitative metrics. Assessing the performance of LRMs on mathematical questions is challenging. This is due to the open nature of the question. For our experiments, we selected the \emph{Math-Verify}\footnote{\url{https://github.com/huggingface/Math-Verify}} library which is a common metric to assess mathematical problems. It uses text extraction and formal verification. This metric also reported strong correctness compared to other evaluation methods such as Harness \citep{eval-harness} or Qwen-Math Verifier \citep{huang2025accuracyrobustnessstudyrule}.

\newpage

\section{Construction and Validation of the ReasonType taxonomy} \label{app:valid-taxonomy}

This appendix presents the methodology that we adopted to create the ReasonType taxonomy. Importantly, this section demonstrates the robustness of our annotation approach using our taxonomy, further validating our overall methodology.

\subsection{Construction of the taxonomy} \label{app:construction-taxonomy}

In Section \ref{sec:experimental-section}, we provided a high-level description of our methodology to generate a taxonomy of reasoning step. In this section, we will describe our methodology in more details. 

\textbf{Methodology.} Our goal in constructing the \emph{ReasonType} taxonomy is to observe the behavior of LRMs and look at aggregating similar types together to follow the generation of LRMs closely. Because no taxonomy of reasoning steps exists, we used an open-ended labeling procedure: we prompted a model to generate free-form step-type labels, and manually merged common labels together. We took inspiration from previous work, who relied on the summarization and behavior detection capabilities of strong models such as \texttt{GPT-4o-mini}, as this method has proven to be effective \citep{galichin2025icoveredbaseshere, kuznetsov2025featurelevelinsightsartificialtext}. 

\begin{table}[h]
\small
\centering
\begin{tabular}{lccc}
\toprule
\textbf{Dataset} & \textbf{Model} & \textbf{\# Steps} & \textbf{\# Unique Tags} \\ 

\midrule

\multirow{2}{*}{MATH500} & \deepseekLlama{} & 3,840 & 162 \\
& \QwQLarge{} & 3,057 & 179 \\

\bottomrule
\end{tabular}
\caption{Step for taxonomy generation}
\label{tab:taxonomy-generation-traces}
\end{table} 

We first generated $100$ reasoning traces from the MATH500 train dataset, using \deepseekLlama{} and \QwQLarge{}, to obtain a pool of reasoning steps ($20$ samples from each complexity level). We obtained a pool of $6,897$ reasoning steps (see Table \ref{tab:taxonomy-generation-traces}). Each step is then passed to \texttt{GPT-4o-mini} using our Taxonomy prompt, presented in Figure \ref{fig:prompt_taxonomy} below. We obtained an open-ended label for each step-type. 

\begin{figure}[h]
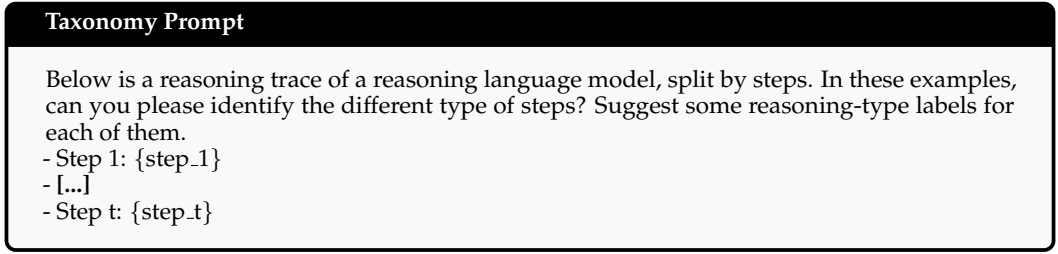

\centering
\footnotesize
\begin{adjustbox}{max width=\textwidth}
\begin{tcolorbox}[colback=gray!5, colframe=black, title=Taxonomy Prompt, fonttitle=\bfseries]

Below is a reasoning trace of a reasoning language model, split by steps. In these examples, can you please identify the different type of steps? Suggest some reasoning-type labels for each of them. \\
- Step 1: \{step\_1\} \\
- \textbf{[...]} \\
- Step t: \{step\_t\}

\end{tcolorbox}
\end{adjustbox}
\caption{Prompt used to generate the Taxonomy}
\label{fig:prompt_taxonomy}
\end{figure}

\begin{figure}[h]
    \centering
    \includegraphics[width=0.95\linewidth]{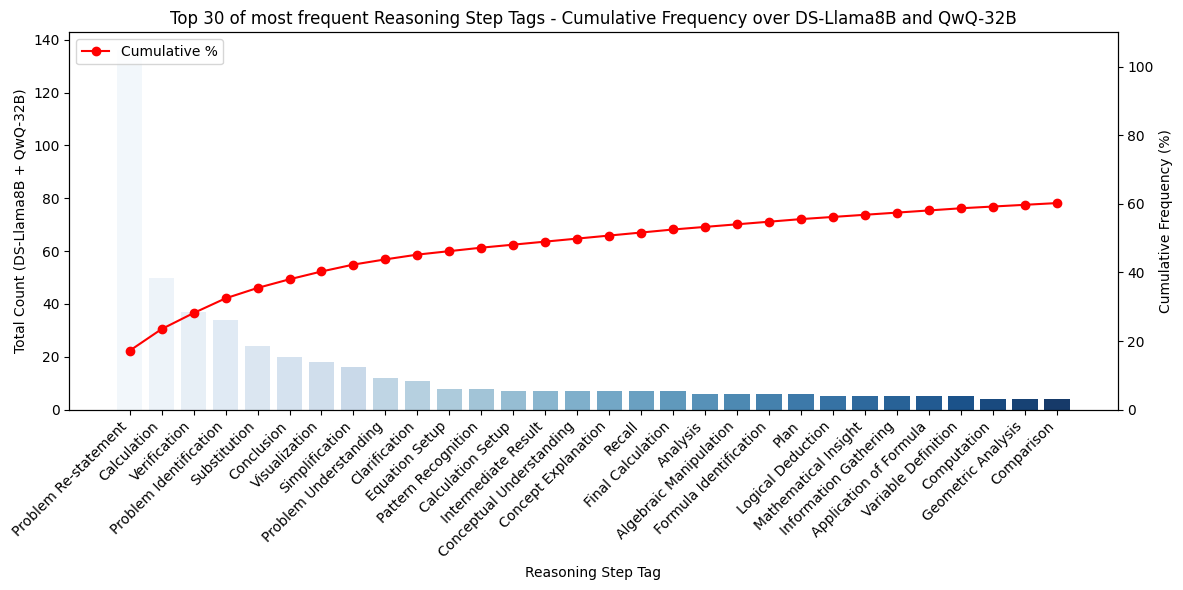}
    \caption{Most frequent labels obtained from open-end label generation}
    \label{fig:frequency-most-occurent-labels}
\end{figure}

Figure \ref{fig:frequency-most-occurent-labels} presents the most frequent labels generated. We observe that, even though our prompting was open-ended (without particular instruction for the model to generate certain type of labels), some labels were obtained frequently, such as \emph{Problem Identification/Re-Statement}, \emph{Substitution}, or \emph{Verification}. We manually inspected the labels obtained, and merged labels that have a common signification. Table \ref{tab:example-of-tag-merged} illustrates several examples. 

\textbf{Takeaways.} While the ReasonType taxonomy might not be optimal, we showed that it reflects the different steps of the model, and offers enough consistency and granularity to analyze the reasoning process of LRMs and to train accurate classifiers for our study. It also echoes previous work, and the categories overlaps with existing taxonomies \citep{galichin2025icoveredbaseshere,marjanovic2026deepseekr1thoughtologyletsthink, minegishi2025topologyreasoningunderstandinglarge, venhoff2025basemodelsknowreason}. Furthermore, the size of our taxonomy (13-14 step-types) is supported by the finding of \citet{venhoff2025basemodelsknowreason}: \emph{"we find [...] that reasoning mechanisms are reasonably well represented using 10 to 20 categories."}.

\textbf{Limitations.} To over-come the domain dependency, we could rely on the OpenThoughts-114k dataset. Indeed, it seems to provide a diverse set of possible behavior for the models. However, it would come at a high processing cost.

\begin{table}[h]
\tiny
\centering
\begin{tabular}{cp{4.2cm}p{4.2cm}}
\toprule
\textbf{Tags from ReasonType} & \textbf{DS-Llama8B} & \textbf{QwQ-32B} \\ 

\midrule

\multirow{2}{*}{Problem Re-statement / Setup} & Problem Re-statement, Problem Identification, Clarification & Problem Re-statement, Coordinate Setup, Step-by-Step Breakdown \\

\midrule

\multirow{1}{*}{Definition Recall} & Recall of Relevant Concepts, Definition Recall & Recall, Rule Recall, Definition Explanation \\

\midrule

\multirow{2}{*}{Formula Substitution} & Value Substitution, Application of Formula, Calculation & Equation Setup, Equation Manipulation, Application of Formula \\

\midrule

\multirow{2}{*}{Exploration} & Approach Exploration, Exploration of Alternatives & Exploration \\

\midrule

\multirow{2}{*}{Self-Talk} & Procedure Explanation, Plan of Action, Comparison of Options & Confirmation, Reflection, Logical Breakdown \\

\midrule

\multirow{2}{*}{Verification} & Confirmation, Verification & Backward Calculation, Example Verification, Assumption Checking \\

\midrule

\multirow{1}{*}{Final Answer} & Final Evaluation, Final Calculation & Final Evaluation, Final Calculation \\

\bottomrule
\end{tabular}
\caption{Example of categories from the ReasonType Taxonmy obtained by merging labels from the annotation - each set of labels were generated using our prompt and sampled reasoning steps, and we manually merged common labels to create the ReasonType taxonomy.}
\label{tab:example-of-tag-merged}
\end{table}

\subsection{Labeling the reasoning traces} \label{sec:appendix-labeling-process}

Table \ref{tab:annotation-steps-data} presents statistics on the number of steps and \texttt{GPT-4o-mini} annotation for each models on the selected datasets. Results are averaged for the seed 42.

\begin{table}[h]
\tiny
\centering
\begin{tabular}{lccccccc}
\toprule
\textbf{Model} & \textbf{Dataset} & \textbf{\# Tok. / Steps} & \textbf{\# Steps / Sample} & \textbf{\# Steps} & \textbf{Runtime / Steps} & \textbf{Runtime / Sample} & \textbf{Tot. Runtime} \\ 

\midrule

\multirow{2}{*}{\texttt{DS-8B}} & GSM8K & 34.58 & 23.64 & 70,911 & 0.87 & 20.71 & 62,143 \\
& MATH500 & 34.11 & 96.27 & 96,270 & 0.88 & 84.86 & 84,861 \\

\midrule

\multirow{5}{*}{\texttt{DS-14B}} & GSM8K & 36.64 & 14.01 & 42,017 & 0.86 & 12.09 & 36,293 \\
& MATH500 & 34.33 & 94.84 & 94,838 & 0.85 & 80.26 & 80,263 \\
& AIME 22-24 & 27.74 & 469.29 & 42,236 & 0.85 & 398.68 & 35,881 \\
& GPQA-Diamond & 38.96 & 186.51 & 36,928 & 0.84 & 156.05 & 30,898 \\
& MMLU-Pro & 35.86 & 64.59 & 90,428 & 0.87 & 56.42 & 78,985 \\

\midrule

\multirow{2}{*}{\texttt{QwQ-32B}} & GSM8K & 45.59 & 40.43 & 121,288 & 0.87 & 35.33 & 105,977 \\
& MATH500 & 38.97 & 117.28 & 117,283 & 0.83 & 97.49 & 97,494 \\

\bottomrule
\end{tabular}
\caption{Avg. \# of steps and annotation runtime per sample - MATH500 and GSM8K are the trained dataset ($1,000$ and $3,000$ samples, respectively). AIME, GPQA and MMLU are the full dataset ($90$, $198$ and $1,400$ samples, respectively). Runtime in seconds.}
\label{tab:annotation-steps-data}
\end{table}

\newpage

\subsection{Reliability of \texttt{GPT-4o-mini} as an annotator} \label{sec:appendix-reliability-annotation}

\textbf{Claim.} Our framework rely on the annotation capability of the \texttt{GPT-4o-mini} model. Since the model is large and achieved great performance on a range of tasks, we assumed that the model is able to provide us with labels of good quality. In this section, we will observe and analyse the reliability of the annotation of the steps by the \texttt{GPT-4o-mini} model.

\textbf{Methodology.} To verify our claim, we sampled $1,000$ reasoning steps of DS-Qwen14B model from its inference on the MATH500 dataset. We then annotated each steps $5$ times using \texttt{GPT-4o-mini}, and observe the agreement of each annotation. In addition, we also compared the annotation agreement between the \texttt{GPT-4o-mini} model, and 3 additional models: \texttt{GPT-4o} (a larger, closed-source model), \texttt{llama-3-3-70b-instruct}, and \texttt{Mixtral-8x22B-Instruct-v0.1} (both open-source and smaller relatively to the two model selected). This setup allows us to assess both the internal consistency of \texttt{GPT-4o-mini} and its alignment with other model annotators.

\textbf{Experimental Design.} We refer to self-model agreement as the \emph{Inner-model agreement} (agreement between different runs of the same model on the same reasoning steps). The inner-model agreement is measured using the Fleiss' kappa metric. Indeed, the Fleiss' kappa measure the agreement between more than $2$ annotators, making it suitable to compare many annotation trials \citep{Moons_2025}. Furthermore, we call agreement between two different model the \emph{Inter-model agreement}. To compute this, we first selected the most consistent label generated across the different runs for each model, and we measured the Cohen's Kappa \citep{badshah2025referenceguidedverdictllmsasjudgesautomatic}. For this experiment, we used the OpenAI default decoding parameters for the GPT models (temperature = $1.0$ and top-p = $1.0$). Same parameter was set for the other models selected.

\begin{table}[h]
\footnotesize
\centering
\begin{tabular}{lcccc}
\toprule
\textbf{Models} & GPT-4o-mini & GPT-4o & Llama-3-3-70b & Mixtral-8x22B \\ 

\midrule

GPT-4o-mini & \textbf{0.780} & 0.601 & 0.457 & 0.392 \\
GPT-4o &  & \textbf{0.799} & 0.445 & 0.384 \\
Llama-3-3-70b &  &  & \textbf{0.722} & 0.398 \\
Mixtral-8x22B &  &  &  & \textbf{0.587} \\

\bottomrule
\end{tabular}
\caption{Agregation metrics of the annotation process - $1,000$ samples reasoning steps from the \deepseekQwen{} model on the MATH500 dataset -  \textbf{Fleiss' kappa} in bold (diagonal - inner-model aggregation) - \textbf{Cohen's kappa} otherwise (inter-model aggregation)}
\label{tab:reliability-annotation}
\end{table}

\textbf{Internal consistency of \texttt{GPT-4o-mini}.} The Fleiss' Kappa score of $0.780$ (see Table \ref{tab:reliability-annotation} in the diagonal) indicates high agreement among multiple independent runs of \texttt{GPT-4o-mini} on the same reasoning steps. This demonstrates that the model produces stable and consistent annotations when prompted repeatedly. We observe that the Fleiss' Kappa score seems to decrease as the model size becomes smaller.

\textbf{\texttt{GPT-4o-mini} against other models.} When comparing \texttt{GPT-4o-mini} with other models, we observe significant agreement with \texttt{GPT-4o}, with a Cohen's Kappa of $0.601$, and moderate agreement with smaller models ($0.457$ with Llama70b and $0.392$ with Mixtral). The higher agreement with \texttt{GPT-4o} suggests that larger models tend to produce more stable and higher-quality annotation, while smaller models exhibits more variability.

\textbf{Takeaway.} Overall, these results supports that \texttt{GPT-4o-mini} is a reliable annotator for reasoning steps. Indeed, for our annotation task, the model appears to be consistent accross multiple runs, and shows meaningful agreement with other strong models. The results obtained on smaller models confirms our decision of selecting \texttt{GPT-4o-mini} as a annotator to label the reasoning steps of LRMs.

\newpage

\subsection{Reason-Type, a robust taxonomy for identifying reasoning behaviors} \label{sec:appendix-validation-taxonomy-shuffled}


\textbf{Objective.} This ablation study is looking at further validating our ReasonType taxonomy. In other words, we are investigating whether our proposed taxonomy captures meaningful distinctions in reasoning steps. We are looking to demonstrate that: 

\begin{enumerate}[leftmargin=*, itemsep=0pt]

    \item The ReasonType taxonomy enable semantic distinction of the type of reasoning.

    \item Our annotation method with the \texttt{GPT-4o-mini} model, coupled with the ReasonType taxonomy, is a robust method to access to the ground-truth labels of the reasoning steps. 
    
\end{enumerate}

\textbf{Methodology.} To address our objective, we compare the performance of BERT classifiers across Original labels (OG - from \texttt{GPT-4o-mini} annotation using the ReasonType taxonomy), and shuffled labels for three step-types, namely: \emph{Verification}, \emph{Exploration} and \emph{Self-Talk}. For the shuffled labels version, we took the exact same proportion of positive labels as in Original datasets, and used random shuffle with a seed of $42$. Each experiment is run on the same training and held-out testing dataset (see Section \ref{sec:st-es-evaluation}), using the MATH500 training dataset on the DS-Qwen14B model. Instead of training Step-Tagging classifiers (identifying 3-classes, including \emph{constructive} and \emph{evaluative} classes), we trained binary step-taggers following the same training configuration (see Appendix \ref{sec:appendix-step-tagging-training}). We set the label of selected class to $1$, and put other labels to $0$. To compare performances, we report both training loss, and classification metrics (precision and recall on both classes, along with macro and micro average.)

\begin{minipage}{0.4\linewidth}
    \textbf{Evaluation.} Figure \ref{fig:shuffled_training_loss} shows the training loss of the Original and Shuffled versions, for the three labels. Models trained on the Original labels presents significant lower losses, and are smoothly decreasing. It demonstrate that the Original datasets contains meaningful patterns between reasoning steps and their labels. In comparison, the models trained on shuffled labels present almost constant loss, relatively higher than the one from the Original labels.
\end{minipage}
\hfill
\begin{minipage}{0.6\linewidth}
    \centering
    \includegraphics[width=0.9\linewidth]{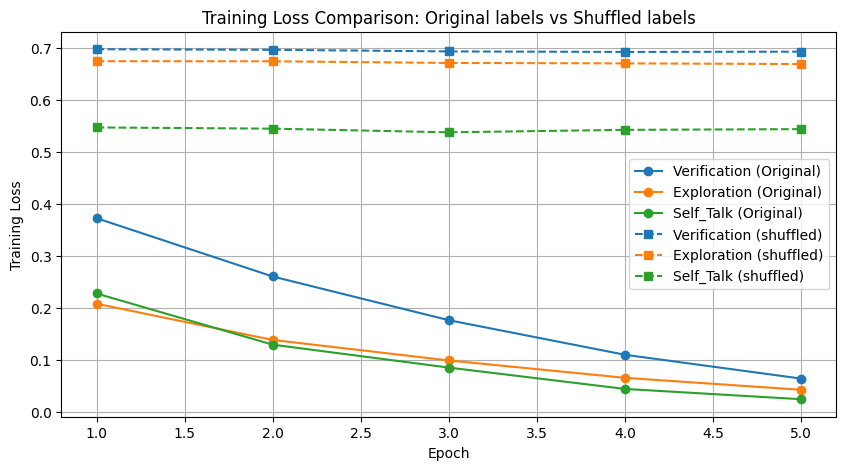}
    \captionof{figure}{\small Training losses - ReasonType vs. Shuffled labels}
    \label{fig:shuffled_training_loss}
\end{minipage}

\begin{figure}[h]
    \centering
    \begin{subfigure}{0.48\textwidth}
        \includegraphics[width=\linewidth]{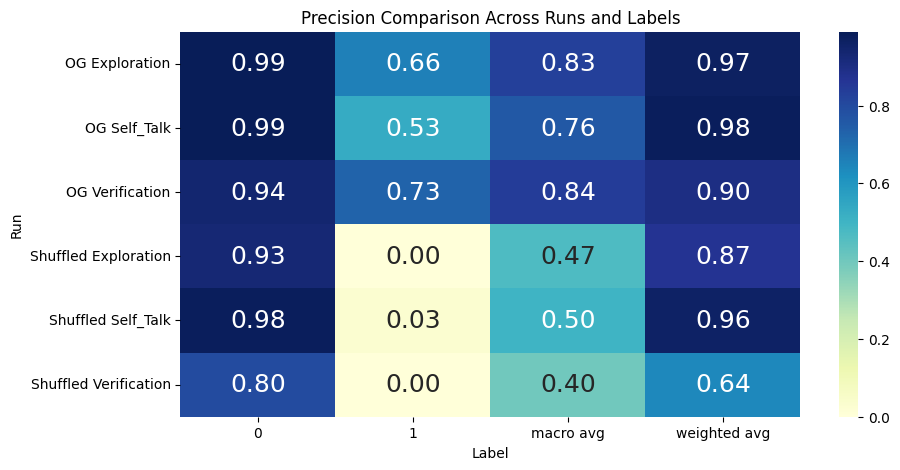}
        \caption{Precision}
        \label{fig:shuffled_precision}
    \end{subfigure}
    \hfill
    \begin{subfigure}{0.48\textwidth}
        \includegraphics[width=\linewidth]{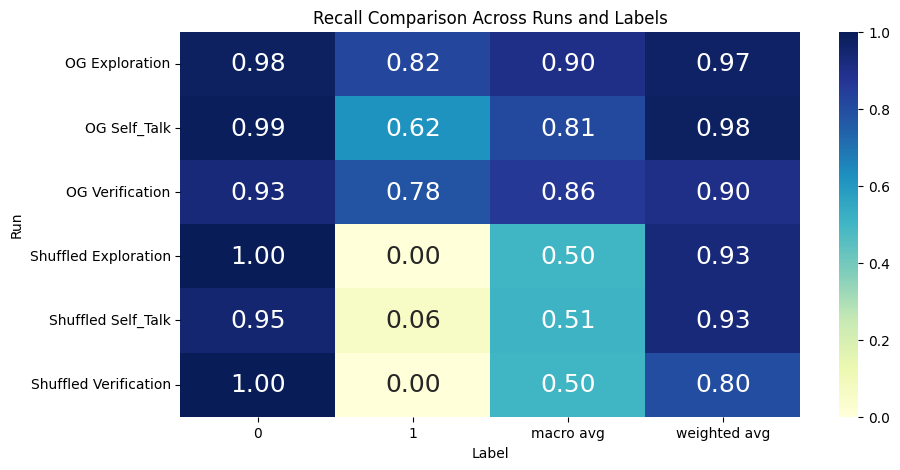}
        \caption{Recall}
        \label{fig:shuffled_recall}
    \end{subfigure}
    \vspace{-0.2cm}
    \caption{Precision and Recall - ReasonType vs. Shuffled labels}
    \label{fig:shuffled_metrics}
    \vspace{-0.2cm}
\end{figure}

Furthermore, Figures \ref{fig:shuffled_precision} and \ref{fig:shuffled_recall} show the Precision and Recall classification metrics on the testing dataset, respectively. For Original runs, both classes ($0$ and $1$) achieve good performance despite dataset imbalancity, with Macro average Precision and Recall lying between $0.76$ and $0.90$ across labels. In comparison, shuffled runs presents poor results, with models failing in predicting positive classes - Precision and Recall of class $1$ between $0.00$ and $0.06$. Along with the training loss, theses metrics highlight that the models trained on shuffles labels cannot learn meaningful relations between steps and labels. In comparison, Original labels (from the ReasonType taxonomy) resulted in satisfying model performance, and smooth training.

\textbf{Takeaway.} Overall, these results comforts our finding that the ReasonType taxonomy labels enable annotation methods to results in reasoning steps carrying semantic meaning.

\newpage

\subsection{Generalizability of the ReasonType taxonomy} \label{sec:appendix-generalization-taxonomy}

We proved in Appendix \ref{sec:appendix-reliability-annotation} that \texttt{GPT-4o-mini} is a reliable annotator model for our taxonomy and the task of tagging the steps. However, the taxonomy is generated from specific samples of specific models. We already proved to some extend that the Taxonomy is generalizable. Indeed, we trained our step-tagging classifier on reasoning steps obtained using DS-Qwen14B, a model that was not used to derive the taxonomy (see Appendix \ref{app:construction-taxonomy}). Moreover, we obtained satisfying performance of our step-tagging classifier on models and datasets that were not used to for training, namely DS-Llama8B and QwQ-32B on MATH500 and GSM8K, and DS-Qwen14B on GSM8K, AIME and MMLU-Pro (see Section \ref{sec:st-es-evaluation}).

\textbf{Claim.} To further validate our taxonomy, we will test its applicability to 2 additional reasoning models. Our claim is that if the annotation process from \texttt{GPT-4o-mini} results in accurate training of step-tagging classifiers, it means that the ReasonType taxonomy is applicable to other models, therefore generalizable. To evaluate this, we trained binary BERT classifiers -- similarly as Appendix \ref{sec:appendix-validation-taxonomy-shuffled} -- on four step-types, namely: \emph{Problem Re-statement}, \emph{Exploration}, \emph{Self-Talk} and \emph{Verification}.

\textbf{Methodology.} To verify our claim, we inferred 2 additional models on both MATH500 and GSM8K, specifically \texttt{microsoft/Phi-4-reasoning} and \texttt{Qwen/Qwen3-30B-A3B-Thinking-2507} since they are both reasoning models, and comes from additional providers or training methods. We then trained \textbf{$4$} BERT classifiers for each model and dataset. We used $500$ and $3,000$ samples from the MATH500 and GSM8K train datasets, respectively. 

\begin{figure}[h]
    \centering
    \begin{subfigure}{0.48\textwidth}
        \includegraphics[width=\linewidth]{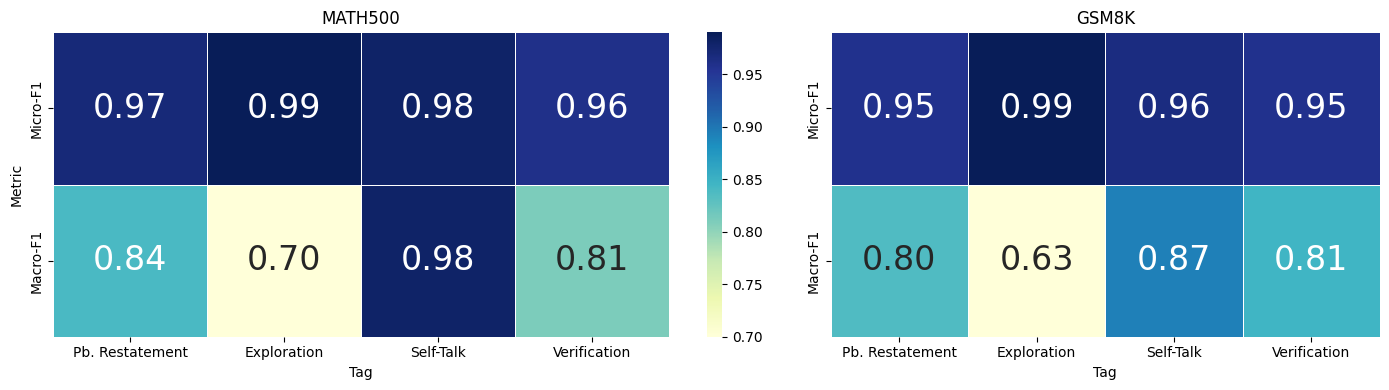}
        \caption{Phi-4}
        \label{fig:phi_4_reasoning}
    \end{subfigure}
    \hfill
    \begin{subfigure}{0.48\textwidth}
        \includegraphics[width=\linewidth]{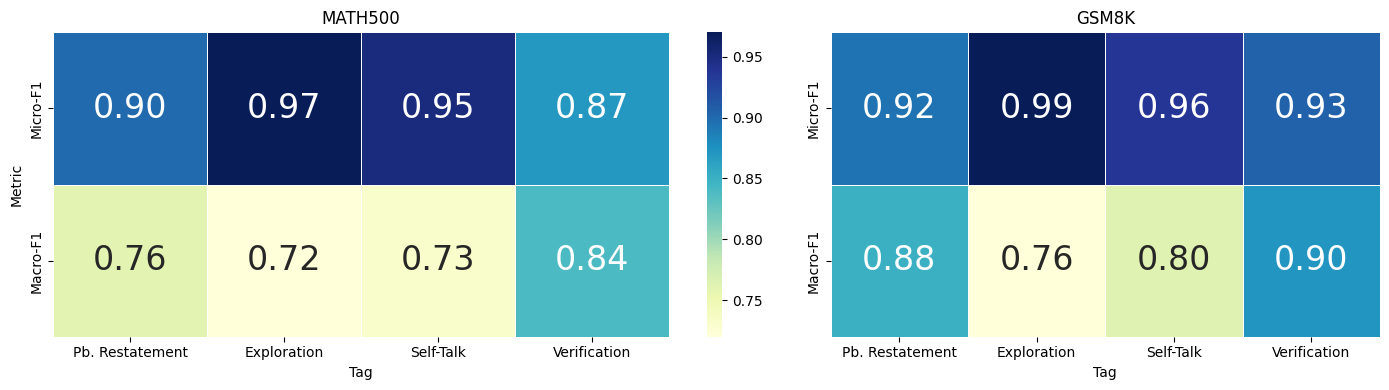}
        \caption{Qwen3-30B-A3B}
        \label{fig:Qwen3-30B-A3B}
    \end{subfigure}
    \vspace{-0.2cm}
    \caption{Performance of Step-Taggers - seed 42}
    \label{fig:performance_new_models}
    \vspace{-0.2cm}
\end{figure}

\textbf{Performances of the Step-Taggers.} We split the resulting annotated datasets following random 80:20 train/test split. Figure \ref{fig:phi_4_reasoning} and \ref{fig:Qwen3-30B-A3B} show the micro-F1 and macro-F1 for the Phi-4 and Qwen3-30B-A3B models on MATH500 and GSM8K, respectively. We observe satifying performances of Step-Taggers on both models. Specifically, on the \texttt{Phi-4} model, we obtained between $0.7$ to $0.98$ and $0.63$ to $0.87$ macro-F1 on MATH500 and GSM8K datasets, respectively. The lower macro-F1 on Exploration can be explained by its low representation in the datasets (around 1\% of labels in both datasets). 

Similarly, \texttt{Qwen3-30B-A3B} obtained satifying perfromance on both MATH500 and GSM8K, with $0.72$ to $0.84$ and $0.76$ to $0.90$ macro-F1, respectively. Performances of both models are comparable to results obtained on the DS-Qwen14B using the original dataset (see Appendix \ref{sec:appendix-validation-taxonomy-shuffled}).

\textbf{Takeaways.} We interpret the satisfying performances of the Step-Taggers trained on other models as further validating the applicability of our taxonomy to other models. Indeed, these models were not used to create our \emph{ReasonType} taxonomy, but still resulted in step-type identification performance of BERT classifiers trained on their reasoning traces using our methodology.

\newpage

\section{Baselines}

\subsection{Ideal Early-Stopping} \label{sec:appendix-ideal-early-stopping}

\textbf{Algorithm.} Algorithm \ref{alg:ies_algorithm} presents the Ideal Early-Stopping $\mathcal{IES}$ baseline algorithm.

\begin{algorithm}
\footnotesize
\caption{Ideal Early-Stopping $\mathcal{IES}$}
\label{alg:ies_algorithm}
\begin{algorithmic}[1]
\Require Prompt $x$; reasoning delimiter $\alpha \in V$; Ground-Truth Answer $y_{\text{gold}}$; Answer checker $\beta(\text{prediction}, \text{ground\_truth})$; Reasoning Language Model $\mathcal{M}$; tokenizer $\mathcal{T}$; EOS token $\gamma$; 

\State $y \gets \mathcal{T}(x)$ \color{gray} \Comment{Tokenize the input} \color{black}
\State $S_{running} \gets [\,]$; \color{gray} \Comment{Initialize output} \color{black}
\State $t \gets 0$
\State $b \gets True$ \color{gray} \Comment{Initialize stopping criteria} \color{black}

\While{$b$} \color{gray} \Comment{Generate until constraint breaks} \color{black}
    \State Generate step $s_i$ using $\mathcal{M}$ and $\alpha$
    \State $y \gets s_i$
    \State $y_{\text{current answer}} \gets \mathcal{P}_{\text{exit}}(y)$ \color{gray} \Comment{Retrieve the current best answer} \color{black}

    \If{$\beta(y_{\text{current answer}}, y_{\text{gold}})$}
        \State $b \gets False$ \color{gray} \Comment{Stop generation} \color{black}
        
    \Else
        \State Continue the generation until EOS token
    \EndIf

    \State $t \gets t + 1$
\EndWhile

\State \Return y

\end{algorithmic}
\end{algorithm}

\begin{figure}[h]
    \centering
    \begin{subfigure}{0.32\linewidth}
        \centering
        \includegraphics[width=\linewidth]{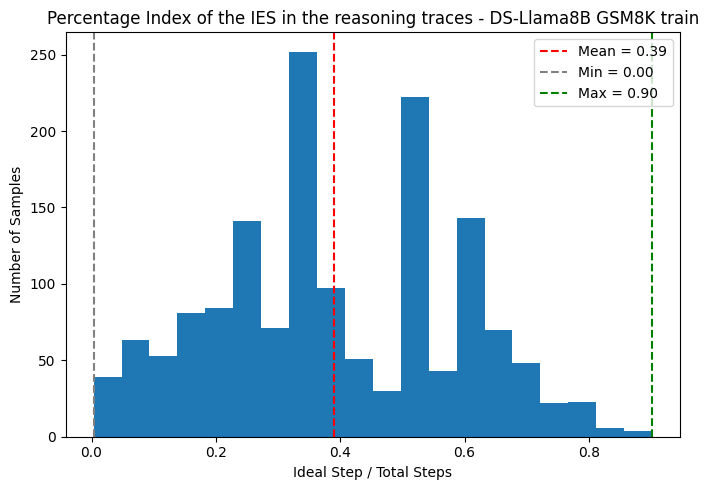}
        \caption{DS-Llama8B on GSM8K}
        \label{fig:DS8B-GSM8K-IES} 
    \end{subfigure}
    \hfill
    \begin{subfigure}{0.32\linewidth}
        \centering
        \includegraphics[width=\linewidth]{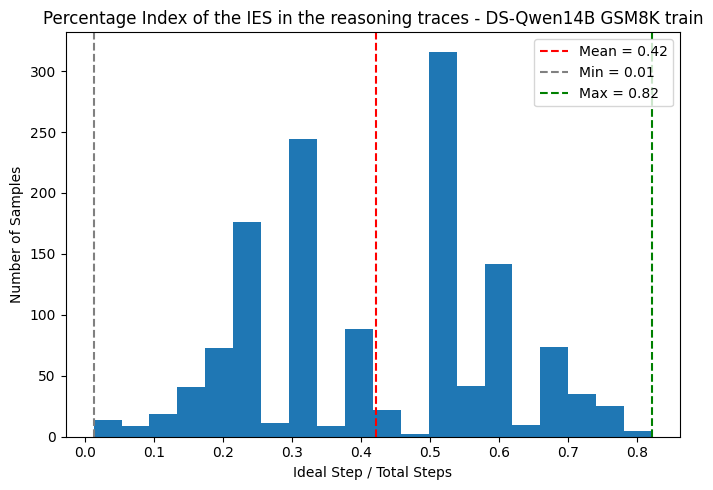}
        \caption{DS-Qwen14B on GSM8K}
        \label{fig:DS14B-GSM8K-IES}
    \end{subfigure}
    \hfill
    \begin{subfigure}{0.32\linewidth}
        \centering
        \includegraphics[width=\linewidth]{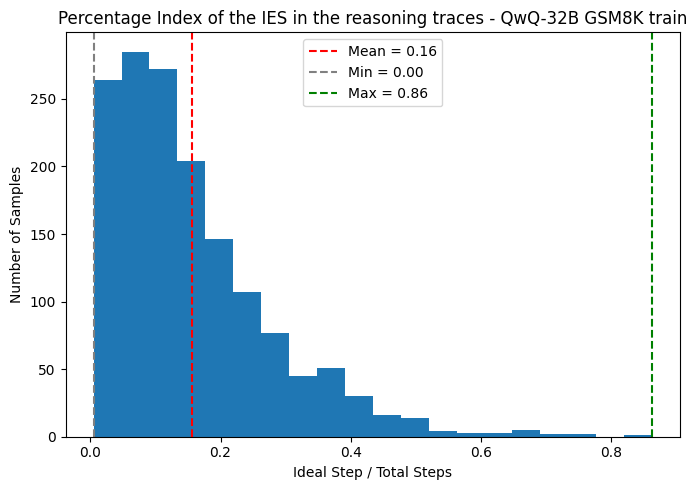}
        \caption{QwQ-32B on GSM8K}
        \label{fig:QwQ32B-GSM8K-IES}
    \end{subfigure}
    \begin{subfigure}{0.32\linewidth}
        \centering
        \includegraphics[width=\linewidth]{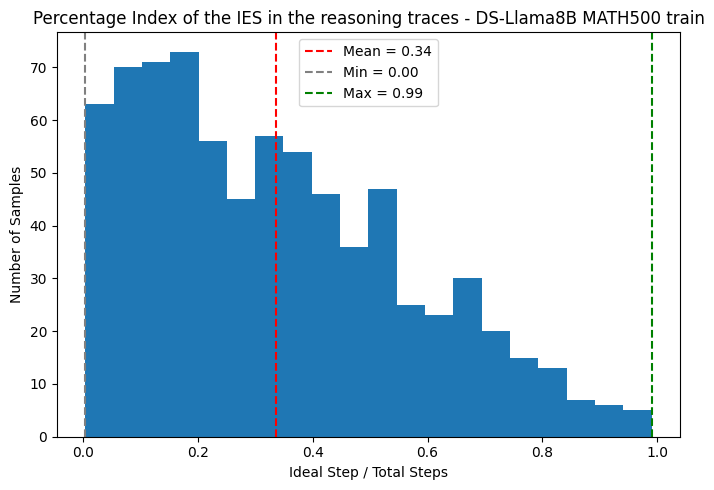}
        \caption{DS-Llama8B on MATH500}
        \label{fig:DS8B-MATH500-IES} 
    \end{subfigure}
    \hfill
    \begin{subfigure}{0.32\linewidth}
        \centering
        \includegraphics[width=\linewidth]{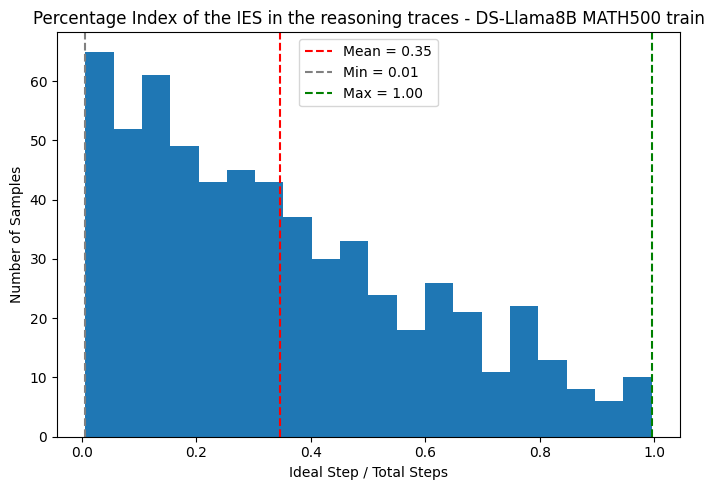}
        \caption{DS-Qwen14B on MATH500}
        \label{fig:DS14B-MATH500-IES}
    \end{subfigure}
    \hfill
    \begin{subfigure}{0.32\linewidth}
        \centering
        \includegraphics[width=\linewidth]{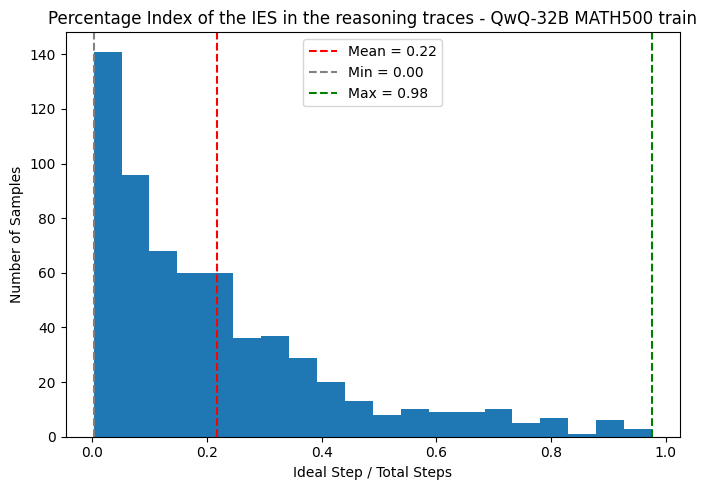}
        \caption{QwQ-32B on MATH500}
        \label{fig:QwQ32B-MATH500-IES}
    \end{subfigure}
    \\
    \begin{subfigure}{0.31\linewidth}
        \centering
        \includegraphics[width=\linewidth]{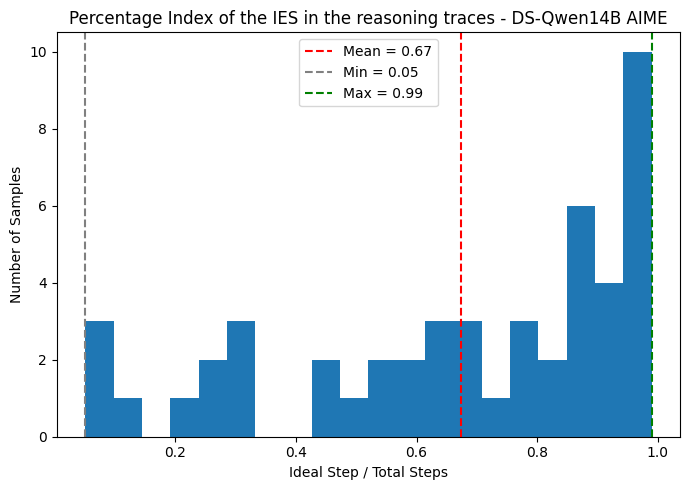}
        \caption{DS-Qwen14B on AIME}
        \label{fig:DS14B-AIME-IES} 
    \end{subfigure}
    \hfill
    \begin{subfigure}{0.31\linewidth}
        \centering
        \includegraphics[width=\linewidth]{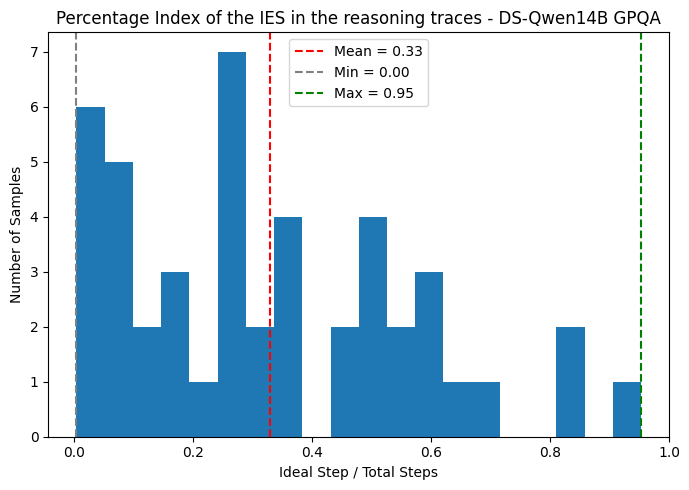}
        \caption{DS-Qwen14B on GPQA}
        \label{fig:DS14B-GPQA-IES}
    \end{subfigure}
    \hfill
    \begin{subfigure}{0.31\linewidth}
        \centering
        \includegraphics[width=\linewidth]{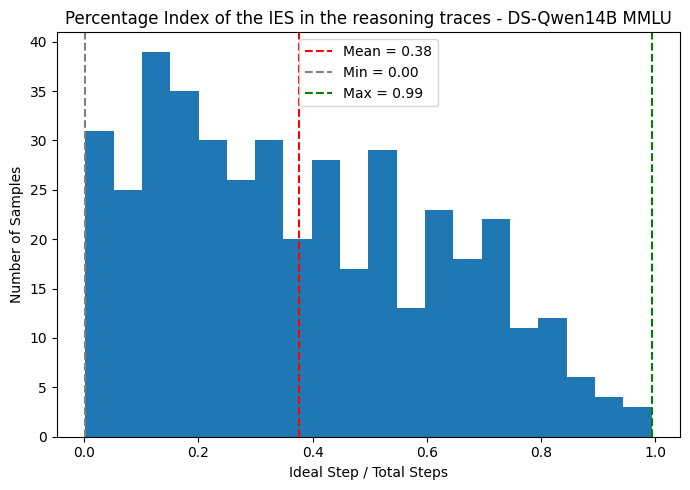}
        \caption{DS-Qwen14B on MMLU-Pro}
        \label{fig:DS14B-MMLU-IES}
    \end{subfigure}
    \vspace{-0.2cm}
    \caption{IES on training datasets - correct answers}
    \label{fig:IES-train}
    \vspace{-0.3cm}
\end{figure}

\textbf{$S_{IES}$ in practice.} Figure \ref{fig:IES-train} shows the distribution of the IES step $S_{IES}$ obtained by applying Algorithm \ref{alg:ies_algorithm} to the traces obtained on the training datasets.

\textbf{Further evidences of reasoning transition phase.} In Section \ref{sec:experimental-section}, we detail that our IES algorithm. In practice, we prompted back the model after each steps to retrieve its current best answer. While this process is expansive (models often generate hundreds of steps for complex problems), it gives us access the model's best answer after each step that is being generated. 

In Section \ref{sec:influence-step-type}, we observe that our ratio $R$ drops around the $S_{IES}$ step. By the nature of the classes $\tau_{\text{constructive}}$ and $\tau_{\text{evaluative}}$, we interpreted the drop of $R$ as a reasoning phase transition. We concluded that before that phase, the model constructed its answer, and past this phase, the model verified its current answer.

\begin{figure}[h]
    \centering
    \begin{subfigure}{0.32\linewidth}
        \centering
        \includegraphics[width=\linewidth]{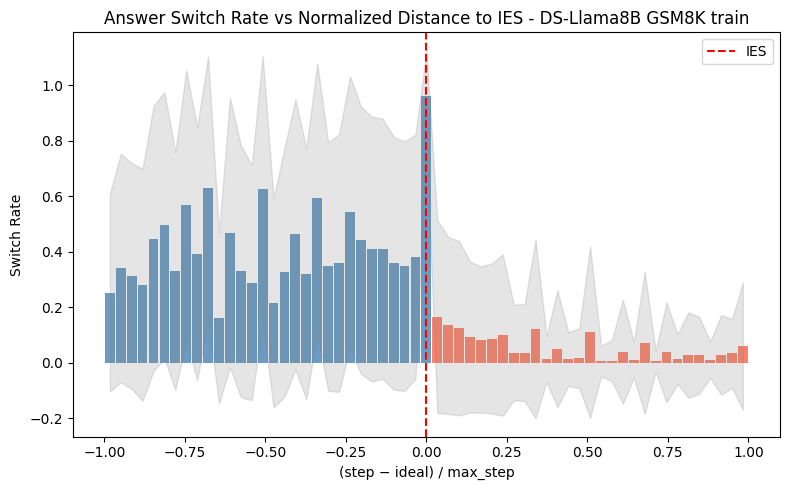}
        \caption{DS-Llama8B on GSM8K}
        \label{fig:DS8B-GSM8K-answer-switch} 
    \end{subfigure}
    \hfill
    \begin{subfigure}{0.32\linewidth}
        \centering
        \includegraphics[width=\linewidth]{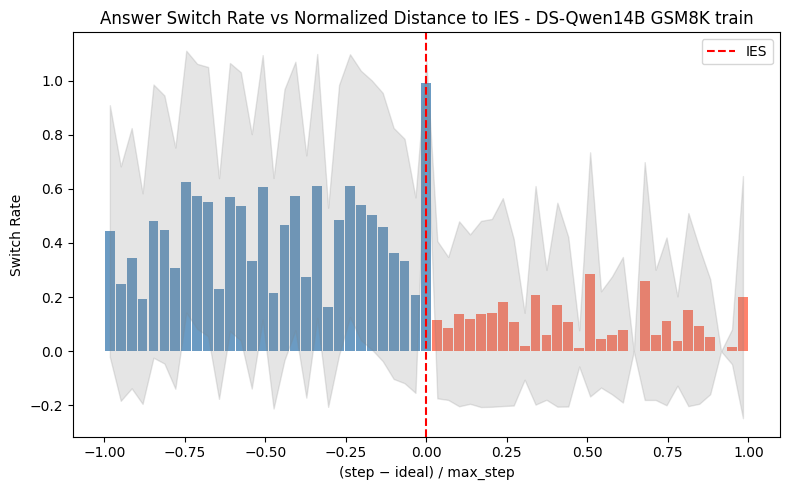}
        \caption{DS-Qwen14B on GSM8K}
        \label{fig:DS14B-GSM8K-answer-switch}
    \end{subfigure}
    \hfill
    \begin{subfigure}{0.32\linewidth}
        \centering
        \includegraphics[width=\linewidth]{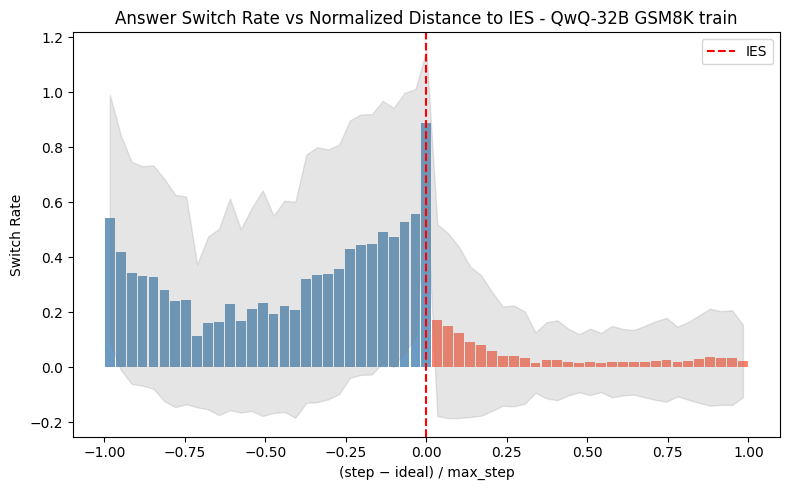}
        \caption{QwQ-32B on GSM8K}
        \label{fig:QwQ32B-GSM8K-answer-switch}
    \end{subfigure}
    \begin{subfigure}{0.32\linewidth}
        \centering
        \includegraphics[width=\linewidth]{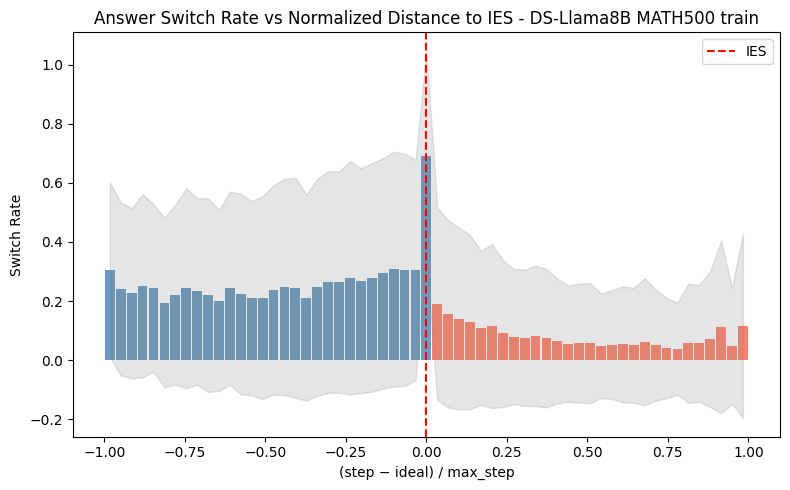}
        \caption{DS-Llama8B on MATH500}
        \label{fig:DS8B-MATH500-answer-switch} 
    \end{subfigure}
    \hfill
    \begin{subfigure}{0.32\linewidth}
        \centering
        \includegraphics[width=\linewidth]{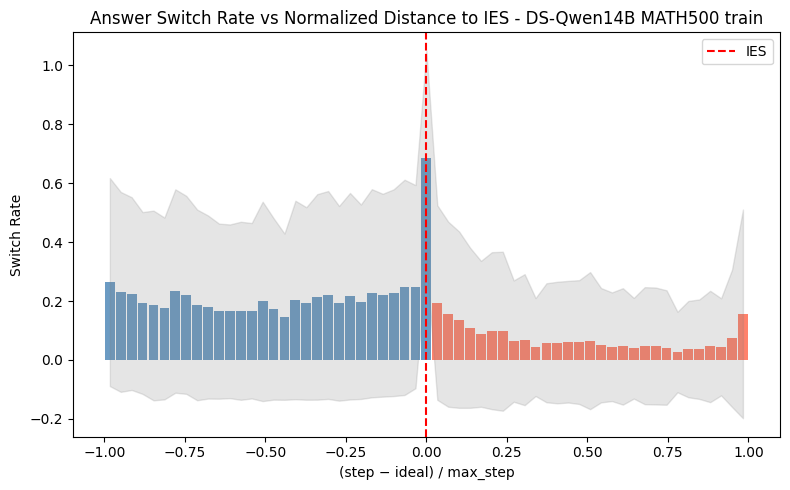}
        \caption{DS-Qwen14B on MATH500}
        \label{fig:DS14B-MATH500-answer-switch}
    \end{subfigure}
    \hfill
    \begin{subfigure}{0.32\linewidth}
        \centering
        \includegraphics[width=\linewidth]{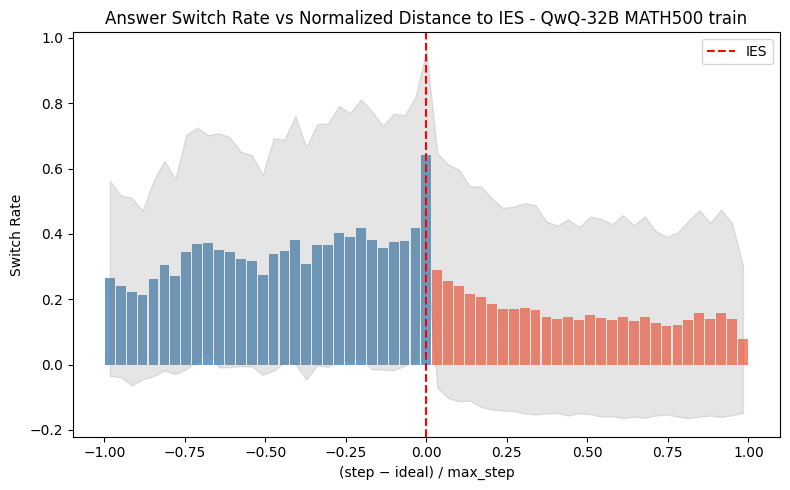}
        \caption{QwQ-32B on MATH500}
        \label{fig:QwQ32B-MATH500-answer-switch}
    \end{subfigure}
    \vspace{-0.2cm}
    \caption{Switch rate of model's answer on training datasets - $s_i \in S_{\text{before}}$ and $s_i \in S_{\text{after}}$ in blue and red bars, respectively}
    \label{fig:answer-switch-rate}
    \vspace{-0.3cm}
\end{figure}

While it is an observation of the model's behaviors (type of reasoning), we miss an observation of the model's state with regards to its current answer. Figure \ref{fig:answer-switch-rate} address this, and show how frequently the model's current best answer changes through the reasoning. To plot such figures, we inspect the content in the "\textbackslash boxed\{\dots\}" answer between consecutive steps. Indeed, we prompted back the model forcing the model to include a "\textbackslash boxed\{\dots\}" in its answer generation for each step. If the content is different between two consecutive steps, we interpret that the model switched its current best answer. 

We define the switch rate as the proportion of sample in the dataset including a switch between consecutive steps at a particular step index. We observe that before the IES step, the model switches its current answer frequently. After the IES step, we observe that the model stabilizes, and conserves its current answer.

\newpage

\begin{figure}[h]
    \centering
    \begin{subfigure}{0.32\linewidth}
        \centering
        \includegraphics[width=\linewidth]{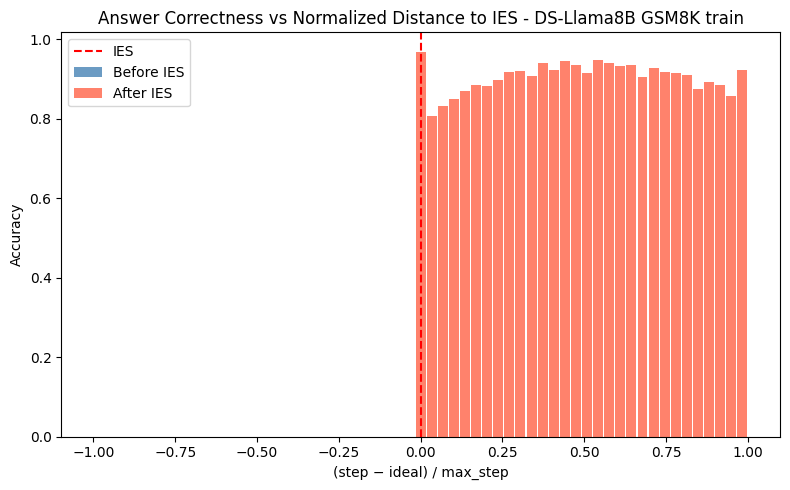}
        \caption{DS-Llama8B on GSM8K}
        \label{fig:DS8B-GSM8K-answer-correct} 
    \end{subfigure}
    \hfill
    \begin{subfigure}{0.32\linewidth}
        \centering
        \includegraphics[width=\linewidth]{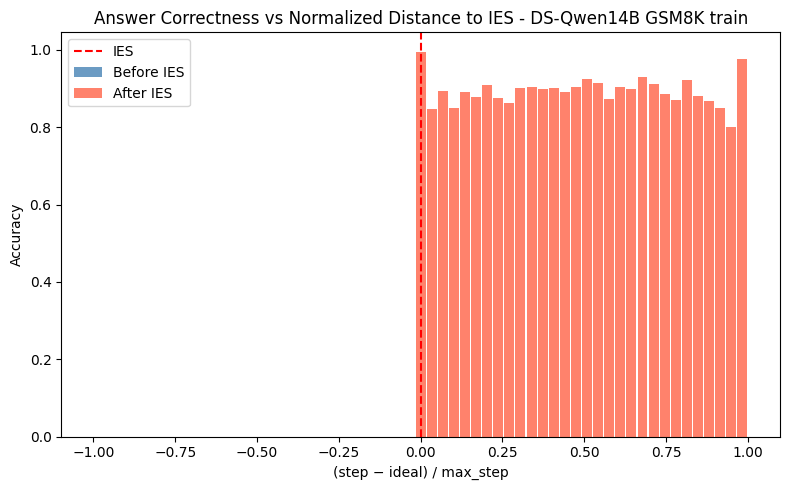}
        \caption{DS-Qwen14B on GSM8K}
        \label{fig:DS14B-GSM8K-answer-correct}
    \end{subfigure}
    \hfill
    \begin{subfigure}{0.32\linewidth}
        \centering
        \includegraphics[width=\linewidth]{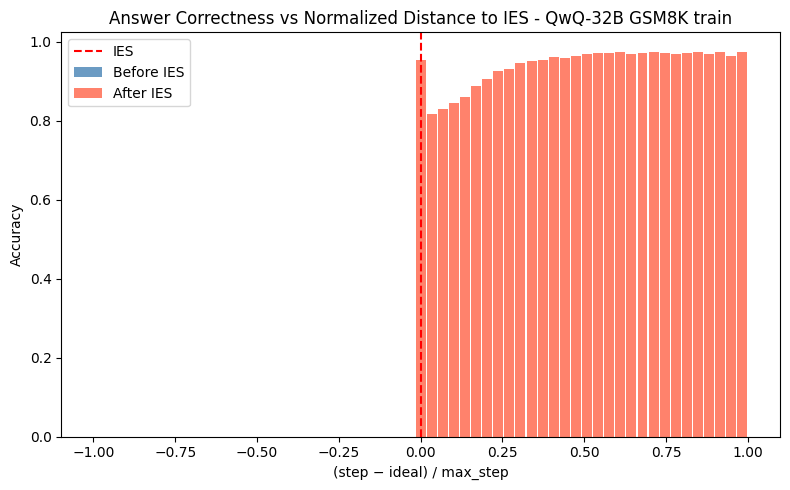}
        \caption{QwQ-32B on GSM8K}
        \label{fig:QwQ32B-GSM8K-answer-correct}
    \end{subfigure}
    \begin{subfigure}{0.32\linewidth}
        \centering
        \includegraphics[width=\linewidth]{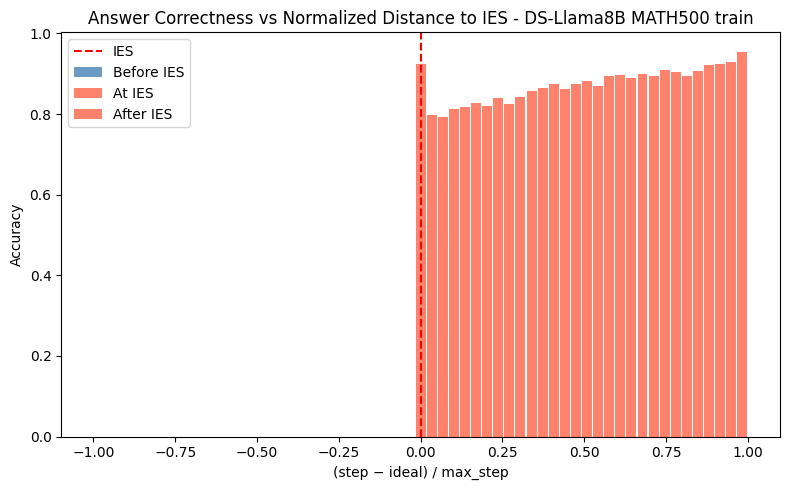}
        \caption{DS-Llama8B on MATH500}
        \label{fig:DS8B-MATH500-answer-correct} 
    \end{subfigure}
    \hfill
    \begin{subfigure}{0.32\linewidth}
        \centering
        \includegraphics[width=\linewidth]{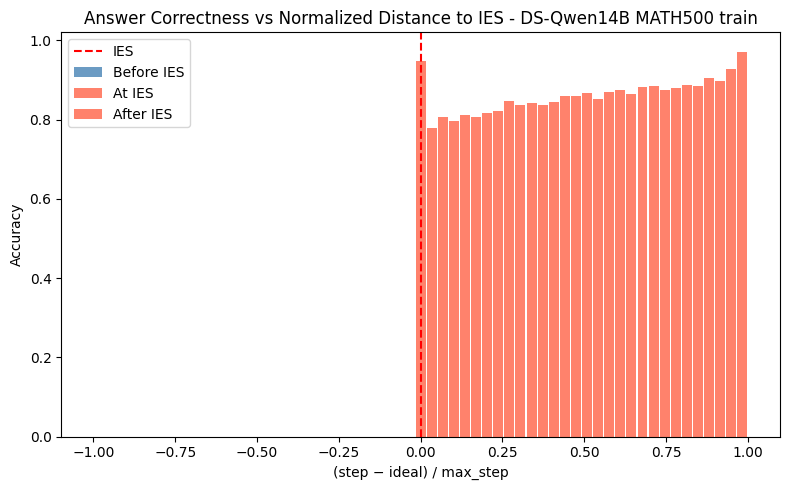}
        \caption{DS-Qwen14B on MATH500}
        \label{fig:DS14B-MATH500-answer-correct}
    \end{subfigure}
    \hfill
    \begin{subfigure}{0.32\linewidth}
        \centering
        \includegraphics[width=\linewidth]{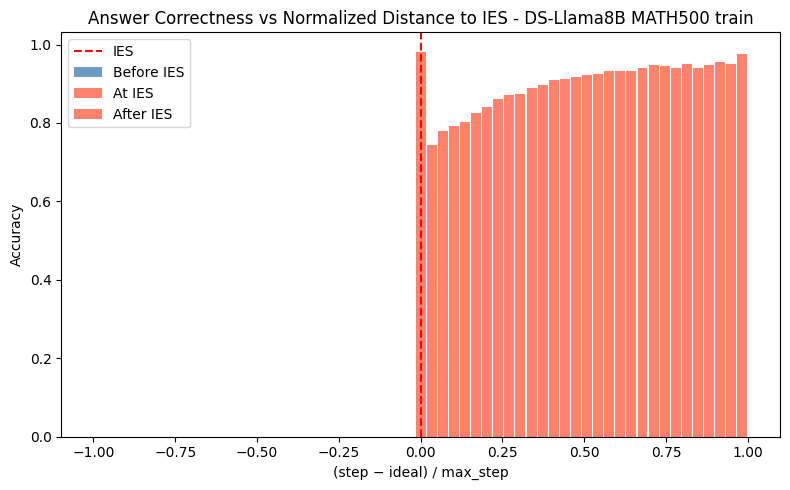}
        \caption{QwQ-32B on MATH500}
        \label{fig:QwQ32B-MATH500-answer-correct}
    \end{subfigure}
    \vspace{-0.2cm}
    \caption{Answer correctness on training datasets} 
    \label{fig:answer-correctness-rate}
    \vspace{-0.3cm}
\end{figure}

\textbf{Answer correctness before and after IES step.} Furthermore, we want to analyze how the correctness propagates through the reasoning. Such information could be useful to better understand how to manage the reasoning process and when to early-stop. Figure \ref{fig:answer-correctness-rate} shows the correctness of the model through its reasoning. Instead of monitoring how frequent the model switches its answer, we evaluated the "\textbackslash boxed\{\dots\}" generated, at each step, with the ground-truth. 

On Figure \ref{fig:answer-correctness-rate}, we observe an accuracy of 0 before the IES step ($x=0$). This is expected, since the IES step is the first time where the model reaches a correct answer. Furthermore, the correctness peak at the IES step, and then drops. This can be attributed to the fact that after the IES, the model still switches its answer.

Figure \ref{fig:answer-switch-rate} and \ref{fig:answer-correctness-rate} partially explains why early-stopping is a hard problem. After the IES step, the model switches its answer less frequently, but still does. It means that sometimes the model reached the correct answer (IES), but move away from it time to time (and it seems to be caused by the self-verification/validation of the model -- see Section \ref{sec:influence-step-type}, where we concluded that the reasoning behavior of LRMs is often dominated by \emph{Evaluative} steps). So if a sample early-stop after the IES step and before the end of the CoT, the TRACES algorithm sometimes early-stops with the wrong answer (explaining why we have minimal drop in the accuracy for moderated values of $\delta$).

\newpage

\subsection{Prompt-guided budget compression} \label{sec:appendix-prompt-guided-baselines}

\begin{figure}[h]
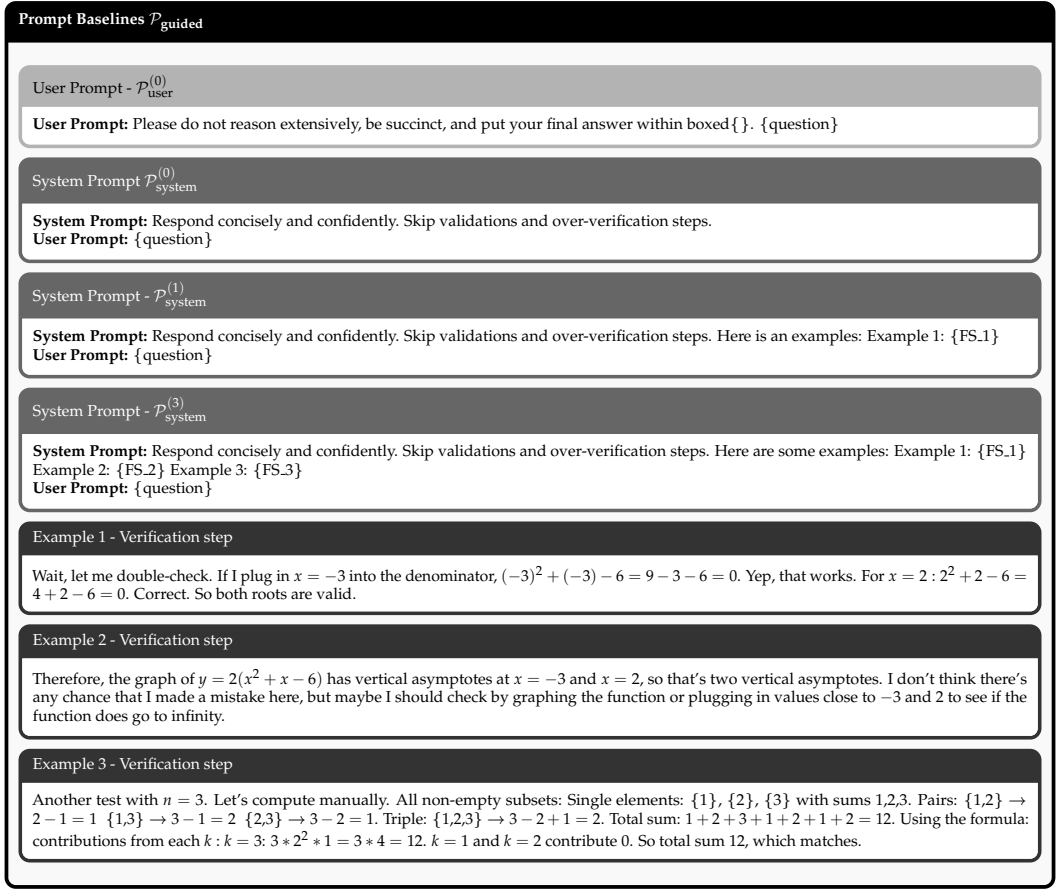

\centering
\tiny
\begin{adjustbox}{max width=\textwidth}
\begin{tcolorbox}[colback=gray!5, colframe=black, title=Prompt Baselines $\mathcal{P}_{\text{guided}}$, fonttitle=\bfseries, left=1pt, right=1pt]

\begin{tcolorbox}[
  colback=white, 
  colframe=black!30, 
  title=User Prompt - $\mathcal{P}^{(0)}_{\text{user}}$, 
  coltitle=black!30!black,
  top=2pt,
  bottom=2pt,
  boxsep=2pt,
  width=\textwidth,
  left=2pt, 
  right=2pt
]
\textbf{User Prompt:} Please do not reason extensively, be succinct, and put your final answer within boxed\{\}. \{question\}
\end{tcolorbox}


\begin{tcolorbox}[colback=white, 
    colback=white, 
    colframe=black!60, 
    title=System Prompt $\mathcal{P}^{(0)}_{\text{system}}$,
    top=2pt,
  bottom=2pt,
  boxsep=2pt,
  width=\textwidth,
  left=2pt, 
  right=2pt
]
\textbf{System Prompt:} Respond concisely and confidently. Skip validations and over-verification steps.

\textbf{User Prompt:} \{question\}

\end{tcolorbox}

\begin{tcolorbox}[colback=white, 
    colback=white, 
    colframe=black!60, 
    title=System Prompt - $\mathcal{P}^{(1)}_{\text{system}}$,
    top=2pt,
  bottom=2pt,
  boxsep=2pt,
  width=\textwidth,
  left=2pt, 
  right=2pt
]
\textbf{System Prompt:} Respond concisely and confidently. Skip validations and over-verification steps. Here is an examples: Example 1: \{FS\_1\}

\textbf{User Prompt:} \{question\}

\end{tcolorbox}

\begin{tcolorbox}[colback=white, 
    colback=white, 
    colframe=black!60, 
    title=System Prompt - $\mathcal{P}^{(3)}_{\text{system}}$,
    top=2pt,
  bottom=2pt,
  boxsep=2pt,
  width=\textwidth,
  left=2pt, 
  right=2pt
]
\textbf{System Prompt:} Respond concisely and confidently. Skip validations and over-verification steps. Here are some examples:  Example 1: \{FS\_1\}  Example 2: \{FS\_2\}  Example 3: \{FS\_3\}

\textbf{User Prompt:} \{question\}

\end{tcolorbox}

\begin{tcolorbox}[colback=white, 
    colback=white, 
    colframe=black!80, 
    title=Example 1 - Verification step,
    top=2pt,
  bottom=2pt,
  boxsep=2pt,
  width=\textwidth,
  left=2pt, 
  right=2pt
]

Wait, let me double-check. If I plug in $x = -3$ into the denominator, $(-3)^2 + (-3) -6 = 9 -3 -6 = 0$. Yep, that works. For $x =2: 2^2 +2 -6 =4 +2 -6=0$. Correct. So both roots are valid. 

\end{tcolorbox}

\begin{tcolorbox}[colback=white, 
    colback=white, 
    colframe=black!80, 
    title=Example 2 - Verification step,
    top=2pt,
  bottom=2pt,
  boxsep=2pt,
  width=\textwidth,
  left=2pt, 
  right=2pt
]

Therefore, the graph of $y=2\/(x^2 +x -6)$ has vertical asymptotes at $x= -3$ and $x=2$, so that's two vertical asymptotes. I don't think there's any chance that I made a mistake here, but maybe I should check by graphing the function or plugging in values close to $-3$ and $2$ to see if the function does go to infinity.

\end{tcolorbox}

\begin{tcolorbox}[colback=white, 
    colback=white, 
    colframe=black!80, 
    title=Example 3 - Verification step,
    top=2pt,
  bottom=2pt,
  boxsep=2pt,
  width=\textwidth,
  left=2pt, 
  right=2pt
]

Another test with $n=3$. Let's compute manually. All non-empty subsets: Single elements: \{1\}, \{2\}, \{3\} with sums 1,2,3. Pairs: \{1,2\} → $2-1=1$\; \{1,3\} → $3-1=2$\; \{2,3\} → $3-2=1$. Triple: \{1,2,3\} → $3 -2 +1=2$. Total sum: $1+2+3 +1+2+1 +2 = 12$. Using the formula: contributions from each $k:$ $k=3$: $3*2^2*1= 3*4=12$. $k=1$ and $k=2$ contribute $0$. So total sum $12$, which matches.

\end{tcolorbox}

\end{tcolorbox}
\end{adjustbox}
\caption{Prompt baselines} 
\label{fig:prompt_baselines}
\end{figure}

\newpage

\subsection{Token-count budget compression} \label{sec:appendix-token-count-baselines}

\textbf{Motivation.} In Section \ref{sec:related-work}, we framed our contributions and within existing gaps in the literature. Specifically, a motivation of our work is that existing early-stopping methods are either fixed (the early-stopping is fixed before the inference, using prompt-compression or token-count budgets) or dynamic (the stopping decision is taken during the inference, based on the model's internals). 

First, most existing dynamic methods are gray/white box. Indeed, \citet{yang2025dynamicearlyexitreasoning,wang2025entropylangletextttthinkrangle} rely on the logits or on the model's internals which reduces the applicability of early-stopping mechanisms. In comparison, our approach is black-box since it only relies the text generated by the CoT (which is the case for certain APIs such as Ollama, or closed-source models - those models does not gives the user access to the model's internal). 

Second, in Section \ref{sec:experimental-section}, we introduced the $\mathcal{P_{\text{guided}}}$ baselines, inspired by the related work \citep{lee2025llmscompresschainofthoughttoken}. While these baselines enable a fair comparison (black-box early-stopping), it does not allow a  comparison at equivalent token-count. For this reason, this ablation further evaluate our TRACES framework in comparison to $\mathcal{T_{\text{token-count}}}$ token-count baselines \citep{pu2025thoughtterminatorbenchmarkingcalibratingmitigating,han2025tokenbudgetawarellmreasoning}. Indeed, $\mathcal{T_{\text{token-count}}}$ baselines allows us to target a specific token-count when early-stopping the generation.

\textbf{Claim.} In this section, we argue that TRACES is competitive against $\mathcal{T_{\text{token-count}}}$ baselines. Furthermore, we show that it also offers as much control over the generation as managing the token-count, and does not requires previous knowledge for a given model or dataset. In other words, TRACES is more agnostic to the models and datasets used than the baselines.

\textbf{Methodology.} To implement $\mathcal{T_{\text{token-count}}}$, we stop the generation of the LRMs early, based on fixed token-budget. The token-count of LRMs depends on the model and datasets. For this reason, to build effective $\mathcal{T_{\text{token-count}}}$ methods, the literature showed that a calibration analysis is required before-hand \citep{pu2025thoughtterminatorbenchmarkingcalibratingmitigating,han2025tokenbudgetawarellmreasoning}. To calibrate our token-count baseline effectively, we assume to have access to the average token-count over samples for a given model and dataset. 

\begin{figure}[h]
    \centering
    \begin{subfigure}{0.3\linewidth}
        \centering
        \includegraphics[width=\linewidth]{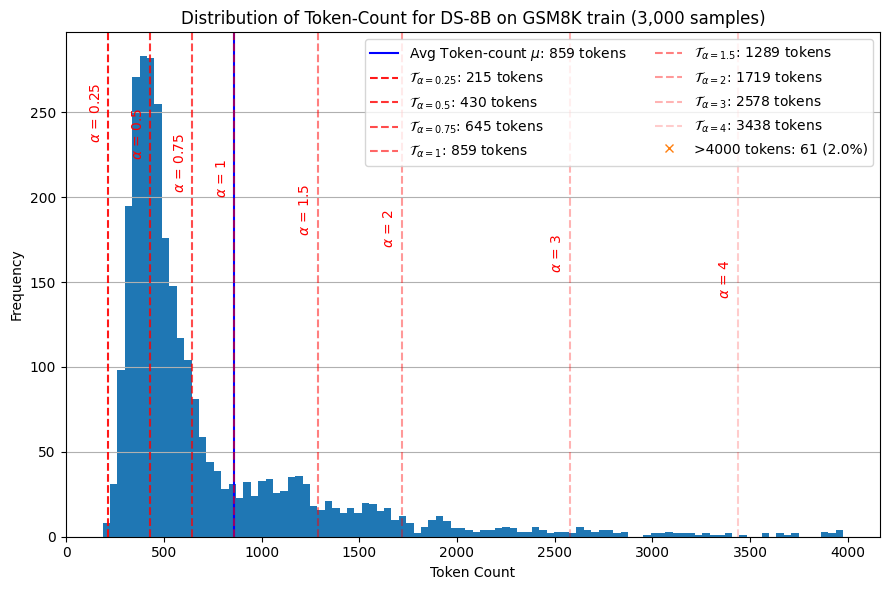}
        \caption{GSM8K - DS-8B}
        \label{fig:token_count_ds8b_gsm8k}
    \end{subfigure}
    \hfill
    \begin{subfigure}{0.3\linewidth}
        \centering
        \includegraphics[width=\linewidth]{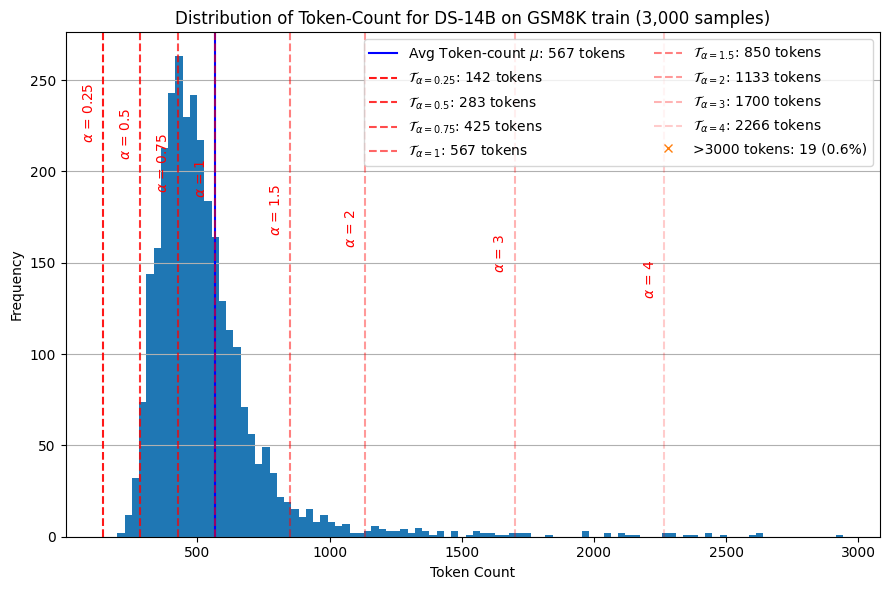}
        \caption{GSM8K - DS-14B}
        \label{fig:token_count_ds14b_gsm8k}
    \end{subfigure}
    \hfill
    \begin{subfigure}{0.3\linewidth}
        \centering
        \includegraphics[width=\linewidth]{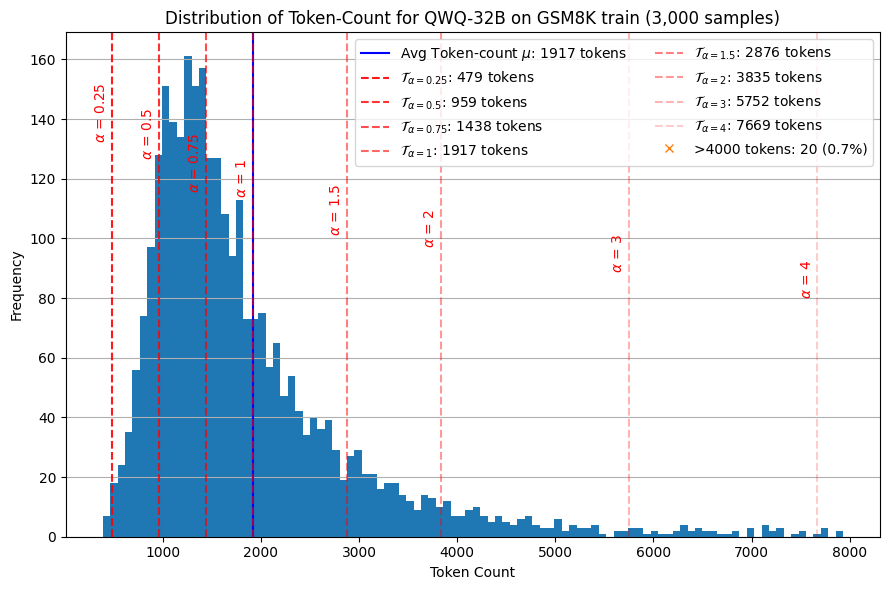}
        \caption{GSM8K - QwQ-32B}
        \label{fig:token_count_qwq32b_gsm8k}
    \end{subfigure}
    
    \begin{subfigure}{0.3\linewidth}
        \centering
        \includegraphics[width=\linewidth]{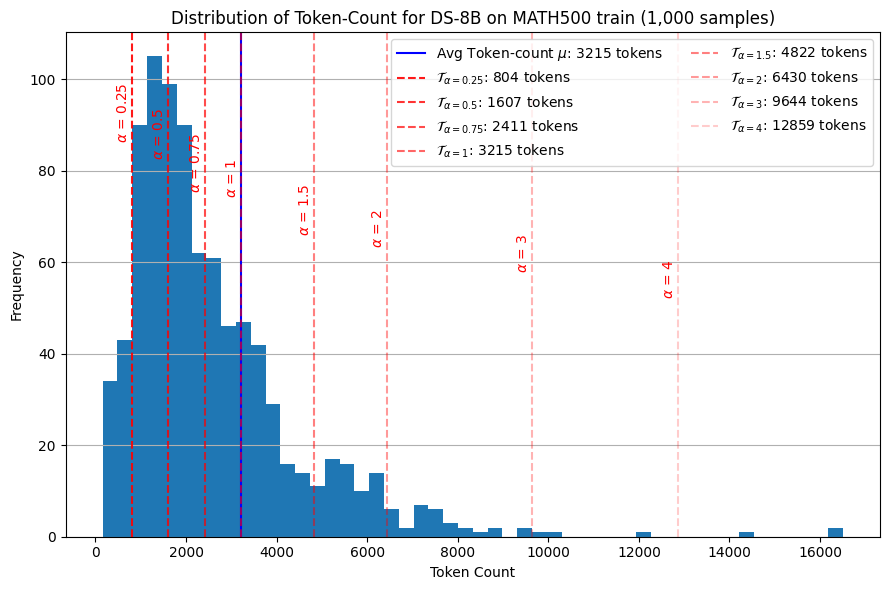}
        \caption{MATH500 - DS-8B}
        \label{fig:token_count_ds8b_math500}
    \end{subfigure}
    \hfill
    \begin{subfigure}{0.3\linewidth}
        \centering
        \includegraphics[width=\linewidth]{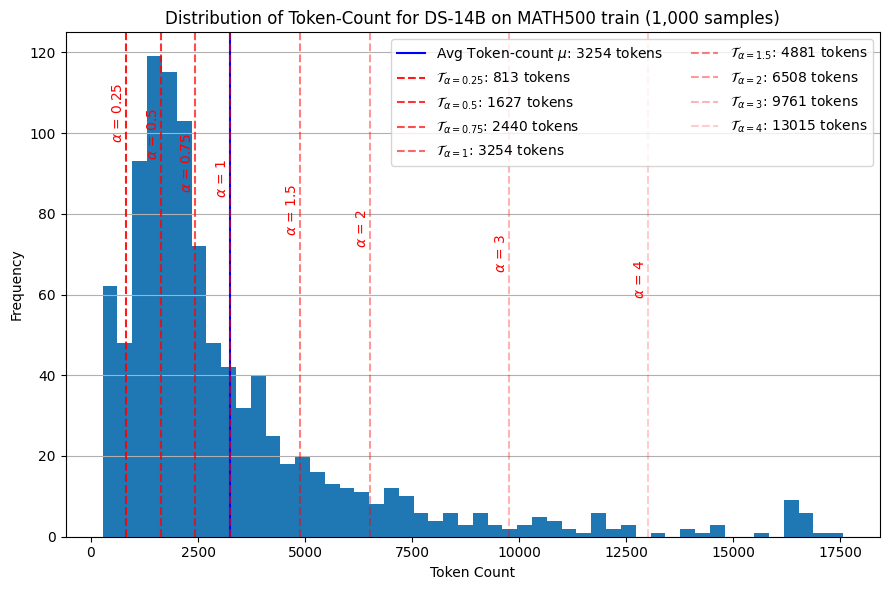}
        \caption{MATH500 - DS-14B}
        \label{fig:token_count_ds14b_math500}
    \end{subfigure}
    \hfill
    \begin{subfigure}{0.3\linewidth}
        \centering
        \includegraphics[width=\linewidth]{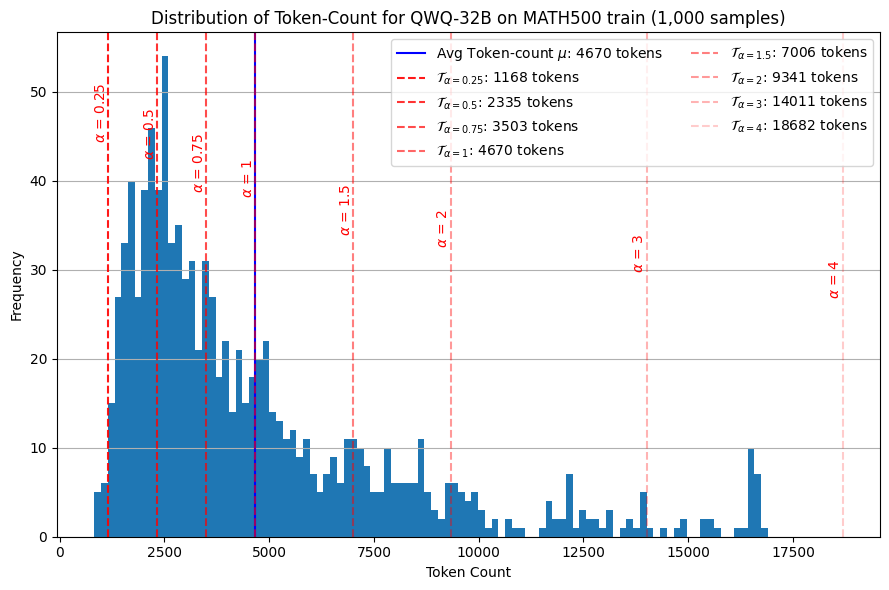}
        \caption{MATH500 - QwQ-32B}
        \label{fig:token_count_qwq32b_math500}
    \end{subfigure}

    \begin{subfigure}{0.3\linewidth}
        \centering
        \includegraphics[width=\linewidth]{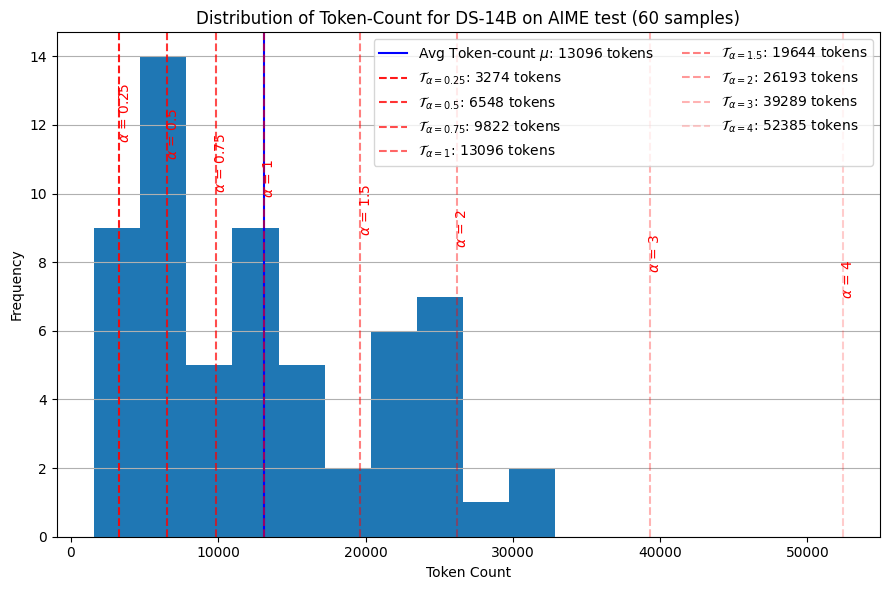}
        \caption{AIME - DS-14B}
        \label{fig:token_count_ds14b_aime}
    \end{subfigure}
    \hfill
    \begin{subfigure}{0.3\linewidth}
        \centering
        \includegraphics[width=\linewidth]{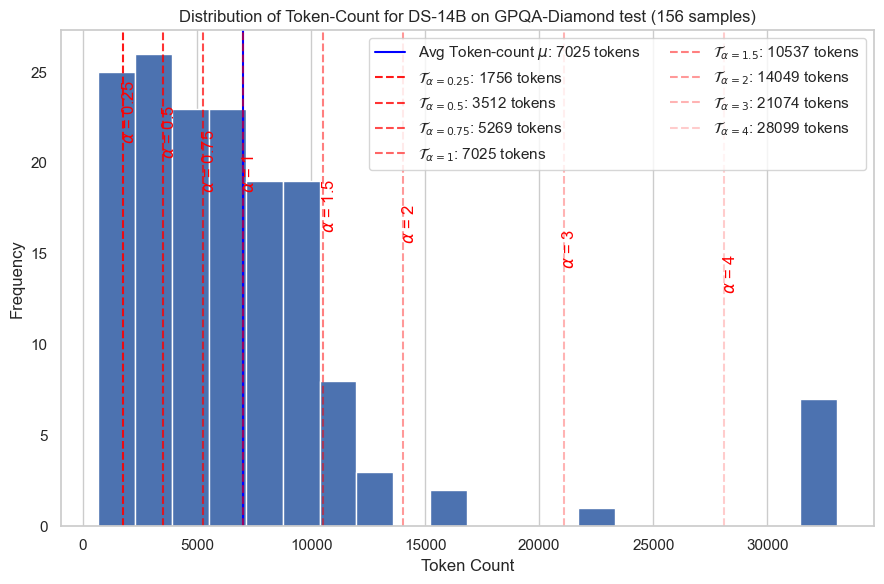}
        \caption{GPQA - DS-14B}
        \label{fig:token_count_ds14b_gpqa}
    \end{subfigure}
    \hfill
    \begin{subfigure}{0.3\linewidth}
        \centering
        \includegraphics[width=\linewidth]{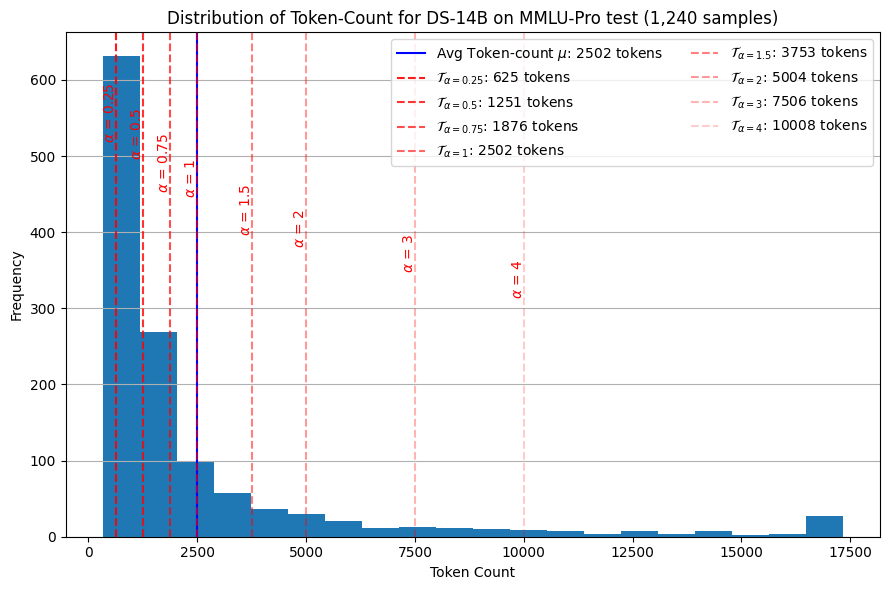}
        \caption{MMLU - DS-14B}
        \label{fig:token_count_ds14b_mmlu}
    \end{subfigure}
    \vspace{-0.2cm}
    \caption{Distribution of Token-count on training datasets - \deepseekLlama{}.} 
    \label{fig:distribution-token-train-data}
    \vspace{-0.3cm}
\end{figure}

\textbf{Selecting the token-budget.} The average token-count budget is useful to know the computation efforts needed by a model for a given dataset (without intervention in the generation). However, the token-count is volatile across the samples, and depends on both models and datasets. Figure \ref{fig:distribution-token-train-data} shows the distribution of token-count on training datasets. We observe that the token-count is variable across samples of a same dataset. To make our $\mathcal{T_{\text{token-count}}}$ baseline simple yet effective, we suggest to fix the token-budget proportionally to the average token-count of models and datasets combination. Namely, we stop the generation early if the token-count exceed $\eta = \alpha \times \mu_{\text{model, dataset}}$, where $\eta$ is the token-count limit, $\mu_{\text{model, dataset}}$ the average token-count for a given model and dataset, and $\alpha$ is a factor that varies. For our experimentation, we select $\alpha \in [0.25, 0.5, 0.75, 1.0, 1.5, 2.0, 3.0, 4.0]$. These various values allows to set budgets around the value of the average token-count. In Figure \ref{fig:distribution-token-train-data}, the red vertical dotted lines represents the different $\mathcal{T_{\text{token-count}}}$ that we selected.

\begin{figure}[h]
    \centering
    \begin{subfigure}{0.3\linewidth}
        \centering
        \includegraphics[width=\linewidth]{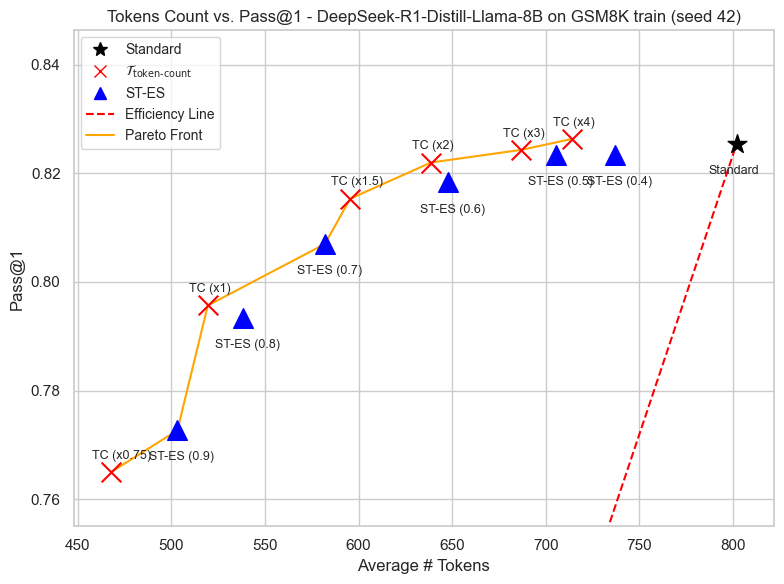}
        \caption{GSM8K - DS-8B}
        \label{fig:token_count_base_ds8b_gsm8k}
    \end{subfigure}
    \hfill
    \begin{subfigure}{0.3\linewidth}
        \centering
        \includegraphics[width=\linewidth]{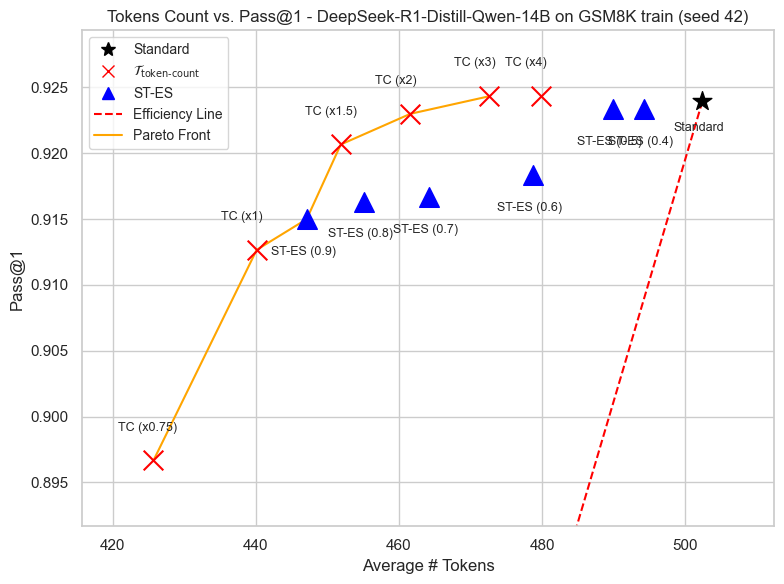}
        \caption{GSM8K - DS-14B}
        \label{fig:token_count_base_ds14b_gsm8k}
    \end{subfigure}
    \hfill
    \begin{subfigure}{0.3\linewidth}
        \centering
        \includegraphics[width=\linewidth]{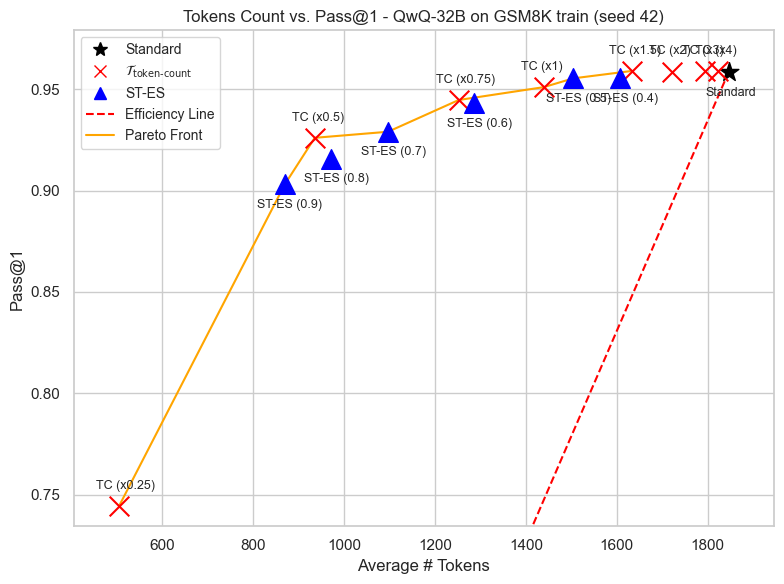}
        \caption{GSM8K - QwQ-32B}
        \label{fig:token_count_base_qwq32b_gsm8k}
    \end{subfigure}
    
    \begin{subfigure}{0.3\linewidth}
        \centering
        \includegraphics[width=\linewidth]{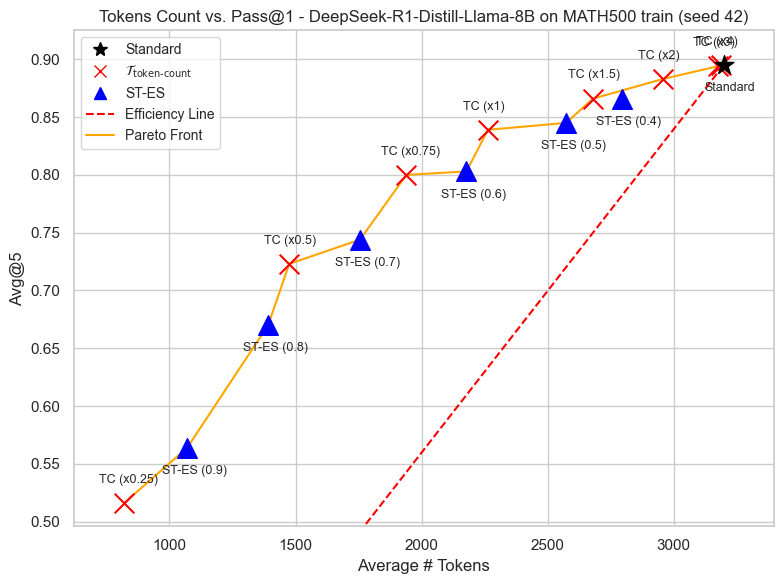}
        \caption{MATH500 - DS-8B}
        \label{fig:token_count_base_ds8b_math500}
    \end{subfigure}
    \hfill
    \begin{subfigure}{0.3\linewidth}
        \centering
        \includegraphics[width=\linewidth]{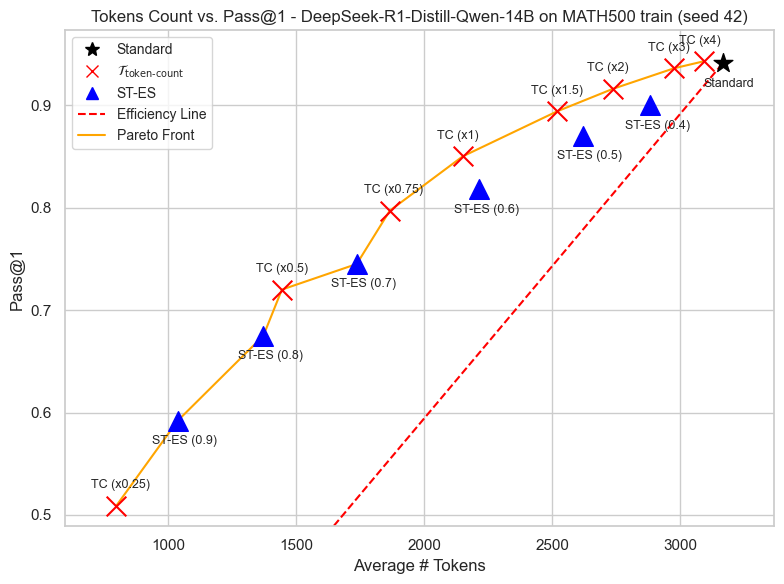}
        \caption{MATH500 - DS-14B}
        \label{fig:token_count_base_ds14b_math500}
    \end{subfigure}
    \hfill
    \begin{subfigure}{0.3\linewidth}
        \centering
        \includegraphics[width=\linewidth]{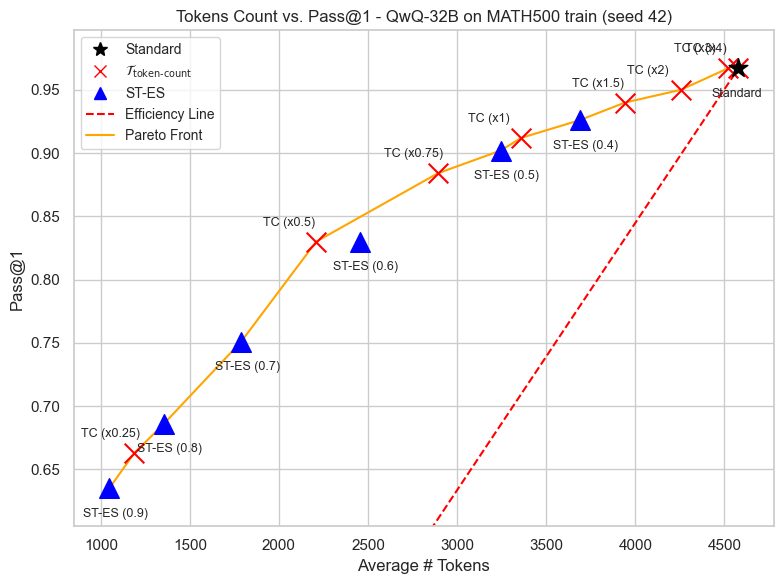}
        \caption{MATH500 - QwQ-32B}
        \label{fig:token_count_base_qwq32b_math500}
    \end{subfigure}

    \begin{subfigure}{0.3\linewidth}
        \centering
        \includegraphics[width=\linewidth]{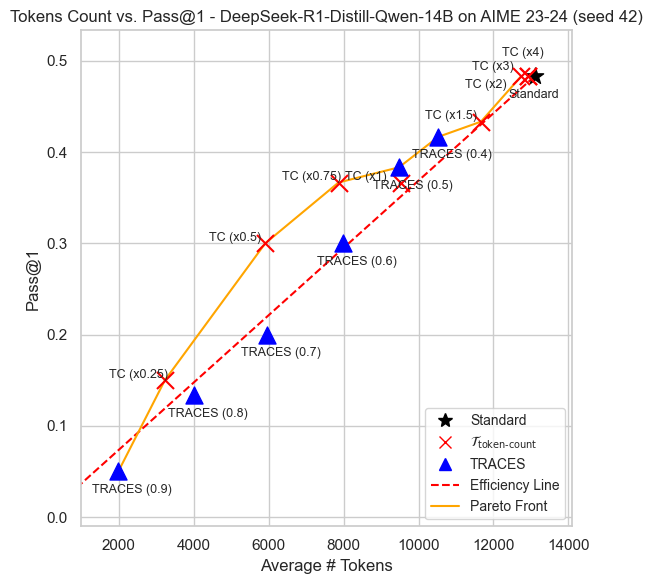}
        \caption{MATH500 - DS-8B}
        \label{fig:token_count_base_ds14b_aime}
    \end{subfigure}
    \hfill
    \begin{subfigure}{0.3\linewidth}
        \centering
        \includegraphics[width=\linewidth]{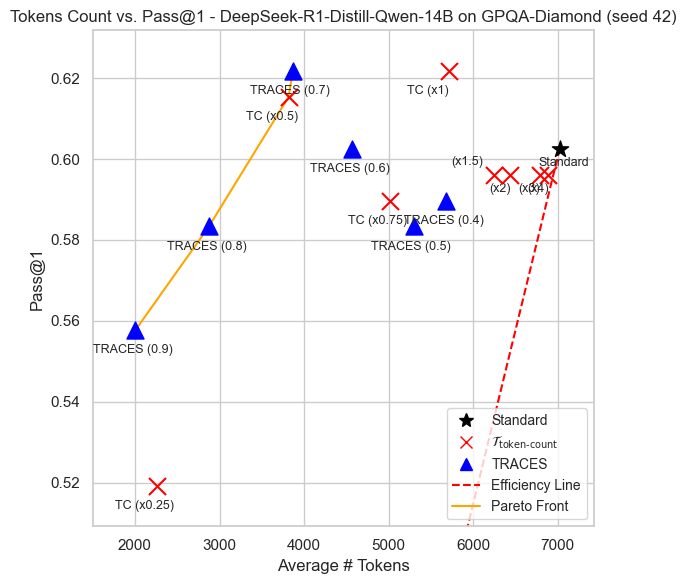}
        \caption{MATH500 - DS-14B}
        \label{fig:token_count_base_ds14b_gpqa}
    \end{subfigure}
    \hfill
    \begin{subfigure}{0.3\linewidth}
        \centering
        \includegraphics[width=\linewidth]{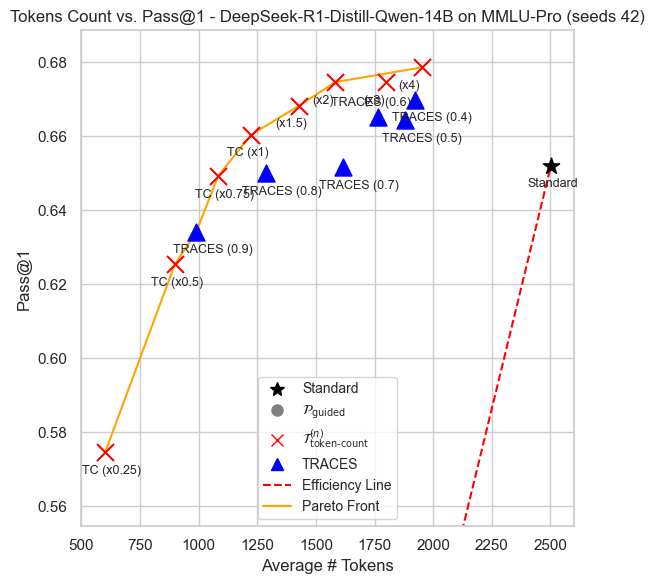}
        \caption{MATH500 - QwQ-32B}
        \label{fig:token_count_base_ds14b_mmlu}
    \end{subfigure}
    \vspace{-0.2cm}
    \caption{ST-ES vs. Token-count $\mathcal{T_{\text{token-count}}}$ baseline - Pass@$1$}
    \label{fig:token-count-budget}
    \vspace{-0.3cm}
\end{figure}

\textbf{$\mathcal{T_{\text{token-count}}}$ baselines.} Figure \ref{fig:token-count-budget} shows the performance of TRACES compared to the $\mathcal{T_{\text{token-count}}}$ baselines. First, we observe that our design of $\mathcal{T_{\text{token-count}}}$ enable a fair comparison of TRACES at equivalent token-count. Indeed, we observe most TRACES configurations has a baseline which has approximately a similar token-count. Second, we see that the performance of TRACES configurations matches closely to the performance of the $\mathcal{T_{\text{token-count}}}$ baselines. Indeed, many TRACES configurations lies on the Pareto front, and when $\mathcal{T_{\text{token-count}}}$ outperforms TRACES at same token-count, it is by a maximum of 2\% to 4\% accuracy. These results shows that our TRACES method matches token-count baseline, without requiring information on the computation efforts required by a model on a dataset.

\textbf{Takeaways.} In this section, we compared TRACES to an additional common approach in the literature, namely: token-budgeting methods. We showed that TRACES matches closely the performance of token-count baselines. Importantly, TRACES does not requires previous knowledge from the configurations (token-count specific to models and datasets).

\newpage

\subsection{White-box early-exit method} \label{sec:appendix-white-box-baselines}

Further, comparing TRACES to state-of-the-art baselines such as DEER \citep{yang2025dynamicearlyexitreasoning}, or EAT \citep{wang2025entropylangletextttthinkrangle} is interesting and necessary to understand how our framework compares to the relevant literature.

However, as we mentioned in the main body (see Section \ref{sec:related-work}), these baselines operate at a different level than TRACES. Indeed, TRACES is a \emph{black-box early-exit method}. TRACES only operates on the text generated by LRMs, without the model's internals. In contrast, early-exit methods such as DEER \citep{yang2025dynamicearlyexitreasoning} and EAT \citep{wang2025entropylangletextttthinkrangle} operate in a white-box and gray-box setting, respectively, therefore requiring some form of internals from the model. For this reason, we set-up two complementary baselines, namely: the prompt-based baselines (see Appendix \ref{sec:appendix-prompt-guided-baselines}), and the token-count baseline (see Appendix \ref{sec:appendix-token-count-baselines}). Both baselines do not requires model's internals, and operate in the exact same setting as TRACES, allowing a fair comparison. We showed that TRACES performs well compared to these baselines.

To further address this gap, we implemented the DEER algorithm \citep{yang2025dynamicearlyexitreasoning}, allowing us to compare TRACES to a state-of-the-art dynamic early-exit method. Table \ref{tab:TRACES-DEER} presents the token-count, saved token-count, and accuracy obtained by applying DEER on DS-Llama8B and Qwen3-8B on MATH500, GSM8K, and GPQA-Diamond test dataset. 

\begin{table}[h]
\centering
\tiny
\begin{tabular}{ll
                ccc
                ccc
                ccc}
\toprule
\multirow{2}{*}{\textbf{Model}}  &
\multirow{2}{*}{\textbf{Config.}} 
& \multicolumn{3}{c}{\textbf{MATH500}} & \multicolumn{3}{c}{\textbf{GSM8K}} & \multicolumn{3}{c}{\textbf{GPQA-Diamond}} \\
\cmidrule(lr){3-5} \cmidrule(lr){6-8} \cmidrule(lr){9-11} 
& & \# Tokens & Saved (\%) & Pass@1 & \# Tokens & Saved (\%) & Pass@1 & \# Tokens & Saved (\%) & Pass@1 \\
\midrule

\multirow{10}{*}{Qwen-3-8B}
  & Base & 3933.3 & - & 0.882 & 2388.8 & - & 0.949 & 9821.6 & - & 0.545 \\
  
  \cmidrule(lr){2-11} 
  
& DEER & 1440.2 & 63.38 & 0.854 & 638.5 & 73.27 & 0.925 & 5147.6 & 47.59 & 0.424 \\
  
  \cmidrule(lr){2-11} 
  
& $\delta=0.4$ & 3208.9 & 18.43 & 0.894 & 1937.8 & 18.88 & 0.947 & 9043.4 & 7.93 & 0.596 \\
& $\delta=0.5$ & 2866.9 & 27.13 & 0.880 & 1779.2 & 25.52 & 0.942 & 8578.7 & 12.66 & 0.581 \\
& $\delta=0.6$ & 2302.5 & 41.47 & 0.858 & 1437.7 & 39.81 & 0.930 & 7128.7 & 27.42 & 0.540 \\
& $\delta=0.7$ & 1756.4 & 55.36 & 0.802 & 1164.8 & 51.24 & 0.909 & 6084.7 & 38.05 & 0.510 \\
& $\delta=0.8$ & 1376.5 & 65.01 & 0.752 & 1003.9 & 57.97 & 0.899 & 5204.7 & 47.01 & 0.469 \\
& $\delta=0.9$ & 1058.3 & 73.09 & 0.670 & 871.9 & 63.50 & 0.878 & 3700.7 & 62.32 & 0.399 \\

\midrule

\multirow{10}{*}{DS-Llama8B} & Base & 3810.6 & - & 0.878 & 865.8 & - & 0.826 & 9427.6 & - & 0.414 \\
  
  \cmidrule(lr){2-11} 
  
& DEER & 2267.2 & 40.50 & 0.834 & 637.1 & 26.41 & 0.805 & 6376.5 & 32.36 & 0.414 \\
  
  \cmidrule(lr){2-11} 
  
& $\delta=0.4$ & 3213.0 & 15.68 & 0.830 & 722.2 & 16.58 & 0.823 & 8334.7 & 11.59 & 0.485 \\
& $\delta=0.5$ & 2849.8 & 25.21 & 0.832 & 681.3 & 21.31 & 0.823 & 7611.2 & 28.94 & 0.485 \\
& $\delta=0.6$ & 2440.1 & 35.96 & 0.782 & 610.6 & 29.47 & 0.819 & 6698.7 & 28.94 & 0.444 \\
& $\delta=0.7$ & 1781.8 & 53.24 & 0.706 & 550.5 & 36.42 & 0.815 & 5117.1 & 45.72 & 0.429 \\
& $\delta=0.8$ & 1308.6 & 65.65 & 0.654 & 515.7 & 40.44 & 0.803 & 3837.9 & 59.29 & 0.419 \\
& $\delta=0.9$ & 962.3 & 74.75 & 0.564 & 486.4 & 43.82 & 0.786 & 2285.6 & 75.76 & 0.379 \\ 

\bottomrule
\end{tabular}
\caption{Performance of TRACES - Qwen-3-8B and DS-Qwen3-8B - seed 42}
\label{tab:TRACES-DEER}
\end{table}

Table 8 shows that DEER performs well on the two selected models, with 26 to 73\% token-count saving, and minimal accuracy drop (from 0\% to -12.1\%). Importantly observe that TRACES is competitive against this new baseline. With DS-Llama8B, TRACES outperformed DEER on GSM8K (40\% vs. 26\% token-count saving at matched accuracy), on GPQA-Diamond (60\% vs. 32\% token-count saving at matched accuracy), as well as on Qwen3-8B on GPQA-Diamond (0.469 vs. 0.424 accuracy at same token-count saving, i.e. ±47\%). Importantly, TRACES configurations that matched DEER's token-count savings achieve higher accuracy.

On MATH500, TRACES obtained slightly lower token-count savings for Qwen3-8B (40 to 50\%) at matched accuracy drop. Similarly, TRACES obtained 25\% token-count saving at matched accuracy drop on DS-Llama8B (vs. 40\% for DEER). As well, on GSM8K for Qwen3-8B, DEER achieves higher token-count saving than TRACES at matched token-count (70\% vs. 40\%).

Importantly, DEER relies on access to model-internal signals. In contrast, TRACES does not have access to additional information using the model's internals (such as the model's confidence).

\newpage

\section{Algorithms}

\subsection{Step-wise Generation} \label{sec:appendix-step_wise_generation}

To generate the reasoning traces of models step-by-step, we need to modify the \texttt{model.generate} function from Hugging Face. However, this process comes at the cost of latency in model generation since we need to interrupt the generation process at each step. The algorithm is presented in Algorithm \ref{alg:stepwise_generation}.

\begin{algorithm}
\footnotesize
\caption{Step-wise Generation}
\label{alg:stepwise_generation}
\begin{algorithmic}[1]
\Require Prompt $x$; reasoning delimiter $\alpha \in V$; max steps $T_{\max}$; language model $\mathcal{M}$; tokenizer $\mathcal{T}$; EOS token

\State $y \gets \mathcal{T}(x)$ \color{gray} \Comment{Tokenized input} \color{black}
\State $S \gets [\,]$; $\beta \gets \emptyset$ \color{gray} \Comment{Initialize output and buffer} \color{black}
\State $s \gets 0$

\While{$s < T_{\max}$}
    \State $t \gets \mathcal{M}(y)$ \color{gray} \Comment{Generate next token} \color{black}
    \State $y \gets y + t$
    \State $\beta \gets \beta + \mathcal{T}^{-1}(t)$ \color{gray} \Comment{Add decoded token to buffer} \color{black}

    \If{EOS in $y$} \color{gray} \Comment{Stop inference if EOS generated} \color{black}
        \State Append $\beta$ to $S$
        \State \textbf{break}
    \EndIf

    \If{$\beta$ ends with $\alpha$} \color{gray} \Comment{Complete and valid step} \color{black}
        \State Append $\beta$ to $S$ 
        \State $\beta \gets ''$ \color{gray} \Comment{Empty the buffer} \color{black}
        \State $s \gets s + 1$ \color{gray} \Comment{Increase $S$ by one step} \color{black}
    \Else
        \State Continue \color{gray} \Comment{Continue until next $\alpha$ or EOS is generated} \color{black}
    \EndIf
        
\EndWhile

\State \Return $S$
\end{algorithmic}
\end{algorithm}

\newpage

\subsection{Early Stopping algorithm} \label{sec:appendix-es_algorithm}

Algorithm \ref{alg:st_es_algorithm} lists the TRACES criteria. The user needs to define a constraint $\delta$, and input a Step-Tagger module $\phi_{\tau^*}$, which returns 1, 2 and 0 if the step tag is $\tau_{\text{constructive}}$, $\tau_{\text{evaluative}}$, or $\tau_{\text{other}}$ (of another type), respectively. If the constraint breaks, the algorithm stops the generation, and prompts the model with $\mathcal{P}_{\text{exit}}$ to give the current best answer.

Note that we added a fixed condition $w$ on top of $c_{\tau*}(S_{running}, \delta)$. It makes our condition robust with regards to possible fluctuations during the monitoring. Indeed, this additional condition ensures that we stop only when the value of the ratio $R_i$ is lower than $\delta$ for $w=5$ consecutive steps (validating that the model entered a reasoning phase transition).

\begin{algorithm}
\footnotesize
\caption{TRACES}
\label{alg:st_es_algorithm}
\begin{algorithmic}[1]
\Require Prompt $x$; reasoning delimiter $\alpha \in V$; Reasoning Language Model $\mathcal{M}$; tokenizer $\mathcal{T}$; EOS token $\gamma$; Constraint \{$\delta, w=5$\}; Step-Tagger $\phi_{\tau^*}$; Early-Exit Prompt $\mathcal{P}_{\text{exit}}$

\State $y \gets \mathcal{T}(x)$ \color{gray} \Comment{Tokenize the input} \color{black}
\State $S_{running} \gets [\,]$; \color{gray} \Comment{Initialize output} \color{black}
\State $t \gets 0$
\State $f_{\tau_{\text{constructive}}}, f_{\tau_{\text{evaluative}}} \gets 0, 0$ \color{gray} \Comment{Initialize frequency track of $\tau_{\text{constructive}}$ and $\tau_{\text{evaluative}}$} \color{black}

\While{$c_{\tau*}(S_{running}, \delta)$} \color{gray} \Comment{Generate until constraint breaks} \color{black}
    \State Generate step $s_i$ using $\mathcal{M}$ and $\alpha$
    \State $y \gets s_i$

    \State $\tau_i \gets \phi_{\tau^*}(s_i)$ \color{gray} \Comment{Step-Tagging prediction} \color{black}

    \If{$\tau_i \in \tau_{\text{constructive}}$}
        \State $f_{\text{constr}} \gets f_{\text{constr}} + 1$
    \ElsIf{$\tau_i \in \tau_{\text{evaluative}}$}
        \State $f_{\text{eval}} \gets f_{\text{eval}} + 1$
    \EndIf

    \If{$f_{\text{constr}} + f_{\text{eval}} = 0$}
        \State $R \gets 1$
    \Else
        \State $R \gets \frac{f_{\text{constr}}}{f_{\text{constr}} + f_{\text{eval}}}$ \color{gray} \Comment{Compute $R$} \color{black}
    \EndIf

    \If{$R < \delta$}
        \State Append $1$ to $F$ \color{gray} \Comment{Flag sequence} \color{black}
    \Else
        \State Append $0$ to $F$
    \EndIf

    \If{$\sum F_{[-w:]} = w$}
        \State \textbf{break} \color{gray}  \Comment{Trigger early stopping when $R < \delta$ for $w=5$ consecutive steps} \color{black} 
    \EndIf

    \State $t \gets t + 1$
\EndWhile

\State $y \gets \mathcal{M}(y+\mathcal{P}_{\text{exit}})$ \color{gray} \Comment{Infer $\mathcal{M}$ with the early exit prompt} \color{black}

\State \Return y

\end{algorithmic}
\end{algorithm}


\newpage

\section{Training details of lightweight Step-Tagging module} \label{sec:appendix-step-tagging-training}



\textbf{Experimental Design.} We selected the \texttt{bert-base-uncased} sentence classifier \citep{devlin2019bertpretrainingdeepbidirectional} to construct our Step-Tagging framework, including a single hidden layer. Given the large and fine-grained nature of our taxonomy ($13$ distinct step types), training a multi-class classifier is challenging due to significant class imbalance. To address this, we trained a sentence classifier to identify separately 3 classes of steps. Our classifier is trained to assign to steps the tag $1$ and $2$ for step-types in the classes $\tau_{\text{constructive}}$ and $\tau_{\text{evaluative}}$, respectively. Our classifier also assigns the label $0$ for any other step-types -- such steps does not affects our ratio $R$. This approach notably improved detection accuracy across low-frequency categories, and fits our definition of early-stopping constraint: detecting \emph{constructive} and \emph{evaluative} step-type to estimate the ratio $R$.

\begin{figure}[h]
    \centering
    \begin{subfigure}{0.48\linewidth}
        \centering
        \includegraphics[width=\linewidth]{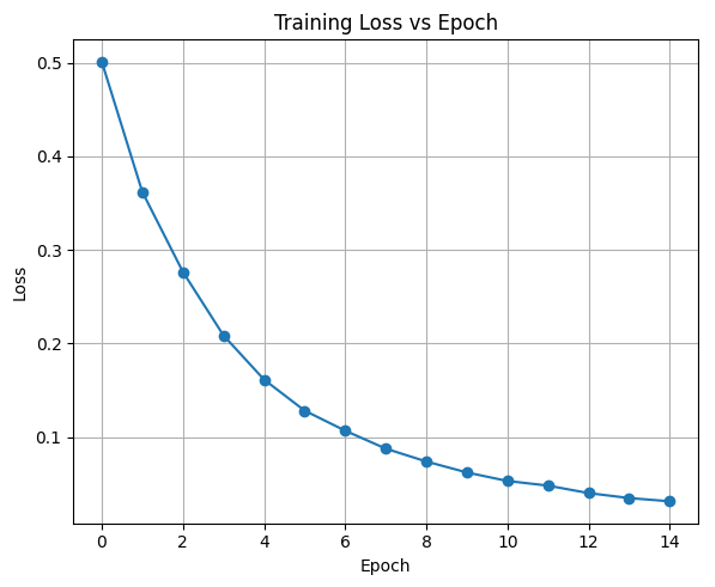}
        \caption{Training loss}
        \label{fig:step-tag-loss}
    \end{subfigure}
    \hfill
    \begin{subfigure}{0.48\linewidth}
        \centering
        \includegraphics[width=\linewidth]{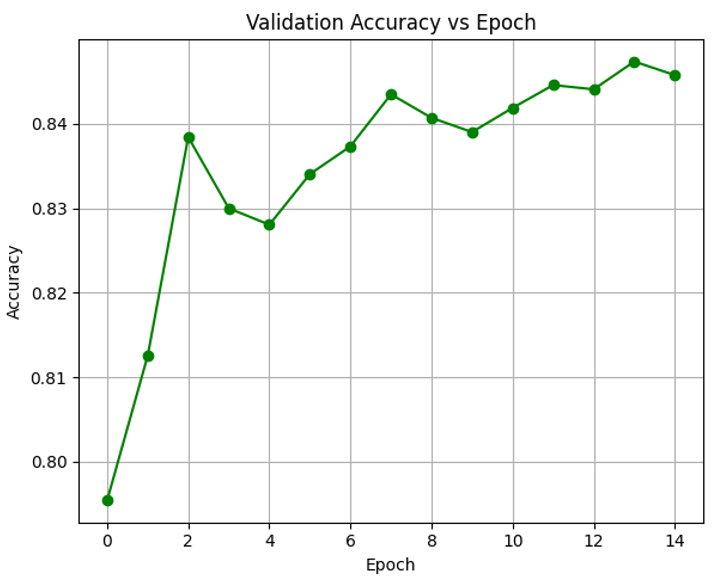}
        \caption{Validation accuracy}
        \label{fig:step-tag-val-acc}
    \end{subfigure}
    \vspace{-0.2cm}
    \caption{Training metrics - Step-Tagging module}
    \label{fig:train-step-tag}
    \vspace{-0.3cm}
\end{figure}

\textbf{Training details.} Figure \ref{fig:step-tag-loss} and \ref{fig:step-tag-val-acc} present the training loss and validation accuracy, respectively. We trained our classifier for $15$ epochs. The batch size is 16 and we used an AdamW optimizer with a learning rate of $2.10^{-5}$. To evaluate the performance of our classifiers, we computed the Macro-F1 and Micro-F1 on the test datasets. While the \emph{Macro-F1} helps to identify the classifier's ability to detect rare classes, the \emph{Micro-F1} offers a more global view on the step detector's performance across all steps.

\newpage

\section{Influencing factors of the TRACES framework}

In this Section, we inspect the influence of the parameters of our approach. While we showed in Section \ref{sec:st-es-evaluation} that our approach seems to generalize well across various models and datasets, we inspect the robustness of our methods across its parameters.

\subsection{Threshold $\delta$} \label{sec:appendix-infl-fact-threshold}



\begin{figure}[h]
    \centering
    \begin{subfigure}{0.45\textwidth}
        \includegraphics[width=\linewidth]{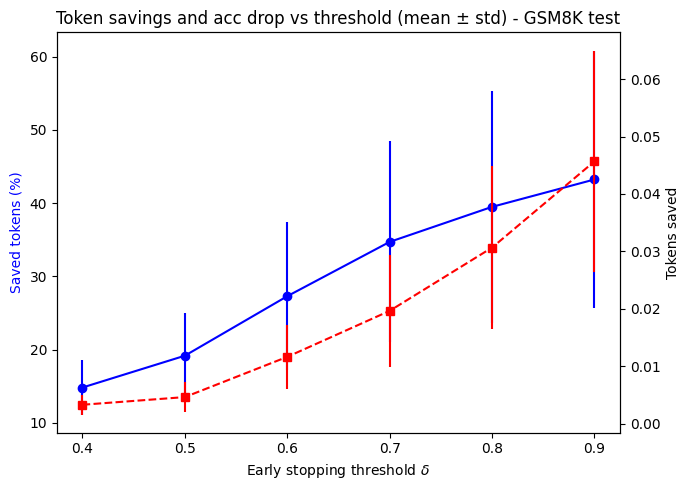}
        \caption{GSM8K}
        \label{fig:GSM8K_threshold}
    \end{subfigure}
    \hfill
    \begin{subfigure}{0.45\textwidth}
        \includegraphics[width=\linewidth]{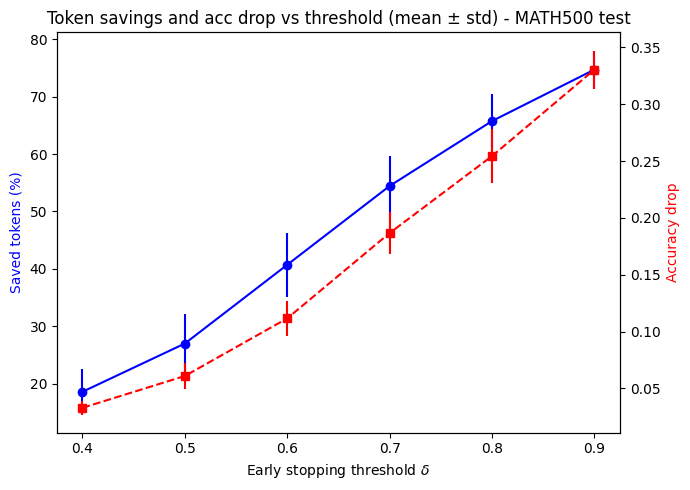}
        \caption{MATH500}
        \label{fig:MATH500_threshold}
    \end{subfigure}
    \vspace{-0.2cm}
    \caption{Efficiency trade-off against threshold $\delta$ over the three selected LRMs}
    \label{fig:threshold_efficiency}
    \vspace{-0.2cm}
\end{figure}

We compared in Section \ref{sec:st-es-evaluation} the performance of our TRACES Early-Stopping criteria. We showed that the value of the threshold can allow adaptive computation at inference time. Figure \ref{fig:threshold_efficiency} present the efficiency trade-off of our criteria based on the value of the thresholds $\delta$, on GSM8K and MATH500 across all selected models.

In Figure \ref{fig:threshold_efficiency}, we observe that reducing the token-count agressively on GSM8K has generally less impact on the model's accuracy than on MATH500. Indeed, we observe for same token-count saving proportion (50\%) the accuracy drop is around $4.5\%$ vs. $17\%$ for GSM8K and MATH500, respectively (with $\delta = 0.9$ and $\delta = 0.7$, respectively). 

First, MATH500 results in higher token-count. It means that for the same proportion, TRACES prune more tokens on MATH500 than on GSM8K. And the validation phase often represents a high proportion on higher token-count datasets (see Figure \ref{fig:IES-train} in Section \ref{sec:appendix-ideal-early-stopping}). Together, it explains why datasets resulting in higher token-count results in higher token-count saving with high value of $\delta$.

Second, we observe that early-exiting is a harder problem for complex datasets (e.g. MATH500 compared to GSM8K). Indeed, we observe in Figure \ref{fig:answer-correctness-rate} in Appendix \ref{sec:appendix-ideal-early-stopping} that the answer correctness is more variable on MATH500 than on GSM8K -- from the IES step to the end of the generation. This observation also contributes to justify why we are observing higher accuracy drop on MATH500 than GSM8K

\textbf{Takeaways:} Together, the observations made in this section supports the fact that TRACES is more sensible on harder reasoning tasks. While TRACES is flexible ($\delta$ can be used to target a wide range of token-count saving), high values of $\delta$ on problems requiring high number of tokens can result in aggressive token-count saving at the cost of the accuracy. 

This can be mitigated by setting lower values of $\delta$, still achieving reasonable token-count saving with minimal accuracy drop. As well, this observation supports that complex reasoning problems needs more evaluative steps to conserve a good accuracy.

\newpage

\subsection{Number and Nature of class $n$} \label{sec:appendix-infl-fact-classes-ratio}



\textbf{Objective.} In this subsection, we assess how robust our criteria is with regards to the step-type classes to compute our ratio $R$. We have seen in Section \ref{sec:influence-step-type} that $\tau_{\text{constructive}} = \{\tau_{\text{Problem-Restatement}}, \tau_{\text{Definition Recall}}\}$ and $\tau_{\text{evaluative}} = \{\tau_{\text{Verification}}, \tau_{\text{Final Conclusion}}\}$ are common across models and datasets. Furthermore, the ratio $R$ plot on training datasets, as well as the evaluation Sections \ref{sec:influence-step-type} and \ref{sec:st-es-evaluation} show that our fixed reasoning transition classes lead to good performance. This ablation is comparing our methodology and framework against different selections of tags.

\begin{table}[h]
    \centering
    \tiny
    \begin{tabular}{cccccccc}
        \toprule
        \multirow{2}{*}{\textbf{\# of tags}} & \multirow{2}{*}{\textbf{Config.}} 
        & \multicolumn{3}{c}{\textbf{$\tau_{\text{constructive}}$}} 
        & \multicolumn{3}{c}{\textbf{$\tau_{\text{evaluative}}$}} \\
        \cmidrule(lr){3-5} \cmidrule(lr){6-8}
        & & \emph{PB Restat} & \emph{Def Recall} & \emph{Context Rep} & \emph{Verif} & \emph{Final Concl} & \emph{Alt App} \\
        \midrule

        \multirow{3}{*}{\textbf{1}} & 1 & \cmark &  &  & \cmark &  &  \\
         & 2 &  & \cmark &  &  & \cmark &  \\
         & 3 &  &  & \cmark &  &  & \cmark \\

        \cmidrule(lr){1-8}
         
        \multirow{3}{*}{\textbf{2}} & 4 & \cmark & \cmark &  & \cmark & \cmark &  \\
         & 5 & \cmark &  & \cmark & \cmark &  & \cmark \\
         & 6 &  & \cmark & \cmark &  & \cmark & \cmark \\

        \cmidrule(lr){1-8}

        \textbf{3} & 7 & \cmark & \cmark & \cmark & \cmark & \cmark & \cmark \\
        
        \bottomrule
    \end{tabular}
    \caption{Alternative reasoning transition classes}
    \label{tab:ablation_n_classes}
\end{table}

\textbf{Alternative Reasoning Transition classes.} Table \ref{tab:ablation_n_classes} present alternative reasoning transition sets of step-types that we use to compute the ratio $R$. We suggest configurations including 1, 2, and 3 step-types per class. To allow fair comparison, we mix different configurations of classes. We conduct this ablation on the \deepseekLlama{} model, on both GSM8K and MATH500 train datasets. Given our results from the main body of the paper, we expect to obtain similar results on other datasets and models.

\begin{figure}[h]
    \centering
    \begin{subfigure}{0.9\textwidth}
        \includegraphics[width=\linewidth]{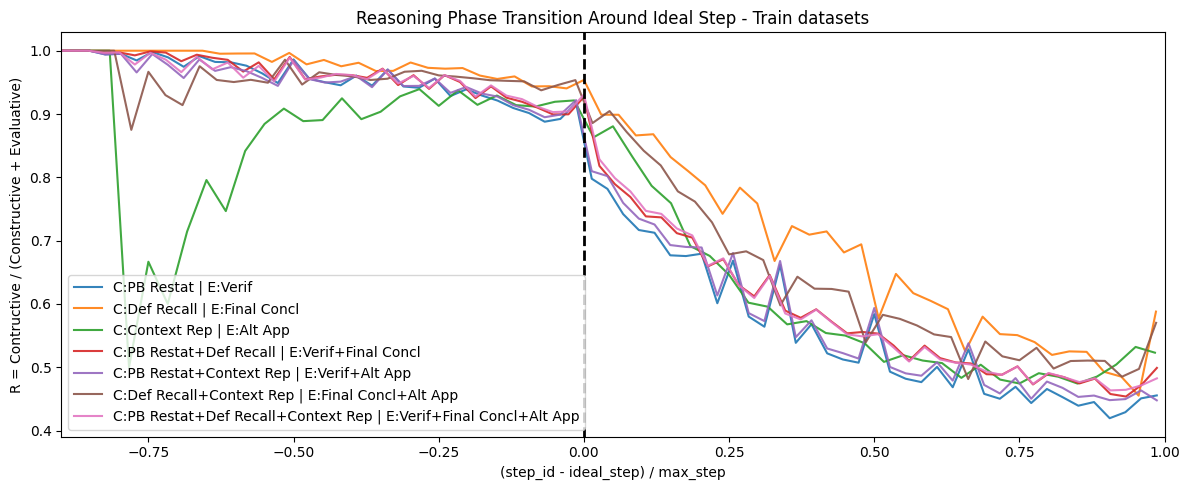}
        \caption{GSM8K}
        \label{fig:GSM8K_reason_phase}
    \end{subfigure}
    \hfill
    \begin{subfigure}{0.9\textwidth}
        \includegraphics[width=\linewidth]{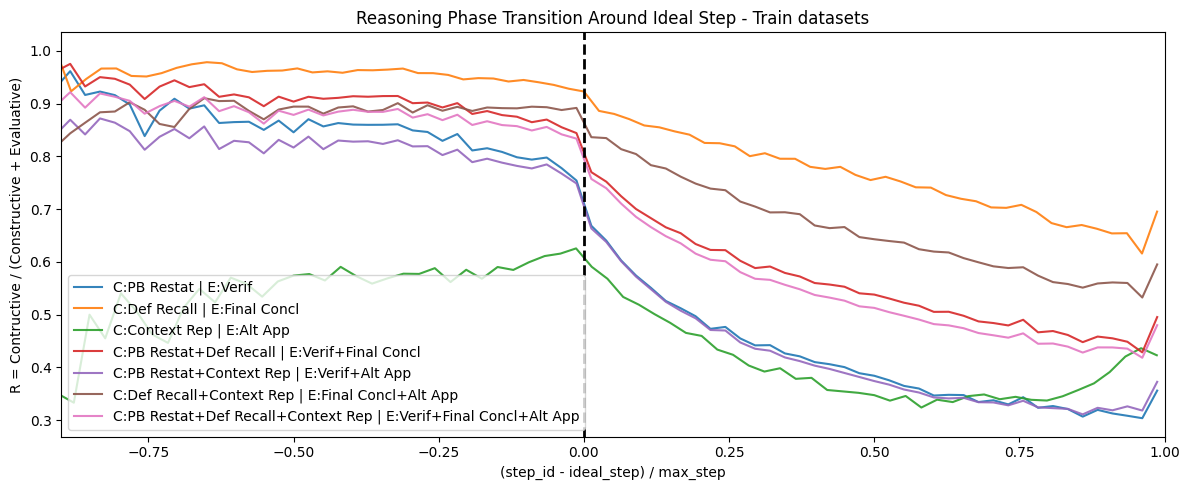}
        \caption{MATH500}
        \label{fig:MATH500_reason_phase}
    \end{subfigure}
    \vspace{-0.2cm}
    \caption{Ratio R for different reasoning transition classes}
    \label{fig:reason_phase_alternative_reason_transition}
    \vspace{-0.2cm}
\end{figure}

\textbf{Reasoning transition.} Figure \ref{fig:reason_phase_alternative_reason_transition} shows the ratio $R$ for each configurations presented for GSM8K and MATH500 train. First, we observe that some ratios are poorly adapted to observe a reasoning transition. Indeed, configurations 3 and 6 lead to a drop of the ratio before the IES step (i.e. $x=0$). This is due to the distribution observed. Indeed, the class \emph{Alternative Exploration} is almost balanced between before and after IES step. It means that these steps appears before IES, decreasing the value of $R$ per its definition.

Conversely, configurations 2, 3 and 6 obtained higher value of $R$ after IES, implying a less pronounced reasoning transition phase. Specifically, these configurations includes  \emph{Context Repetition} but excludes \emph{Problem Re-statement}. The delta $\Delta_{\text{before-after}}$ of \emph{Context Repetition} is lower than the one from \emph{Problem Re-statement}, meaning that some \emph{Context Repetition} appears more frequently after IES, increasing the value of $R$. 

The same observation holds for the MATH500 dataset. Overall, we observe that configurations including \emph{Problem Re-statement}, \emph{Verification} and/or \emph{Final Conclusion} result in similar curves: almost constant ratio $R$ before the IES, and then drops after.

\begin{figure}[h]
    \centering
    \begin{subfigure}{0.49\textwidth}
        \includegraphics[width=\linewidth]{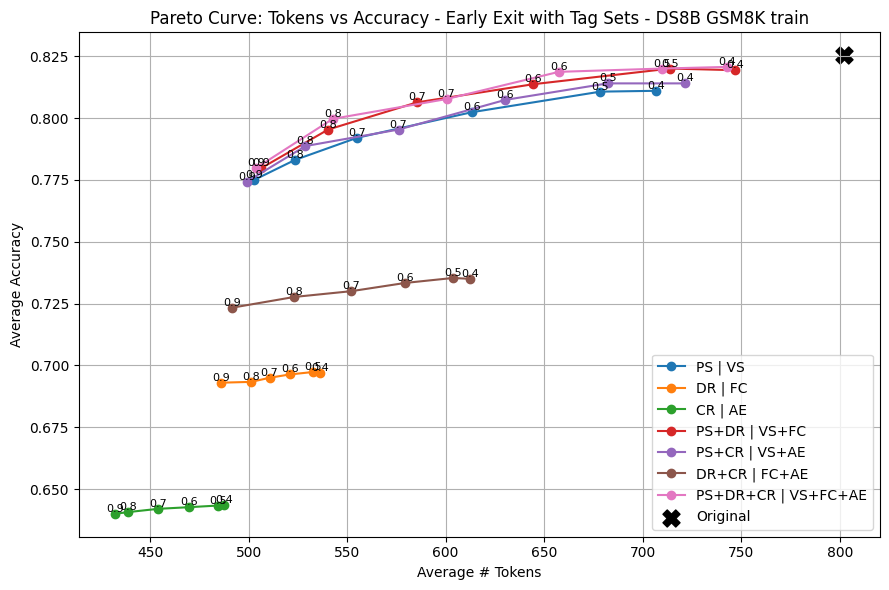}
        \caption{GSM8K}
        \label{fig:GSM8K_reason_phase_altern}
    \end{subfigure}
    \hfill
    \begin{subfigure}{0.49\textwidth}
        \includegraphics[width=\linewidth]{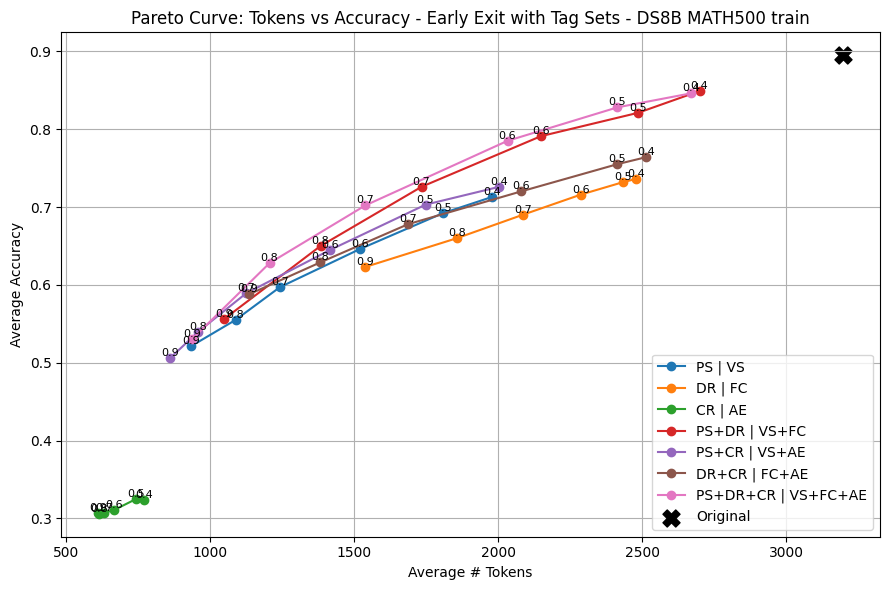}
        \caption{MATH500}
        \label{fig:MATH500_reason_phase_altern}
    \end{subfigure}
    \vspace{-0.2cm}
    \caption{TRACES performance}
    \label{fig:st_es_alternative_reason_transition}
    \vspace{-0.2cm}
\end{figure}

\textbf{TRACES Early-Stopping.} Figure \ref{fig:st_es_alternative_reason_transition} presents the TRACES of the various configurations. We observe that it aligns with Figure \ref{fig:reason_phase_alternative_reason_transition}. Configurations with ratio $R$ dropping before IES stops the generation early, at the cost of the accuracy. All the other configurations lead to similar behavior, confirming our selection.

\textbf{Takeaways.} To conclude, we observed that the choice of the selected step-types affects the quality of the reasoning transition signal, and by extension, the performance of the TRACES framework. First, not all the step-types are equally informative. Classes such as \emph{Alternative Exploration} or \emph{Context Repetition} exhibits more balanced distribution before and after IES, making them poorly informative of the reasoning transition. Furthermore, more classes seems to add noise in the signal, diluting the imbalancity of classes.

\newpage

\subsection{Taxonomy used} \label{sec:appendix-infl-fact-taxonomy}


\textbf{Objective.} In this ablation study, we assess how robust our taxonomy is for the TRACES criteria. To do so, we run the same experiment that we conducted in the main body of the paper on different alternative versions of our original taxonomy. 

\textbf{Alternative Taxonomies.} Our original taxonomy is wide and fine-grained, containing 13 categories of labels (excluding the placeholder label ``Other''). Therefore, we reduce the number of labels in the taxonomy, and grouped similar labels at different levels of abstraction. Table \ref{tab:ablation_taxonomy} shows resulting the taxonomies, considering from $13, 6, 4, 3,$ and $2$ labels. A binary label represent the simplest form of constraint, where we only identify $\tau_{\text{constructive}}$ or $\tau_{\text{evaluative}}$ setp-types.

\begin{table}[h]
    \centering
    \tiny
    \begin{tabular}{ccccc}
        \toprule
        \textbf{Tags ids} & \textbf{6-labels} & \textbf{4-labels} & \textbf{3-labels} & \textbf{2-label} \\
        \midrule
        \textbf{Problem Re-Statement} & \multirow{3}{*}{\textbf{Setup}} & \multirow{3}{*}{\textbf{Early Reasoning}} & \multirow{3}{*}{\textbf{Early Reasoning}} & \multirow{8}{*}{\textbf{Early Reasoning}} \\
        Context Repetition &  &  &  &  \\
        \textbf{Definition Re-call} &  &  &  &  \\ \cmidrule(lr){2-2} \cmidrule(lr){3-3} \cmidrule(lr){4-4}
        Formula Substitution & \multirow{2}{*}{Manipulation} & \multirow{4}{*}{Mid Reasoning} & \multirow{4}{*}{Mid Reasoning} &  \\
        Symbolic Transformation &  &  &  &  \\ \cmidrule(lr){2-2} 
        Edge Case & \multirow{2}{*}{Analysis} &  &  &  \\
        Pattern Recognition &  &  &  &  \\ \cmidrule(lr){2-2} \cmidrule(lr){3-3} \cmidrule(lr){4-4} \cmidrule(lr){5-5}
        Exploration & \multirow{3}{*}{Meta Reasoning} & \multirow{5}{*}{\textbf{Late Reasoning}} & \multirow{7}{*}{\textbf{Late Reasoning}} & \multirow{7}{*}{\textbf{Late Reasoning}} \\
        Interpretation &  &  &  &  \\ 
        Self-Talk &  &  &  &  \\ \cmidrule(lr){2-2}
        \textbf{Verification} & \multirow{2}{*}{\textbf{Checking}} &  &  &  \\
        Heuristic / Intuition &  &  &  &  \\ \cmidrule(lr){2-2} \cmidrule(lr){3-3}
        \textbf{Final Conclusion} & End Reasoning & End Reasoning &  &  \\
        \bottomrule
    \end{tabular}
    \caption{Alternative taxonomies - we regrouped labels at different levels of abstraction to observe the impact of the taxonomy on the TRACES early-stopping criteria}
    \label{tab:ablation_taxonomy}
\end{table}

\textbf{Methodology.} To address our objective, we performed the same analysis  presented in Section \ref{sec:influence-step-type}, using the different taxonomies (i.e. vocabulary of tags $\mathcal{T}$). For each taxonomies, the experiment resulted in \emph{Category distributions} $S_{\text{before}}$ and $S_{\text{after}}$. For each models, we re-used the MATH500 and GSM8K training datasets labeled by \texttt{GPT-4o-mini} using our methodology explicated in Section \ref{sec:influence-step-type}. We then merged labels as in Table \ref{tab:ablation_taxonomy}.

\begin{figure}[h]
    \centering
    \begin{subfigure}{0.49\textwidth}
        \includegraphics[width=\linewidth]{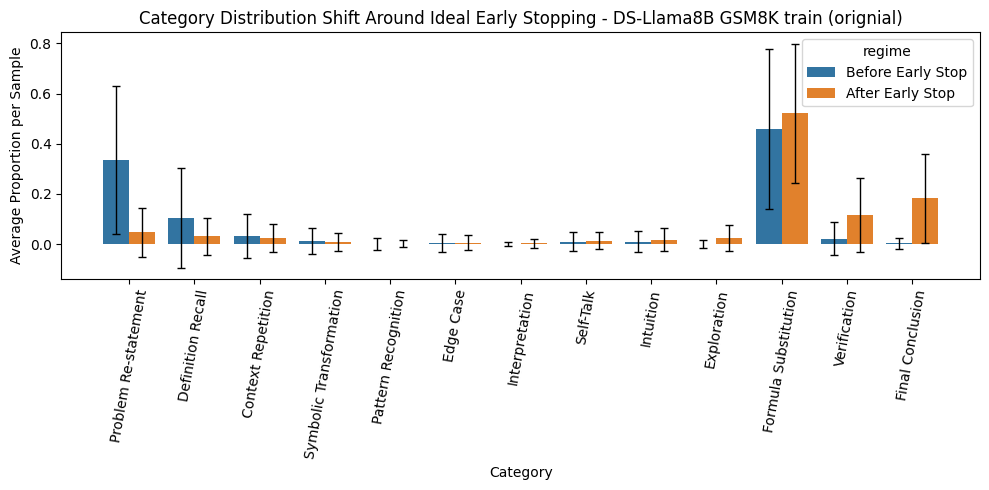}
        \caption{Original - GSM8K}
        \label{fig:GSM8K_alt_taxo_original}
    \end{subfigure}
    \hfill
    \begin{subfigure}{0.49\textwidth}
        \includegraphics[width=\linewidth]{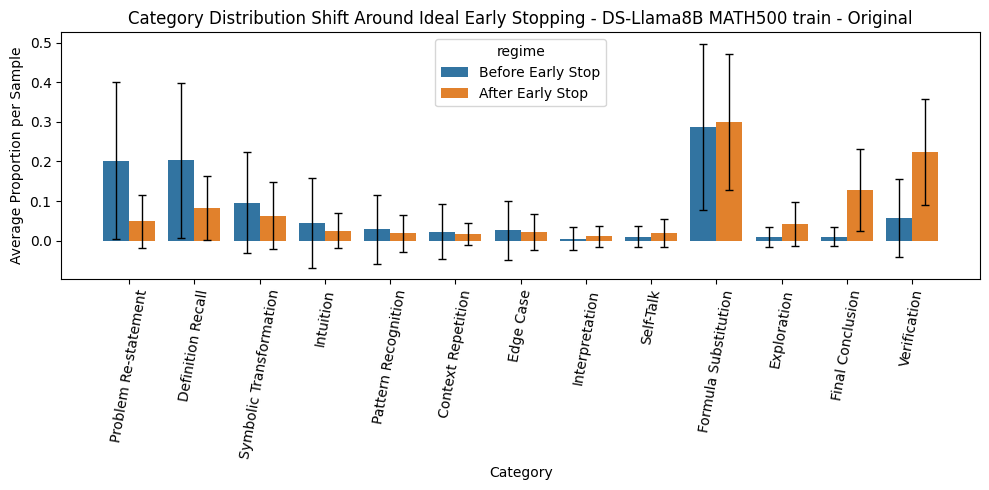}
        \caption{Original - MATH500}
        \label{fig:MATH500_alt_taxo_original}
    \end{subfigure}
    \begin{subfigure}{0.49\textwidth}
        \includegraphics[width=\linewidth]{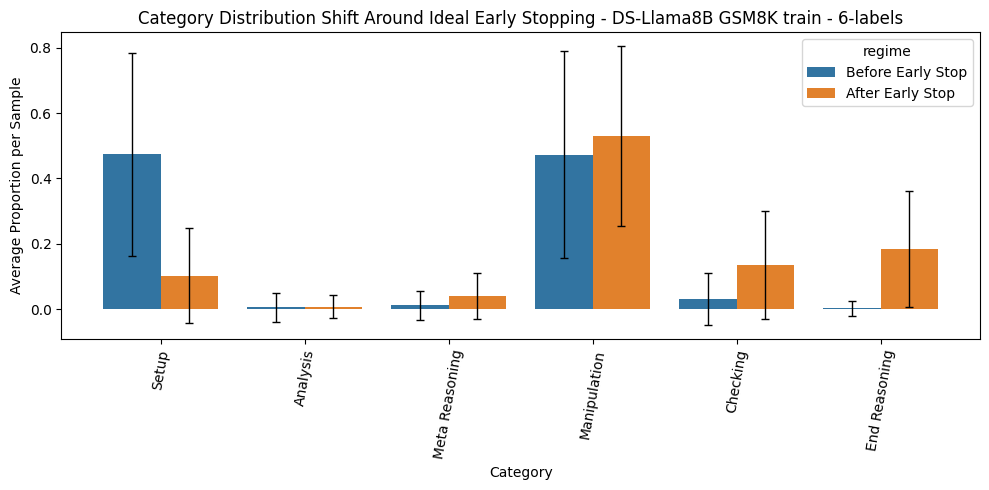}
        \caption{6-labels - GSM8K}
        \label{fig:GSM8K_alt_taxo_6_labels}
    \end{subfigure}
    \hfill
    \begin{subfigure}{0.49\textwidth}
        \includegraphics[width=\linewidth]{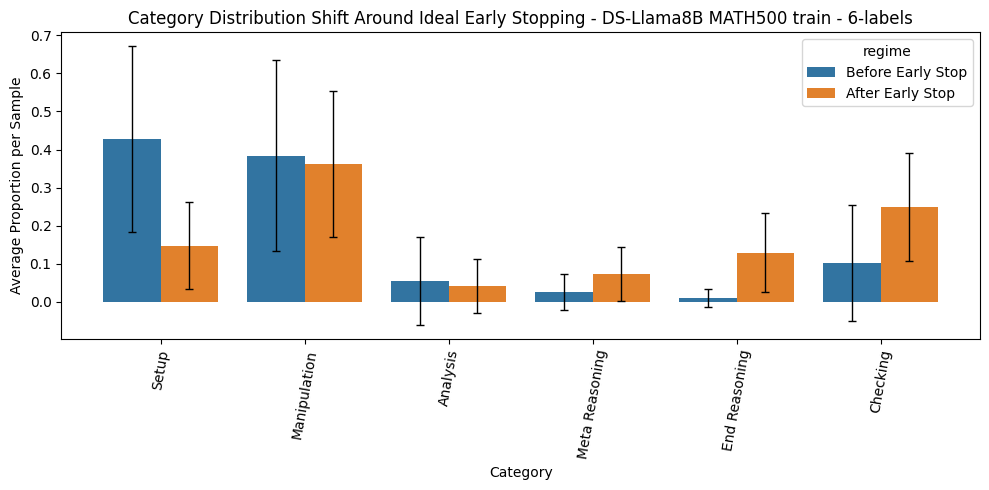}
        \caption{6-labels - MATH500}
        \label{fig:MATH500_alt_taxo_6_labels}
    \end{subfigure}
    \begin{subfigure}{0.49\textwidth}
        \includegraphics[width=\linewidth]{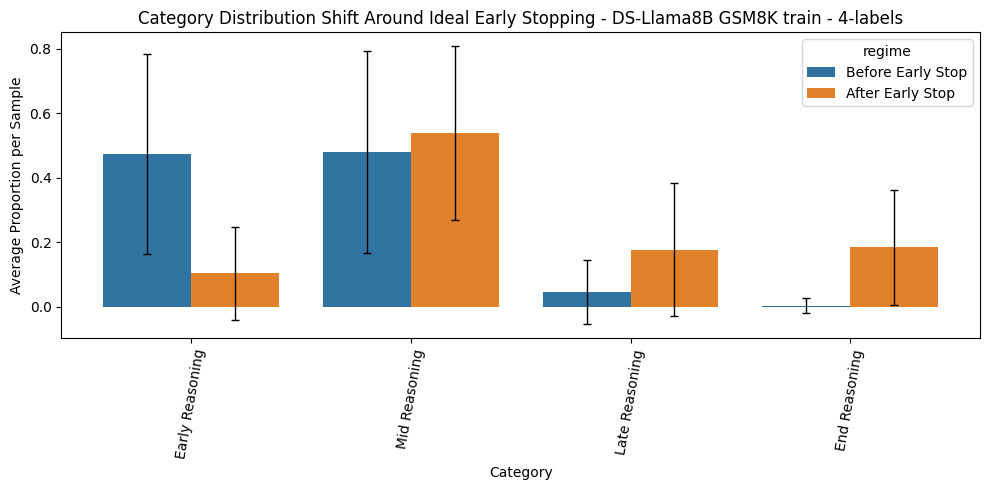}
        \caption{4-labels - GSM8K}
        \label{fig:GSM8K_alt_taxo_4_labels}
    \end{subfigure}
    \hfill
    \begin{subfigure}{0.49\textwidth}
        \includegraphics[width=\linewidth]{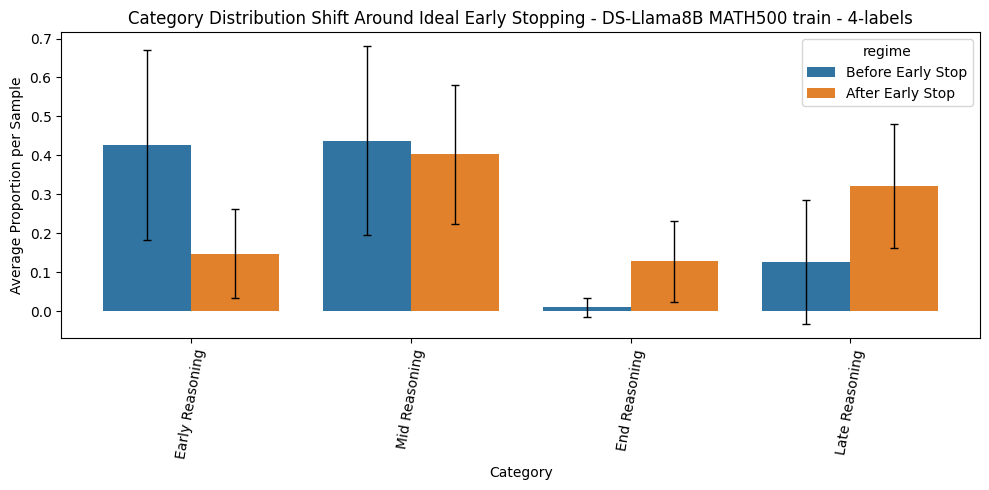}
        \caption{4-labels - MATH500}
        \label{fig:MATH500_alt_taxo_4_labels}
    \end{subfigure}
    \begin{subfigure}{0.49\textwidth}
        \includegraphics[width=\linewidth]{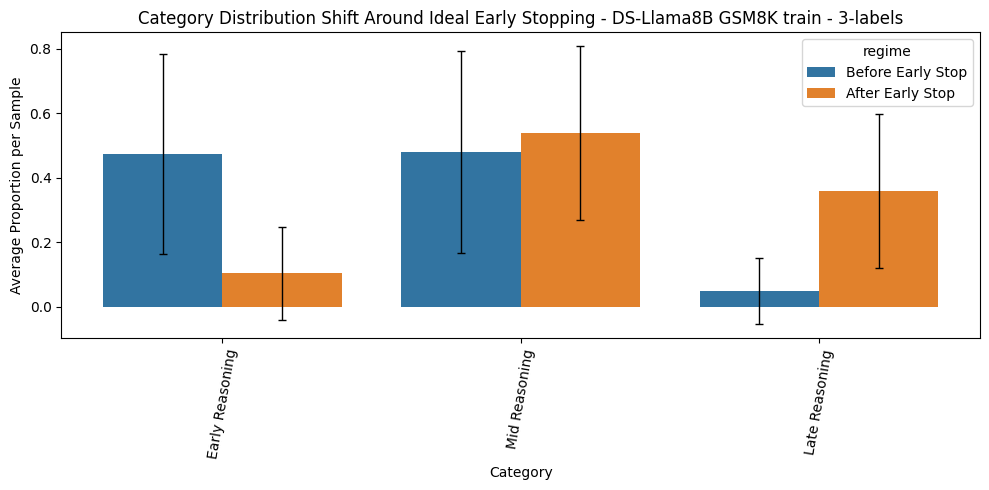}
        \caption{3-labels - GSM8K}
        \label{fig:GSM8K_alt_taxo_3_labels}
    \end{subfigure}
    \hfill
    \begin{subfigure}{0.49\textwidth}
        \includegraphics[width=\linewidth]{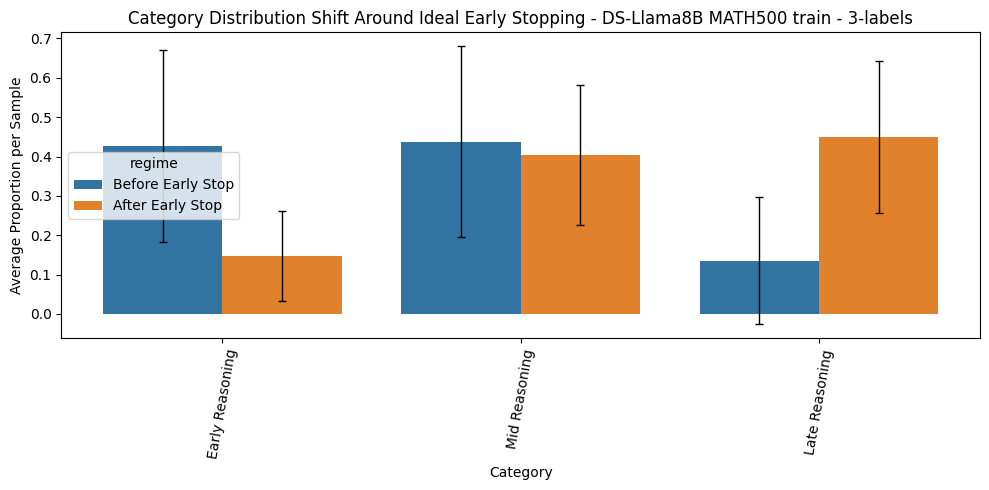}
        \caption{3-labels - MATH500}
        \label{fig:MATH500_alt_taxo_3_labels}
    \end{subfigure}
    \begin{subfigure}{0.49\textwidth}
        \includegraphics[width=\linewidth]{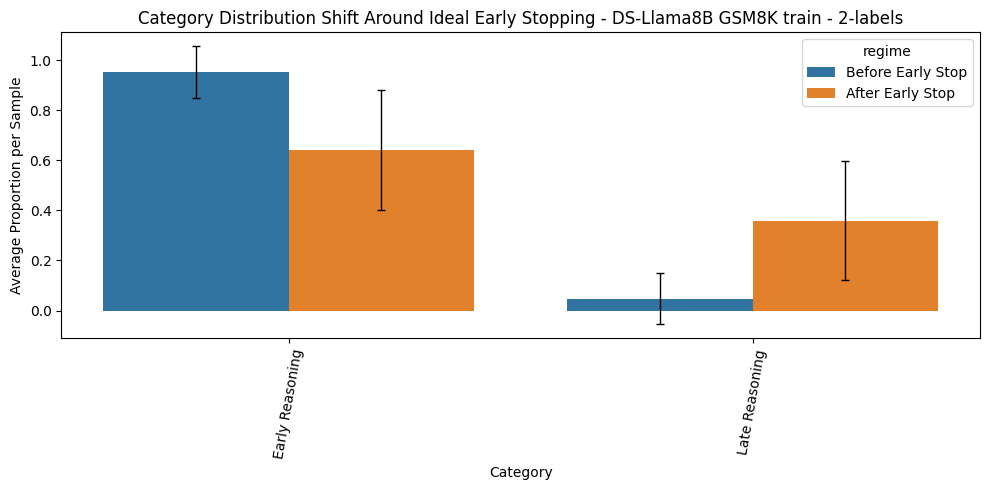}
        \caption{2-labels - GSM8K}
        \label{fig:GSM8K_alt_taxo_2_labels}
    \end{subfigure}
    \hfill
    \begin{subfigure}{0.49\textwidth}
        \includegraphics[width=\linewidth]{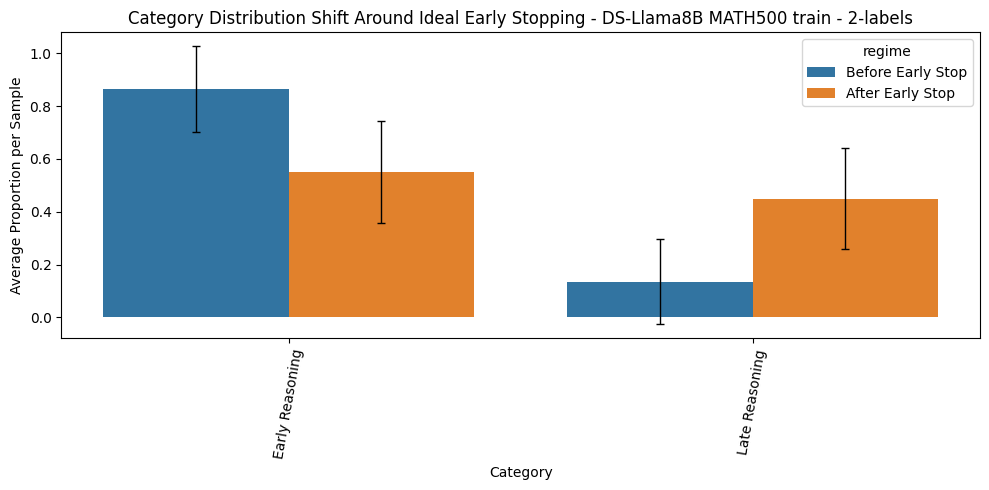}
        \caption{2-labels - MATH500}
        \label{fig:MATH500_alt_taxo_2_labels}
    \end{subfigure}
    \vspace{-0.2cm}
    \caption{Distribution of labels around $S_{\text{IES}}$ - Alternative taxonomies on DS-Llama8B}
    \label{fig:distribution_alternative_taxonomy}
    \vspace{-0.2cm}
\end{figure}

\textbf{Distribution of labels around $S_{\text{IES}}$.} Figure \ref{fig:distribution_alternative_taxonomy} presents the distribution of the tags for $S_{\text{before}}$ and $S_{\text{after}}$ (orange and blue bars, respectively). We note that for some labels, merging tags together in higher level classes enphasis the difference in distribution between tags before and after $S_{\text{IES}}$. 

For each alternative taxonomies, we selected specific $\tau_{\text{constructive}}$ and $\tau_{\text{alternative}}$ to compute the ratio $R$. Specifically, we set $\tau_{\text{constructive}}$ to \emph{Setup}, and \emph{Early Reasoning} for labels $6$ and $4,3,2$, respectively. Similarly, we set $\tau_{\text{evaluative}}$ to \emph{Checking}, and \emph{Late Reasoning} for labels $6$ and $4,3,2$, respectively. Selected tags for each categories are highlighted in bold in Table \ref{tab:ablation_taxonomy}.

\newpage

\begin{figure}[t]
    \centering
    \begin{subfigure}{0.8\textwidth}
        \includegraphics[width=\linewidth]{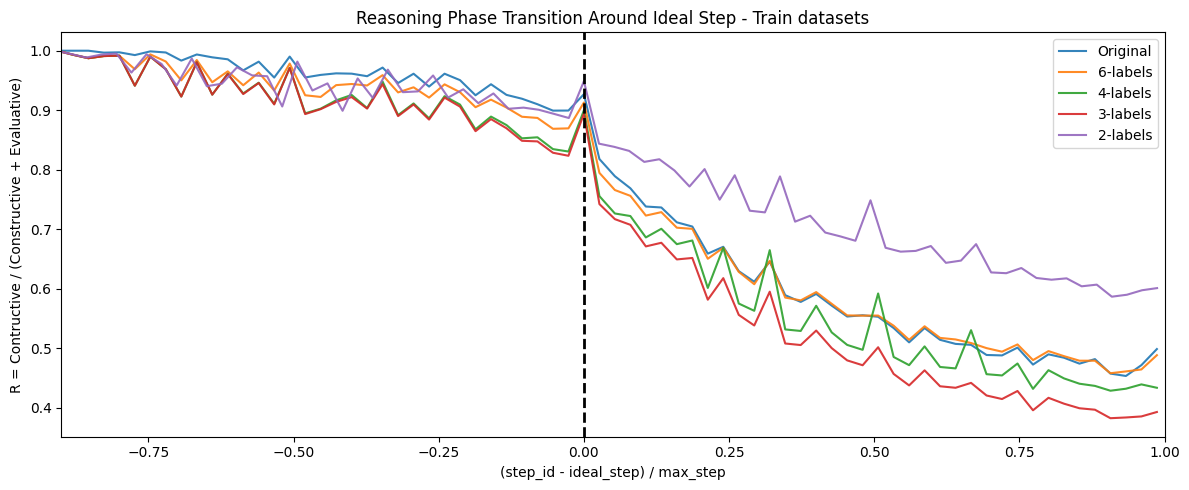}
        \caption{GSM8K}
        \label{fig:GSM8K_reason_phase_alt_taxo}
    \end{subfigure}
    \hfill
    \begin{subfigure}{0.8\textwidth}
        \includegraphics[width=\linewidth]{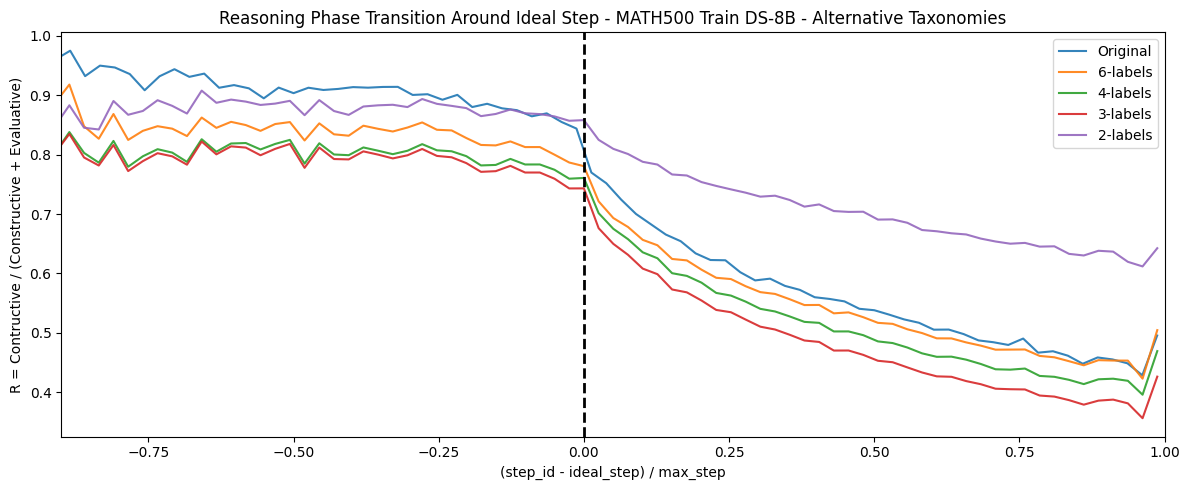}
        \caption{MATH500}
        \label{fig:MATH500_reason_phase_alt_taxo}
    \end{subfigure}
    \vspace{-0.2cm}
    \caption{Ratio R for different taxonomies - DS-Llama8B}
    \label{fig:reason_phase_alternative_taxonomies}
    \vspace{-0.2cm}
\end{figure}

\textbf{Ratio $R$ for each taxonomy.} Figure \ref{fig:reason_phase_alternative_taxonomies} presents the ratio for the different alternative taxonomies. Across both GSM8K and MATH500 datasets, we observe a that transition pattern is common over the different levels of abstractions. Before $S_{\text{IES}}$, the ratio remains relatively high, indicating that constructive steps dominates the reasoning process. After $S_{\text{IES}}$, the ratio drops sharply, reflecting an increasing proportion of evaluative steps.

It is worth noting that the magnitude of R before $S_{\text{IES}}$ is lower for labels $4$ and $3$, and the drop is less pronounced for 2-labels after $S_{\text{IES}}$. In comparison, $R$ is higher and its drop is cleaner for Original and 6-labels taxonomies. 

\begin{figure}[h]
    \centering
    \begin{subfigure}{0.49\textwidth}
        \includegraphics[width=\linewidth]{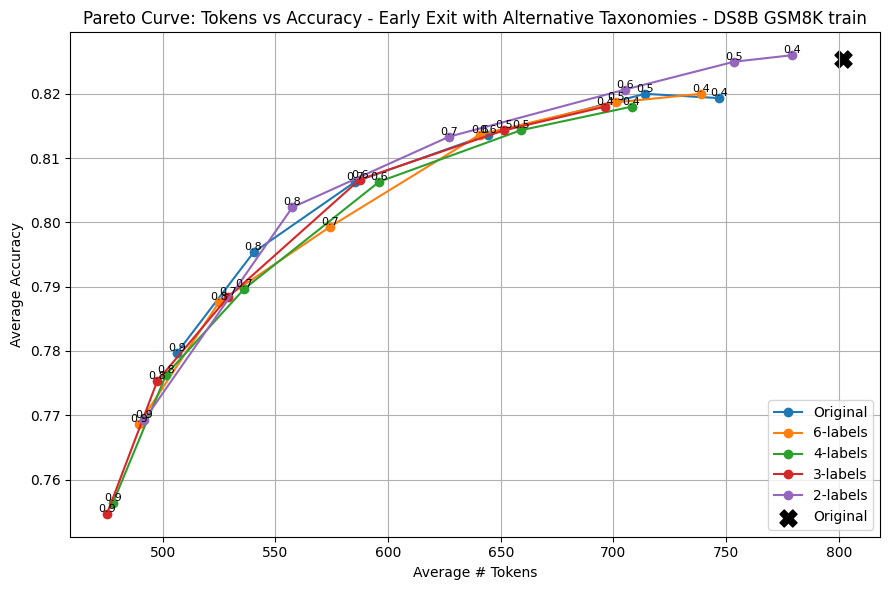}
        \caption{GSM8K}
        \label{fig:GSM8K_alt_taxonomies}
    \end{subfigure}
    \hfill
    \begin{subfigure}{0.49\textwidth}
        \includegraphics[width=\linewidth]{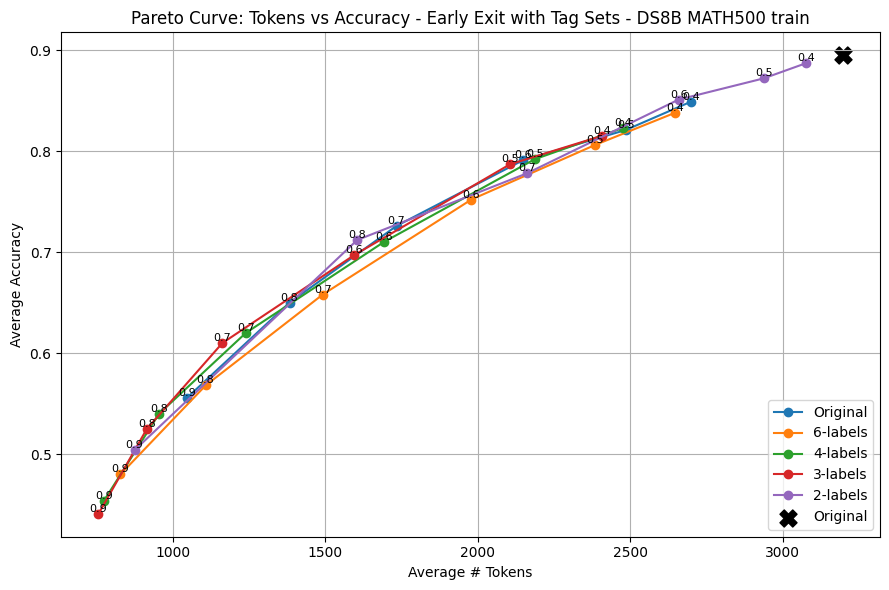}
        \caption{MATH500}
        \label{fig:MATH500_alt_taxonomies}
    \end{subfigure}
    \vspace{-0.2cm}
    \caption{TRACES performance against alternative taxonomies - DS-Llama8B}
    \label{fig:st_es_alt_taxonomies}
    \vspace{-0.2cm}
\end{figure}

\textbf{TRACES performance per taxonomy.} Figure \ref{fig:st_es_alt_taxonomies} compares the TRACES performance obtained using the different taxonomies. The Pareto curves, illustrating the token-count usage and the accuracy, remain very similar across taxonomies for both GSM8K and MATH500. The original and the $6$-labels taxonomies achieve slightly better trade-offs, providing higher accuracy for the for the same token budget. However, the performance degradation is minimal for higher-level taxonomies. This suggests that TRACES does not rely highly on fine-grained reasoning distinctions to identify the reasoning transition.

\textbf{Takeaways.} To conclude, we observe that the reasoning transition pattern remains robust across different level of granularity of the taxonomy. While finer-grained taxonomies seem to enable cleaner signal, all taxonomies resulted in equivalent TRACES performances.

\newpage

\subsection{Performance of the Step-Taggers} \label{sec:appendix-infl-fact-perf-step-tagg}


\textbf{Objective.} In this ablation study, we assess how robust our framework is with regards to the Step-Tagging lightweight module. To do so, we run the experiment that we conducted on different expected performance of the Step-Tagging module. 

\textbf{Motivation.} Even though we obtained satisfying performances, it remains unclear how the performance of the BERT classifiers affects the inference outcomes. To mitigate this limitation, this section looks at how tagging errors affects the efficiency performance of our TRACES framework.

\textbf{Claim.} The goal of our Step-Tagger module is to assign each step one of the following 3 tag: 0 - neural, 1 - constructive and 2 - evaluative. An inaccurate step-tagger would miss to identify the correct class. Intuitively, such errors in the classifications would result in rendering the monitoring process noisy. Indeed, our approach is based on the observation that constructive classes appears more frequently before the IES step, and evaluative classes after. We claim that an inaccurate step-tagger would therefore degrade the identification of the reasoning transition phase.

\textbf{Methodology.} To verify such claim, we selected the \deepseekLlama{} model, on GSM8K and MATH500 train. As we conducted the same methodology on other models, we expect similar results for other models.

\begin{figure}[h]
    \centering
    \begin{subfigure}{0.49\textwidth}
        \includegraphics[width=\linewidth]{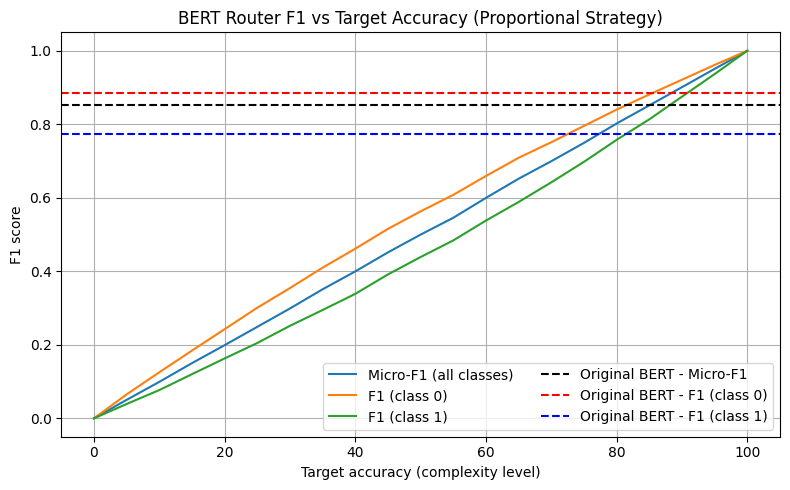}
        \caption{GSM8K}
        \label{fig:GSM8K_simulated_bert_performance}
    \end{subfigure}
    \hfill
    \begin{subfigure}{0.49\textwidth}
        \includegraphics[width=\linewidth]{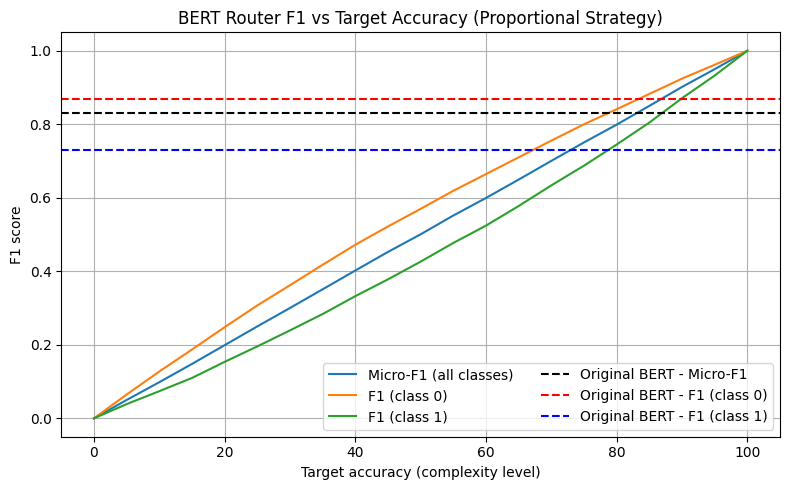}
        \caption{MATH500}
        \label{fig:MATH500_simulated_bert_performance}
    \end{subfigure}
    \vspace{-0.2cm}
    \caption{Simulated performance of the BERT classifiers} 
    \label{fig:simulated_bert_performance}
    \vspace{-0.2cm}
\end{figure}

\textbf{Proportional performance:} To ground our analysis more realistically, we introduce a proportionally accurate tagging module, that would get prediction correct following the ground-truth label distribution (from \GPTmini{} annotation process). Indeed, step-types have different distributions, so a balanced tagging module would disproportionally impact the overall metric. We defined $10$ target accuracy from the router, from $0.10$ to $1.00$ micro-F1, with a path of $0.10$. Figure \ref{fig:simulated_bert_performance} describes the strategy we adopted for both datasets, along with the original router accuracy (see Figure \ref{fig:step-tagging-performance} in Section \ref{sec:st-es-evaluation}). 

\begin{figure}[h]
    \centering
    \begin{subfigure}{0.49\textwidth}
        \includegraphics[width=\linewidth]{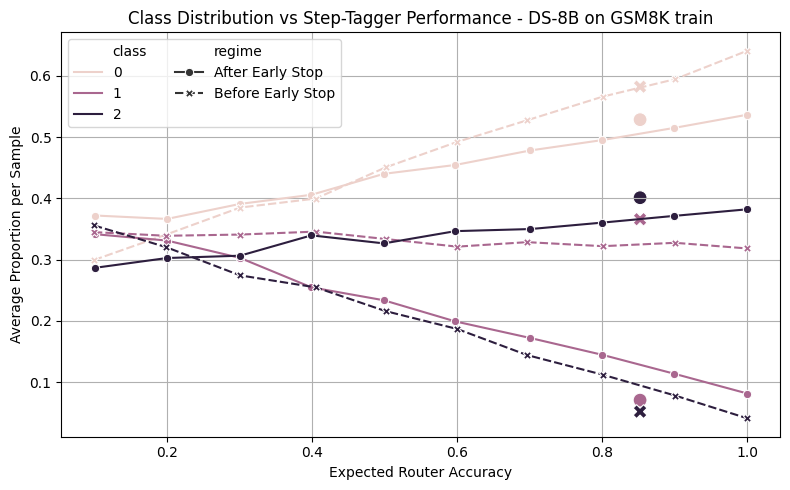}
        \caption{GSM8K}
        \label{fig:GSM8K_distribution_bert_performance}
    \end{subfigure}
    \hfill
    \begin{subfigure}{0.49\textwidth}
        \includegraphics[width=\linewidth]{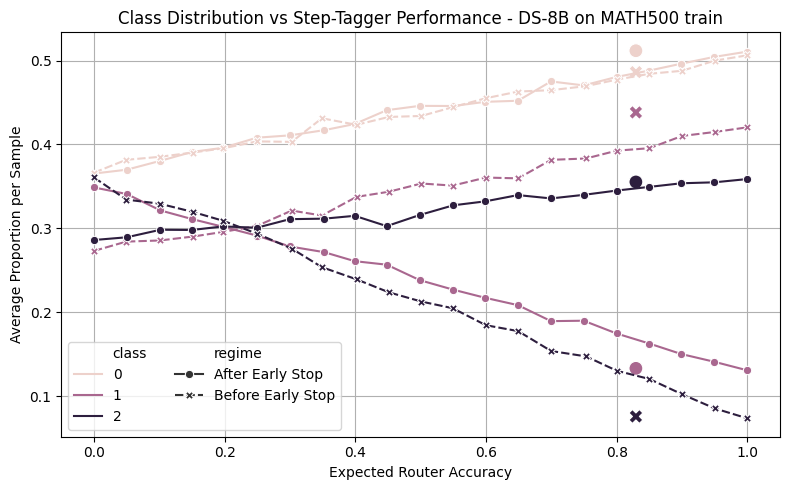}
        \caption{MATH500}
        \label{fig:MATH500_distribution_bert_performance}
    \end{subfigure}
    \vspace{-0.2cm}
    \caption{Distribution $S_{\text{before}}$ and $S_{\text{after}}$ against step-tagger's performance}
    \label{fig:distribution_bert_performance}
    \vspace{-0.2cm}
\end{figure}

\textbf{Distribution before after based on expected router accuracy.} First, we evaluate the distribution of tags generated by the BERT classifiers, before and after the IES step. Figure \ref{fig:distribution_bert_performance} shows the average proportion of the three classes (0 - neutral, 1 - constructive, and 2 - evaluative) against the simulated module's performance. We observe that as the tagger's performance degrades to smaller values of accuracy, the proportion curves converges towards equivalent values. It means that the tagger's performance affects its capacity to identify the pattern that we observed in Section \ref{sec:influence-step-type} (unbalanced distribution of specific classes depending on the state of the reasoning - before or after the IES step).

\begin{figure}[h]
    \centering
    \begin{subfigure}{0.49\textwidth}
        \includegraphics[width=\linewidth]{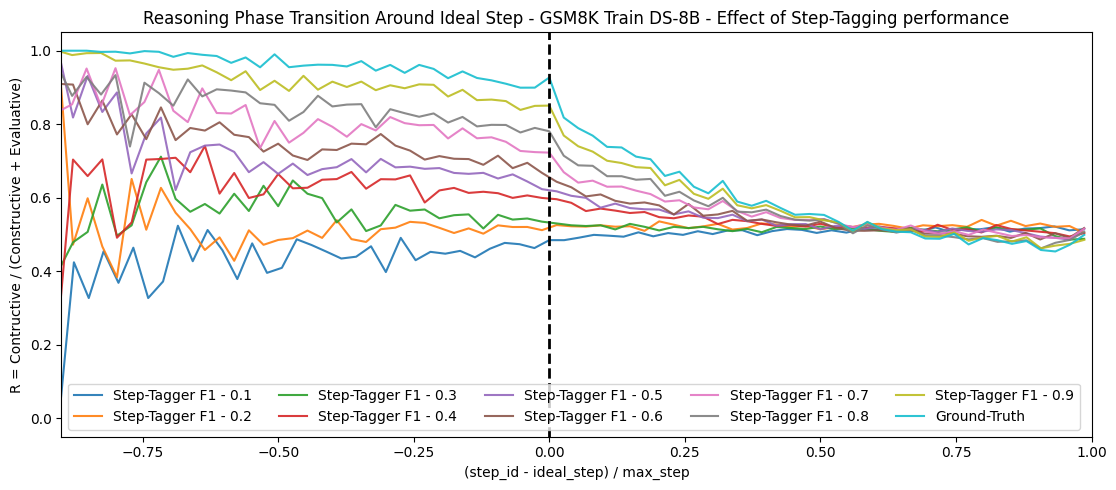}
        \caption{GSM8K}
        \label{fig:GSM8K_ratio_router}
    \end{subfigure}
    \hfill
    \begin{subfigure}{0.49\textwidth}
        \includegraphics[width=\linewidth]{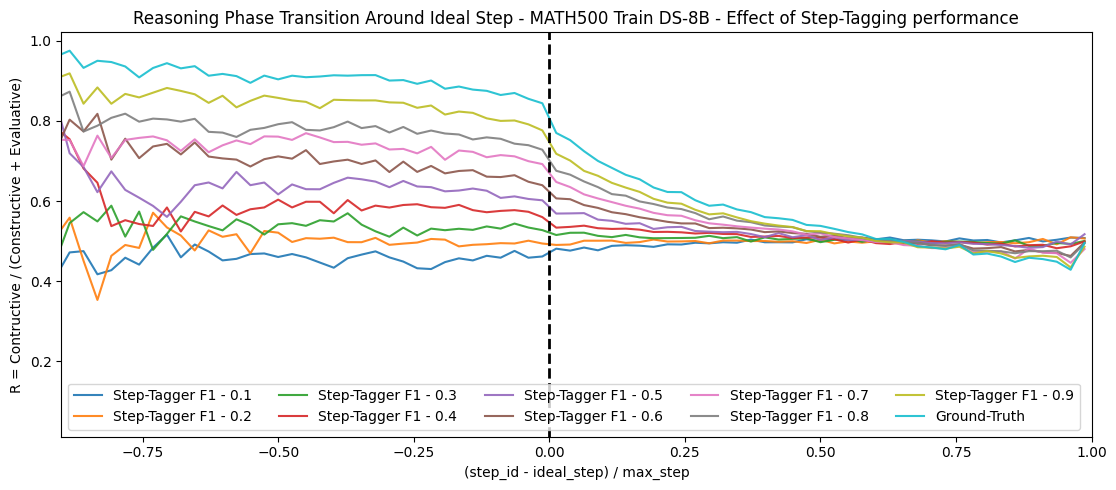}
        \caption{MATH500}
        \label{fig:MATH500_ratio_router}
    \end{subfigure}
    \vspace{-0.2cm}
    \caption{Ratio $R$ against the step-tagger's performance}
    \label{fig:ratio_bert_performance_eval}
    \vspace{-0.2cm}
\end{figure}

\textbf{Ratio $R$ based on expected router accuracy.} Figure \ref{fig:ratio_bert_performance_eval} shows the ratio R for the different values of performance of our tagger module. The Figure confirms our observation from the distribution (Figure \ref{fig:distribution_bert_performance}). As the performance of the Step-Tagger module degrades, the ratio becomes noisy, and the transition phase is less sharp. This is due to the distribution of tags itself. Indeed, we observed in Figure \ref{fig:ratio_bert_performance_eval} that for inaccurate modules (F1 $<$ 0.4), the distributions of each class before and after the IES step is comparable, and so the resulting ratio is almost constant, without drop after the IES.

\begin{figure}[h]
    \centering
    \begin{subfigure}{0.49\textwidth}
        \includegraphics[width=\linewidth]{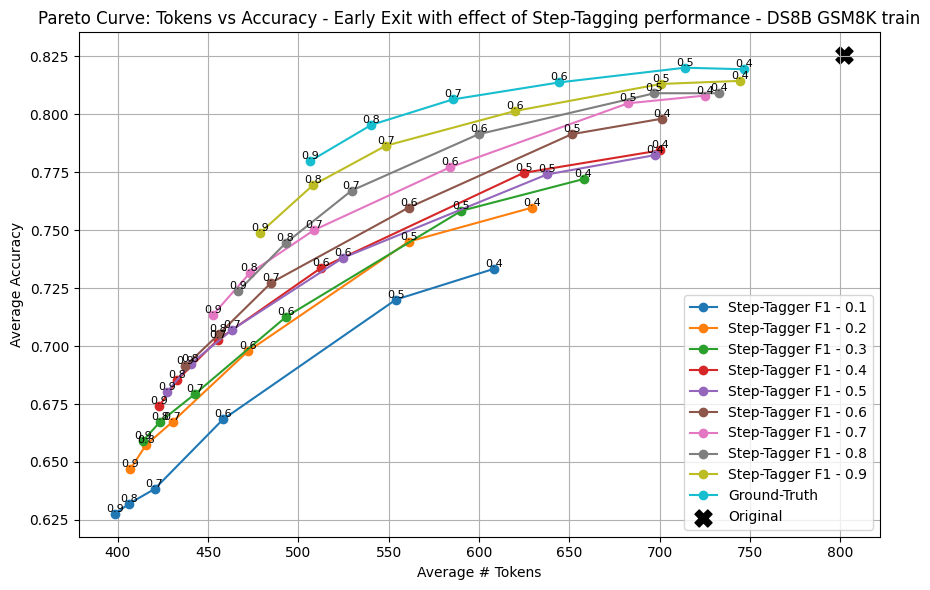}
        \caption{GSM8K}
        \label{fig:GSM8K_st_es_router_performance}
    \end{subfigure}
    \hfill
    \begin{subfigure}{0.49\textwidth}
        \includegraphics[width=\linewidth]{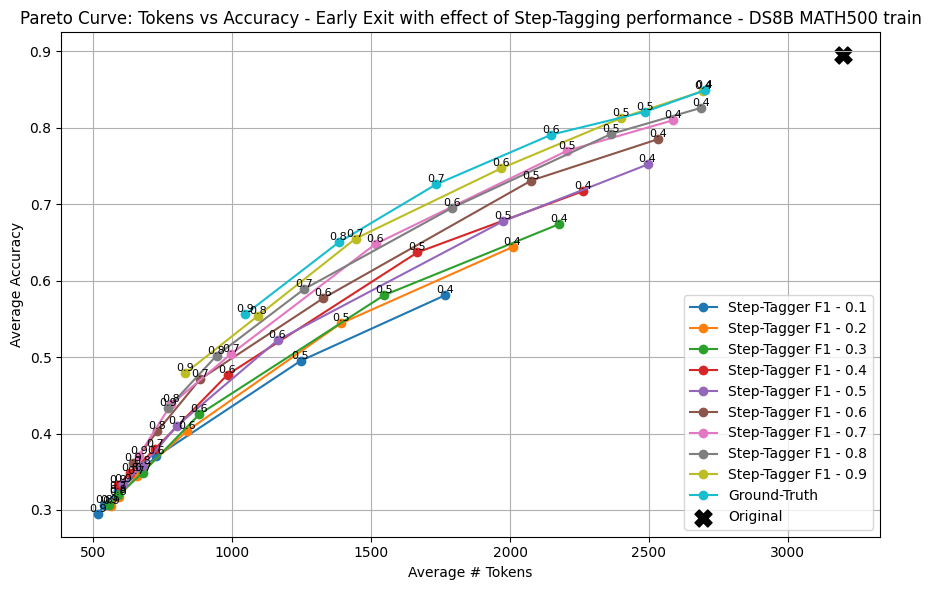}
        \caption{MATH500}
        \label{fig:MATH500_st_es_router_performance}
    \end{subfigure}
    \vspace{-0.2cm}
    \caption{TRACES against the step-tagger's performance}
    \label{fig:ratio_bert_performance}
    \vspace{-0.2cm}
\end{figure}

\textbf{TRACES based on expected router accuracy.} On the strength of our analysis, we applied the classifiers to our TRACES framework. In Figure \ref{fig:ratio_bert_performance}, each curve represent the TRACES obtained for different level of performance of the step-taggers. We note that the ground-truth obtained the best performance (forming a Pareto front - best accuracy for a given token-count). As the router's performance degrades, the accuracy of TRACES drops.

\textbf{Takeaways.} Overall, we confirm our claims. When the Step-Tagging module performs poorly (F1 $< 0.4$), the distribution of tags before and after IES becomes undistinguishable. As a consequence, the ratio $R$ loses its  drop at the IES step and no transition phase is observable. It makes it harder to identify the correct step to early-stop.

However, beyond a certain threshold (F1 $> 0.7$), the TRACES curves remain close to the ground-truth Pareto front, suggesting that the framework is robust to minor tag errors. This is important, as it implies that a lightweight sentence classifier with reasonable (but not perfect) performance is sufficient to preserve the gains of our approach.

\newpage

\subsection{Cross-model transferability of the Step-Taggers} \label{sec:appendix-classifiers-new-models}

The cross-model transferability of the step classifier is important for a broader applicability of TRACES to different reasoning models. For this reason, we trained a unique classifier on reasoning traces from a single model (DS-Qwen14B - Qwen2.5-14B base model) on MATH500 and GPQA training datasets. Figure \ref{fig:step-tagging-performance} in Section \ref{sec:st-es-evaluation} (see main body) shows a good transferability of our classifier to other models (DS-Llama8B - Llama-3.1-8B base model, and QwQ-32B - Qwen-2.5-32B base model with a different RLHF training). As well, the module transfers well on different datasets (GSM8K, MMLU-Pro, and AIME).

Furthermore, we ran additional evaluations to confirm the cross-model transferability. We selected three additional models:

\begin{itemize}
    \item \textbf{DeepSeek-R1-0528-Qwen3-8B}, a R1-Distilled model based on a more recent model architecture (i.e. Qwen3 rather than Qwen2.5).

    \item \textbf{Qwen3-8B}, a reasoning model with a more recent architecture, but also obtained with a different training procedure than R1-Distilled models.

    \item \textbf{GPT-OSS-120B}, a much larger reasoning model from a different provider, and also trained with a different training procedure than R1-Distilled models.
\end{itemize}

We inferred reasoning traces on MATH500 and GPQA-Diamond test datasets, and annotated the reasoning traces obtained to establish the ground-truth labels. We then ran our classifier on the annotated reasoning steps. As in Figure \ref{fig:step-tagging-performance} (Section \ref{sec:st-es-evaluation}), Table \ref{tab:performance_step_tagger_cross_model_general} presents the performance of our lightweight classifier on the reasoning traces of these two additional models.

\begin{table}[h]
\centering
\tiny
\begin{tabular}{l
                ccc
                ccc}
\toprule
\multirow{2}{*}{\textbf{Metrics}}  
& \multicolumn{3}{c}{\textbf{MATH500}} 
& \multicolumn{3}{c}{\textbf{GPQA-Diamond}} \\
\cmidrule(lr){2-4} \cmidrule(lr){5-7}
& Qwen3-8B & DS-Qwen3-8B & gpt-oss-120b & Qwen3-8B & DS-Qwen3-8B & gpt-oss-120b \\
\midrule

Class 0 F1 (Others) & 0.86 & 0.83 & 0.85 & 0.86 & 0.82 & 0.79 \\
Class 1 F1 (Constructive) & 0.74 & 0.69 & 0.72 & 0.77 & 0.75 & 0.70 \\
Class 2 F1 (Evaluation) & 0.83 & 0.70 & 0.79 & 0.75 & 0.50 & 0.70 \\

\midrule

Macro-F1 & 0.81 & 0.74 & 0.78 & 0.79 & 0.69 & 0.73 \\
Micro-F1 & 0.83 & 0.77 & 0.80 & 0.82 & 0.77 & 0.74 \\

\bottomrule
\end{tabular}%
\caption{Performance of step-classifier on new models}
\label{tab:performance_step_tagger_cross_model_general}
\end{table}

\textbf{Takeaways.} We observe a good transferability of our trained classifier on a models having a different architectures and training methods. Indeed, Macro-F1 scores range from 0.73 to 0.81, confirming the transferability across different architecture and RLHF training methods.

\newpage

\subsection{Summary of Takeaways} \label{sec:summary-TRACES}

In this Section, we evaluated potential factors of influence of TRACES. Our ablation studies enable us to better assess the robustness, of our framework. Our takeaways are as follow:

\begin{enumerate}

    \item \textbf{Flexibility and Zero-Shot adaptive computation:} We showed that TRACES allows to manage the computation precisely through its parameter $\delta$. Across models and datasets, we could expect specific token-count saving corresponding to values of $\delta$. Furthermore, our framework does not requires any knowledge on the dataset or model. The training is a one-time exercise, and TRACES does not requires calibration (as opposed to existing early-exit methods).

    \item \textbf{Selected classes matter to compute for ratio $R$:} To compute the ratio $R$, we selected specific step-types to form \textbf{Constructive} and \textbf{Evaluative} classes. We observed that the selection does have an impact on the performance of the framework. To achieve better results, the classes should carry specific information about the completeness of the model towards the correct answer. Specifically, steps included in the \textbf{Constructive} class should appear mainly before the IES step. Conversely, steps included in the \textbf{Evaluative} class should be very frequent after the IES step.

    \item \textbf{TRACES robust to the Taxonomy:} We showed that the granularity of classes does not strongly affects the performance of our framework. While finer-grained taxonomy better help to identify specific reasoning behavior, TRACES still work relatively well using taxonomies including smaller number of step-types.

    \item \textbf{Performance of Step-Taggers:} Further, we showed that the performance of the step-tagging modules affects the effectiveness of TRACES. Importantly, we showed that relatively accurate classifiers (F1 $>$ $0.7$) still results in satisfying performance of TRACES.

    \item \textbf{Performance of Step-Taggers:} Finally, we demonstrated the cross-model transferability of our step-tagging module on three additional models of different architecture and training paradigms, across two reasoning benchmarks.
    
\end{enumerate}

\newpage

\section{Analysis of the cost of TRACES}

\subsection{Latency analysis of the Lightweight step-tagging module} \label{sec:appendix-latency-analysis}

In our experimentation, we reported the number of tokens generated by the models as a proxy for the resources and latency of the inferences of the models. Indeed, we performed the experimentation offline and the number of tokens could have been computed without the need of re-running the inferences. 

\textbf{Motivations.} By definition, our early-stopping criteria requires an external module to perform online annotation of the generated reasoning steps. To further validate our approach, this sub-section aims to analyze the latency introduced by our lightweight step classifiers.

\textbf{Methodology.} Our experiments have been performed offline. However, we have access to both standard runtime, number of tokens, and the number of tokens of the early-stopped samples. In addition, we recorded the runtime of the lightweight classifiers when annotating each steps. To allow fair comparison of the inference latency of the early-stopped samples, we need to estimate the early-stopped runtime.

The runtime of the early-stopped samples is composed of three components:

\begin{equation}
r_{\text{TRACES}} = r_{\text{stopped}} + r_{\text{step\_classifier}} + r_{\text{completion}}
\end{equation}

where $r_{\text{TRACES}}$ is the total runtime of the early-stopped sample, $r_{\text{stopped}}$ is the runtime of the model generating tokens from the start to the early-stopping condition, $r_{\text{step\_classifier}}$ is the runtime of the lightweight classifier annotating each reasoning step, and $r_{\text{completion}}$ is the additional runtime of the model when prompted to generate its current best answer (see Early-Stopping criteria and Answer Forcing in Section \ref{sec:st-es-evaluation}).

\begin{figure}[h] 
    \centering
    \begin{minipage}{0.3\linewidth}
        \includegraphics[width=\linewidth]{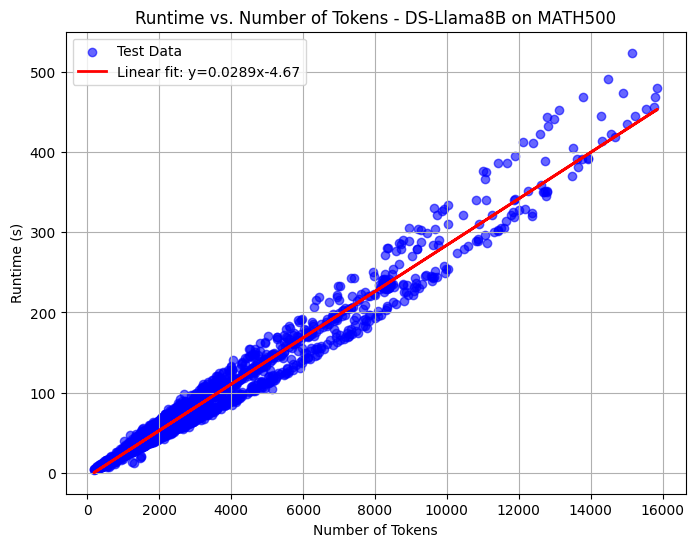}
        \subcaption{\small DS-Llama8B}
        \label{fig:latency_ds_llama8b}
    \end{minipage}
    \hfill
    \begin{minipage}{0.3\linewidth}
        \includegraphics[width=\linewidth]{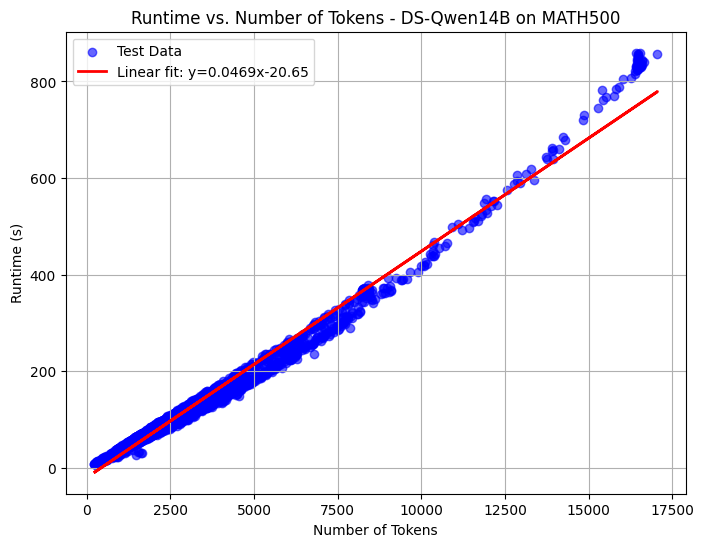}
        \subcaption{\small DS-Qwen14B}
        \label{fig:latency_ds_qwen14b}
    \end{minipage}
    \hfill
    \begin{minipage}{0.3\linewidth}
        \includegraphics[width=\linewidth]{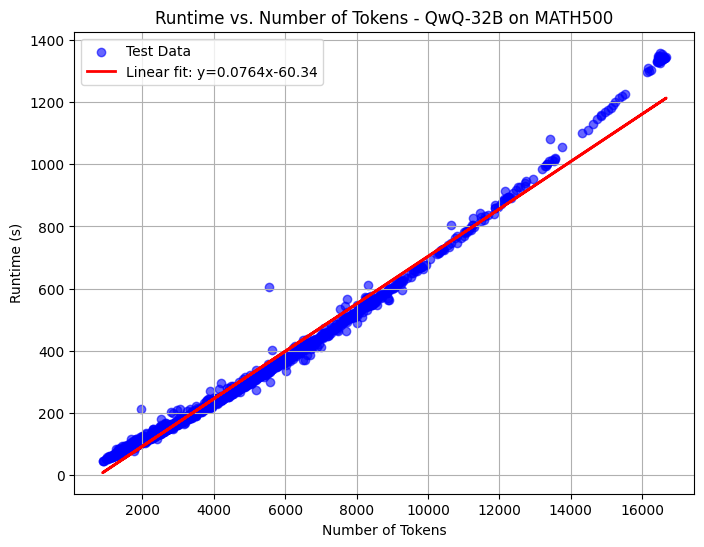}
        \subcaption{\small QwQ-32B}
        \label{fig:latency_qwq32b}
    \end{minipage}
    \vspace{-0.2cm}
    \caption{Linear relationship between number of tokens and runtime}
    \label{fig:linear assumption}
\end{figure}

\textbf{Linear Assumption.} While $r_{\text{step\_classifier}}$ and $r_{\text{completion}}$ are accessible though our computations, we need to estimate $r_{\text{stopped}}$ since we ran our experimentation offline. In the literature, researchers seems to acknowledge a linear relationship between the runtime and the number of token generated: \emph{``During the AR [Auto-Regressive] decoding, output tokens are generated sequentially, conditioned on all previously generated tokens. As a result, the decoding time (i.e. inference latency) increases linearly with the decoding length.''} \citep{oh2022improvingtopkdecodingnonautoregressive}. This assumption is further validated by our experimentation. Indeed, Figure \ref{fig:linear assumption} reports linear regressions between the number of tokens and the runtime of the three selected LRMs, on the MATH500 over the 5 seeds we used.

\textbf{Estimation of $r_{\text{stopped}}$.} Assuming linear relationship between number of tokens generated and the runtime, we estimate $r_{\text{stopped}}$ by fitting a linear regression between the number of generated tokens and the observed runtime of the full runs (see Figure \ref{fig:linear assumption}). 

\begin{equation}
r_{\text{stopped}} \approx \alpha \cdot N_{\text{TRACES}} + \beta 
\end{equation}

where $N_{\text{tokens}}$ is the number of tokens generated in a full run. For each early-stopped samples, we predict its runtime using the regression model. 

\begin{table}[h]
\tiny
\centering
\begin{tabular}{lcccccccc}
\toprule
\textbf{Dataset} & \textbf{Models} & \textbf{$r_{\text{base}}$ (s)} & \textbf{Config.} & \textbf{$r_{\text{stop}}$ (s)} & \textbf{$r_{\text{class.}}$ (s)} & \textbf{$r_{\text{compl.}}$ (s)} & \textbf{$r_{\text{TRACES}}$ (s)} & \textbf{Speed-up} ($\uparrow$) \\ 

\midrule

\multirow{20}{*}{MATH500} & \multirow{6}{*}{DS-8B} & \multirow{6}{*}{102.32} & TRACES ($\delta - 0.4$) & 83.82 & 0.62 & 5.55 & 89.99 & $\times$1.14 \\
& & & TRACES ($\delta - 0.5$) & 73.93 & 0.55 & 4.67 & 79.15 & $\times$1.29 \\
& & & TRACES ($\delta - 0.6$) & 59.83 & 0.45 & 4.16 & 64.45 & $\times$1.59 \\
& & & TRACES ($\delta - 0.7$) & 43.96 & 0.33 & 4.48 & 48.77 & $\times$2.09 \\
& & & TRACES ($\delta - 0.8$) & 30.08 & 0.23 & 3.77 & 34.08 & $\times$3.00 \\
& & & TRACES ($\delta - 0.9$) & 19.28 & 0.15 & 4.89 & 24.32 & $\times$4.21 \\

\cmidrule(lr){2-9}

& \multirow{6}{*}{DS-14B} & \multirow{6}{*}{138.25} & TRACES ($\delta - 0.4$) & 114.32 & 0.49 & 4.70 & 119.51 & $\times$1.16 \\
& & & TRACES ($\delta - 0.5$) & 103.89 & 0.46 & 5.35 & 109.70 & $\times$1.26 \\
& & & TRACES ($\delta - 0.6$) & 81.63 & 0.38 & 5.01 & 87.02 & $\times$1.59 \\
& & & TRACES ($\delta - 0.7$) & 59.69 & 0.29 & 4.82 & 64.79 & $\times$2.13 \\
& & & TRACES ($\delta - 0.8$) & 41.85 & 0.21 & 4.78 & 46.84 & $\times$2.95 \\
& & & TRACES ($\delta - 0.9$) & 25.72 & 0.14 & 5.11 & 30.97 & $\times$4.46 \\

\cmidrule(lr){2-9}

& \multirow{6}{*}{QwQ-32B} & \multirow{6}{*}{281.51} & TRACES ($\delta - 0.4$) & 199.04 & 0.53 & 5.66 & 205.22 & $\times$1.37 \\
& & & TRACES ($\delta - 0.5$) & 165.51 & 0.46 & 12.46 & 178.43 & $\times$1.58 \\
& & & TRACES ($\delta - 0.6$) & 115.92 & 0.36 & 12.22 & 128.49 & $\times$2.19 \\
& & & TRACES ($\delta - 0.7$) & 70.91 & 0.26 & 5.44 & 76.61 & $\times$3.67 \\
& & & TRACES ($\delta - 0.8$) & 35.79 & 0.17 & 5.39 & 41.35 & $\times$6.81 \\
& & & TRACES ($\delta - 0.9$) & 11.65 & 0.11 & 5.59 & 17.35 & $\times$16.23 \\

\midrule

\multirow{20}{*}{GSM8K} & \multirow{6}{*}{DS-8B} & \multirow{6}{*}{25.83} & TRACES ($\delta - 0.4$) & 21.31 & 0.11 & 3.22 & 24.63 & $\times$1.05 \\
& & & TRACES ($\delta - 0.5$) & 19.95 & 0.09 & 5.08 & 25.13 & $\times$1.03 \\
& & & TRACES ($\delta - 0.6$) & 17.57 & 0.08 & 3.46 & 21.11 & $\times$1.22 \\
& & & TRACES ($\delta - 0.7$) & 15.19 & 0.07 & 3.49 & 18.76 & $\times$1.38 \\
& & & TRACES ($\delta - 0.8$) & 13.48 & 0.06 & 2.92 & 16.46 & $\times$1.57 \\
& & & TRACES ($\delta - 0.9$) & 12.32 & 0.05 & 3.83 & 16.19 & $\times$1.59 \\

\cmidrule(lr){2-9}

& \multirow{6}{*}{DS-14B} & \multirow{6}{*}{27.74} & TRACES ($\delta - 0.4$) & 23.26 & 0.05 & 4.32 & 27.63 & $\times$1.00 \\
& & & TRACES ($\delta - 0.5$) & 22.58 & 0.05 & 4.15 & 26.78 & $\times$1.04 \\
& & & TRACES ($\delta - 0.6$) & 21.31 & 0.05 & 4.32 & 25.67 & $\times$1.08 \\
& & & TRACES ($\delta - 0.7$) & 20.14 & 0.04 & 4.47 & 24.65 & $\times$1.13 \\
& & & TRACES ($\delta - 0.8$) & 19.40 & 0.04 & 4.27 & 23.71 & $\times$1.17 \\
& & & TRACES ($\delta - 0.9$) & 18.71 & 0.03 & 4.37 & 23.12 & $\times$1.20 \\

\cmidrule(lr){2-9}

& \multirow{6}{*}{QwQ-32B} & \multirow{6}{*}{117.28} & TRACES ($\delta - 0.4$) & 91.22 & 0.22 & 4.88 & 96.32 & $\times$1.22 \\
& & & TRACES ($\delta - 0.5$) & 81.78 & 0.19 & 5.04 & 87.02 & $\times$1.35 \\
& & & TRACES ($\delta - 0.6$) & 63.42 & 0.15 & 11.28 & 74.84 & $\times$1.57 \\
& & & TRACES ($\delta - 0.7$) & 47.72 & 0.12 & 5.23 & 53.07 & $\times$2.21 \\
& & & TRACES ($\delta - 0.8$) & 38.39 & 0.09 & 5.26 & 43.75 & $\times$2.68 \\
& & & TRACES ($\delta - 0.9$) & 30.51 & 0.08 & 5.41 & 36.01 & $\times$3.26 \\

\bottomrule
\end{tabular}
\caption{Latency analysis of the TRACES framework - For each datasets and models, we reported the average runtime per sample, across the 5 seeds. Each runtime $r$ is expressed in seconds. We reported the standard runtime $r_{\text{base}}$ corresponding to the runtime during standard inference. The runtime $r_{\text{TRACES}}$ is the sum of the runtime corresponding to stopped inference ($r_{\text{stop}}$), the step classifier ($r_{\text{class.}}$), and the final completion ($r_{\text{compl.}}$). The runtime Speed-up is computed between $r_{\text{standard}}$ and $r_{\text{TRACES}}$.}
\label{tab:latency-analysis}
\end{table}

\newpage

\textbf{Latency analysis.} Table \ref{tab:latency-analysis} reports the latency analysis of the experiments. The runtime analysis demonstrates a speed-up ranging from \textbf{1.2} to \textbf{3} for configurations offering limited accuracy loss, meaning that the framework leads to faster model inference for every configurations, compared to the standard inference. Importantly, the Table shows a \emph{very low runtime of the step classifiers} (around 0.01-0.02s per reasoning steps), being two magnitudes lower than the auto-regressive generation of the models. This is because we employed BERT classifiers. It is worth noting that the final completion have more impact on the overall runtime, but is still limited compared to the model's generation runtime since we only allowed a specific token budget (additional $100$ tokens).

\textbf{Online latency.} Even though we showed that the TRACES framework reduce the latency of the model's generation, we suspect that an online implementation of the framework could result in higher latency. Indeed, although the step classification itself is fast and does not hurt the efficiency gains, our framework requires to pause the model's generation frequently to perform step classification. To this means, this process can introduce additional latency as suggested by \cite{yang2025dynamicearlyexitreasoning}. However, their findings suggest that this additional latency should be limited. 

\textbf{Takeaway.} This section present a latency analysis of our TRACES framework. The latency introduced by the classification process of the BERT models appears to be minimal compared to the generation's latency of the reasoning models. Future work should look at comparing runtime of offline and online implementations to validate our findings.

\newpage

\subsection{Online latency of TRACES} \label{sec:appendix-online-latency}

In this section, we implemented an online version of the TRACES algorithm (see Algorithm \ref{alg:st_es_algorithm} in Appendix \ref{sec:appendix-es_algorithm}). To do so, we relied on the Algorithm \ref{alg:stepwise_generation} (see Appendix \ref{sec:appendix-es_algorithm}), where we produce the reasoning traces step-by-step to monitor the generation after each step. We generated reasoning traces token-by-token and did the parsing and monitoring online. We inferred our three selected models (DS-8B, DS-14B and QwQ-32B) on MATH500 test using seed 42, and compared the latency and token/s rate. Table \ref{tab:online-efficiency-TRACES} below presents such analysis.

\begin{table}[h]
    \tiny
    \centering
    \begin{tabular}{ccccccc}
        \toprule
        \textbf{Model} & \textbf{Setting} & \textbf{Avg Runtime (s)} & \textbf{Median Runtime (s)} & \textbf{p95 Runtime (s)} & \textbf{Avg Tokens} & \textbf{Avg Tok/s} \\
        \midrule
        \multirow{6}{*}{DS-8B} & Baseline & 95.7  & 55.9  & 349.6 & 3810 & 43.0 \\
         & TRACES $\delta$=0.5   & 63.0 \tiny{($-34\%$)}  & 39.6  & 191.5 & 2768 \tiny{($-27\%$)} & 41.8 \\
         & TRACES $\delta$=0.6   & 78.1 \tiny{($-18\%$)}  & 48.9  & 275.8 & 2335 \tiny{($-39\%$)} & 28.2 \\
         & TRACES $\delta$=0.7   & 42.7 \tiny{($-55\%$)}  & 26.4  & 159.3 & 1935 \tiny{($-49\%$)} & 42.2 \\
         & TRACES $\delta$=0.8   & 37.6 \tiny{($-61\%$)}  & 25.8  & 125.3 & 1539 \tiny{($-60\%$)} & 37.6 \\
         & TRACES $\delta$=0.9   & 26.6 \tiny{($-72\%$)}  & 21.2  &  52.7 & 1169 \tiny{($-69\%$)} & 39.8 \\
         \midrule
         \multirow{6}{*}{DS-14B}
         & Baseline              & 128.8 & 77.2  & 422.4 & 3361 & 28.6 \\
         & TRACES $\delta$=0.5   & 94.2 \tiny{($-27\%$)}  & 57.4  & 306.3 & 2567 \tiny{($-24\%$)} & 25.9 \\
         & TRACES $\delta$=0.6   & 79.7 \tiny{($-38\%$)}  & 48.8  & 282.6 & 2123 \tiny{($-37\%$)} & 25.1 \\
         & TRACES $\delta$=0.7   & 77.8 \tiny{($-40\%$)}  & 49.1  & 277.6 & 1757 \tiny{($-48\%$)} & 21.0 \\
         & TRACES $\delta$=0.8   & 59.3 \tiny{($-54\%$)}  & 40.8  & 169.2 & 1425 \tiny{($-58\%$)} & 22.1 \\
         & TRACES $\delta$=0.9   & 45.3 \tiny{($-65\%$)}  & 37.3  & 100.8 & 1090 \tiny{($-68\%$)} & 21.6 \\
         \midrule
         \multirow{6}{*}{QwQ-32B}
         & Baseline              & 296.47 & 190.3 & 894.08 & 4673 & 17.3 \\
         & TRACES $\delta$=0.5   & 174.9 \tiny{($-41\%$)}  & 105.2  & 446.1 & 2819 \tiny{($-40\%$)} & 17.6 \\
         & TRACES $\delta$=0.6   & 119.5 \tiny{($-59\%$)} & 76.9 & 441.6 & 2305 \tiny{($-50\%$)} & 18.3 \\
         & TRACES $\delta$=0.7   & 94.71 \tiny{($-68\%$)}  & 58.4 & 343.8 & 1876 \tiny{($-60\%$)} & 18.7 \\
         & TRACES $\delta$=0.8   & 72.77 \tiny{($-75\%$)}  & 51.5  & 209.6 & 1489 \tiny{($-68\%$)} & 19.0 \\
         & TRACES $\delta$=0.9   & 57.07 \tiny{($-80\%$)}  & 45.7  & 141.3 & 1201 \tiny{($-74\%$)} & 19.2 \\
\bottomrule
    \end{tabular}
    \caption{Inference efficiency of online implementation of TRACES on MATH500 across thresholds $\delta$. Percentages show reduction relative to the baseline. Seed 42.}
    \label{tab:online-efficiency-TRACES}
\end{table}

We observe that the monitoring and step-by-step generation introduces some latency (around 10 to 30\% lower tok/s rate compared to baseline for 8B and 14B models). Importantly, the latency caused by the monitoring does not affect the overall algorithm. Indeed, we observe that the saved average runtime of same range of saved token-count. This additional experiment confirms our online estimations (see Appendix \ref{sec:appendix-latency-analysis}). TRACES introduces some latency in the generation, but the algorithm itself recovers the latency costs.

\newpage

\subsection{Computational overhead of TRACES} \label{sec:appendix-computational-overhead}

Further, we analyze the computational overhead, in term of FLOPs and latency, introduced by the step classifier. First, we follow the approximation from \cite[Table 1, Eq. 2.2]{kaplan2020scalinglawsneurallanguage} to estimate the FLOPs. For a large language model, the per-step computational cost can be approximated as:

\begin{equation}
C_{\text{forward}} \approx (2N_{\text{LRM}} + 2\times n_{\text{layer}}\times n_{\text{ctx}}\times d_{\text{att}}) \times L_{\text{step}}    
\end{equation}

And for the BERT-based step classifier:

\begin{equation}
C_{\text{forward}} \approx (2N_{\text{BERT}} + 2\times n_{\text{layer}}\times n_{\text{ctx}}\times d_{\text{att}}) \times L_{\text{step}}  
\end{equation}

However, \(N_{\text{LRM}} >> n_{\text{layer}}\times n_{\text{ctx}}\times d_{\text{att}}\) and \(N_{\text{BERT}} >> n_{\text{layer}}\times n_{\text{ctx}}\times d_{\text{att}}\), so we can estimate that the computational overhead is roughly: \(\frac{C_{\text{forward BERT}}}{C_{\text{forward LRM}}} \approx \frac{N_{\text{BERT}}}{N_{\text{LRM}}}\)

Table \ref{tab:flops_overhead} shows the estimated computational overhead per sample computed given the formula (for 8B, 14B and 32B models, given that BERT model is 110M). Further, we computed the runtime of the inference of the BERT classifier, per sample, during standard inference (i.e. without early-stopping). To do so, we inferred the online version of TRACES with \(\delta=0\), and monitored the inference runtime per step of the BERT classifiers. Table 6.1 also reports the average runtime overhead per sample of the classifier during online inference of TRACES, the last column of our table. We note that the runtime overhead obtained aligns with our estimated overhead.

\begin{table}[h]
\centering
\tiny
\begin{tabular}{lccccc}
\toprule
\textbf{Model} & \textbf{Size ratio} & \textbf{Estimated overhead (\%)} & \textbf{Classifier runtime (s)} & \textbf{Runtime overhead (\%)} \\ 
\midrule
DS-8B  &   $\times$73  & 1.38 & 1.44 & 1.51 \\
DS-14B &   $\times$127 & 0.79 & 1.14 & 0.88 \\
QwQ-32B &  $\times$291 & 0.34 & 1.17 & 0.40 \\
\bottomrule
\end{tabular}
\caption{Runtime of the step classifiers per sample - MATH500 test, seed 42. Runtime per step ± 0.01 to 0.02s}
\label{tab:flops_overhead}
\end{table}

Next, we also compare the overall runtime of TRACES ($\delta=0$) with the standard inference. Table \ref{tab:computation_online} compares the average number tokens generated per second, and the overall runtime per sample. We observe that the token rate per second is slightly lower for TRACES compared to the standard inference, resulting in higher average runtime per sample. However, as confirmed per Table 6.1, this is due to the latency introduced by the halt of the inference to monitor the reasoning steps rather than by the classifier itself. Our analysis in Table \ref{tab:online-efficiency-TRACES} (see Appendix \ref{sec:appendix-online-latency}) confirms that the early-stopping mechanism lower the number of tokens, and recover the runtime loss, resulting in lower runtime overall.

\begin{table}[h]
\centering
\tiny
\begin{tabular}{lcccccc}
\toprule
\textbf{Model} & \textbf{Avg \# of tokens} & \textbf{Avg Tok/s (OG)} & \textbf{Avg Tok/s (TRACES)} & \textbf{Avg Runtime (s) (OG)} & \textbf{Avg Runtime (s) (TRACES)} \\ 
\midrule
DS-8B  & 3810.6 & 43.00 ± 3.9 & 36.85 ± 3.2 & 95.65 & 106.31 \\
DS-14B & 3907.6 & 28.57 ± 2.7 & 24.84 ± 1.9 & 128.82 & 151.21 \\
QwQ-32B & 4672.7 & 17.3 ± 1.73 & 15.43 ± 5.55 & 296.47 & 323.83 \\
\bottomrule
\end{tabular}
\caption{Comparison of standard inference (OG) vs. TRACES ($\delta$=0)}
\label{tab:computation_online}
\end{table}

\newpage

\subsection{Training-inference cost trade-off} \label{sec:appendix-training-inference}

Our framework requires training of BERT Step-Taggers to label the steps during LRMs generation. While this is a one-time exercise, this requirement implies additional computation. This section offers an analysis of the training-inference cost trade-off of our TRACES framework.

\textbf{Methodology.} The training cost of our TRACES framework includes three components:

\begin{itemize}[leftmargin=*, itemsep=0pt, topsep=0pt]

    \item \textbf{Training inference.} First, we first need to infer training samples of reasoning datasets. The resulting traces are used to train our BERT classifiers.

    \item \textbf{Annotation cost.} Second, we need to label the reasoning traces using \GPTmini{} to train our BERT classifiers.

    \item \textbf{Training BERT Classifiers.} Finally, we need to train our BERT classifiers to then apply our framework on test samples.
    
\end{itemize}

Each steps implies additional costs. To allow a fair comparison with our gains obtained by our framework, we will quantify these using the saved-runtime compared to standard inference against the token-count at inference of our TRACES framework. 

\textbf{On the size of training datasets.} In our experiments, we deliberately used a larger training dataset than necessary for the TRACES framework to demonstrate its effectiveness (training datasets being twice the size of testing datasets for MATH500, and using an additional reasoning dataset -- GPQA). While this choice inflates the training cost, it ensures that the model performance is not limited by data availability. As shown in Section \ref{sec:influence-step-type}, the framework can achieve good performance using training data from a single model and only two reasoning datasets (with good generalization on other models and datasets). 

\begin{figure}[h] 
    \centering
    \vspace{-0.25cm}
    \begin{minipage}{0.45\linewidth}
        \includegraphics[width=\linewidth]{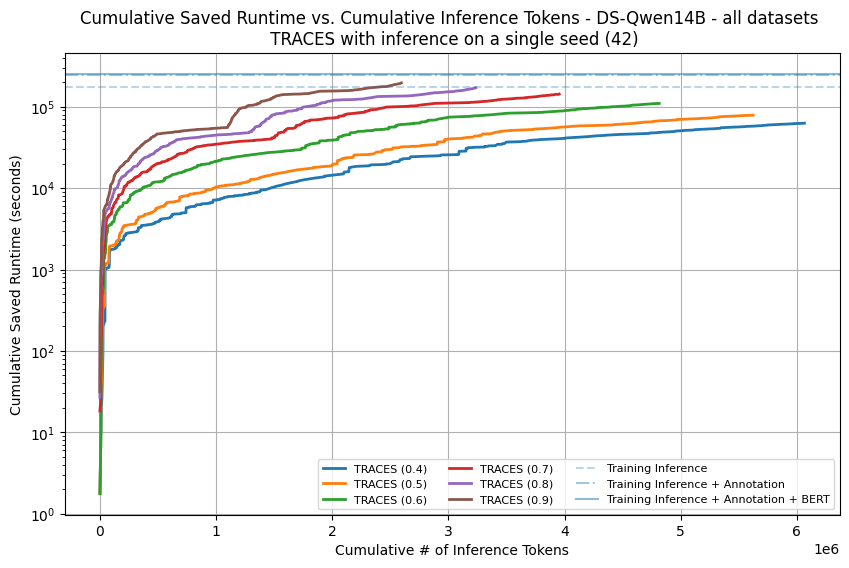}
        \subcaption{Single seed (42)}
        \label{fig:trade_off_single_seed}
    \end{minipage}
    \hfill
    \begin{minipage}{0.5\linewidth}
        \includegraphics[width=\linewidth]{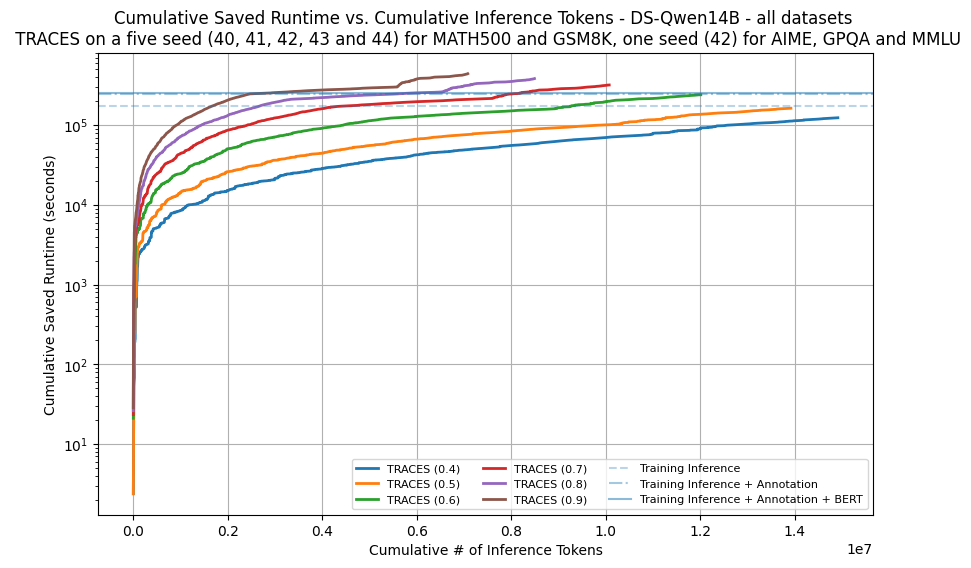}
        \subcaption{Five seeds (40 to 44)}
        \label{fig:trade_off_five_seed}
    \end{minipage}
    \vspace{-0.2cm}
    \caption{\small Training-Inference cost trade-off with original training and combined testing datasets. Figures \ref{fig:trade_off_single_seed} and \ref{fig:trade_off_five_seed} with one and five seeds, respectively, for MATH500 and GSM8K datasets. Single seed for AIME, GPQA and MMLU - DS-Qwen14B}
    \label{fig:training_trade_off_overall}
    \vspace{-0.25cm}
\end{figure}

\textbf{Overall results.} Figure \ref{fig:training_trade_off_overall} shows the Training-Inference cost trade-off overall for the \deepseekQwen{} model. In this Figure, MATH500 and GPQA train datasets are used to train the Step-Tagger module, and each configuration curve sums the saved runtime over the five selected test datasets. Figures \ref{fig:trade_off_single_seed} and \ref{fig:trade_off_five_seed} accumulates the saved runtime per token generated for one and five seeds on MATH500 and GSM8K datasets, respectively.

On Figure \ref{fig:trade_off_single_seed}, we observe that for the most restrictive configurations ($\delta \geq 0.8$), the saved inference almost recovers the training efforts. A large part of the training runtime corresponds to the Training inference (i.e. generate CoT data), and those configurations are above that line. Other configurations (e.g. $\delta \leq 0.7$) offers lower saved runtime through inference. It means that it takes more samples -- and tokens generated -- to recover the training costs. 

Additionally, Figure \ref{fig:trade_off_five_seed} accumulates the saved runtime over the four additional seeds we used for MATH500 and GSM8K. In that case, we observe that the curves for $\delta \geq 0.6$ fully recovers the training costs. Furthermore, we note that our step-tagging module has been used for other models (\deepseekLlama{} and \QwQLarge{}). Therefore, we could also consider summing the saved runtime obtained for these models -- over the five seeds used, and all the configurations would recover the training efforts by far.

\textbf{Estimation of the training-inference trade-off per model and datasets.} To allow a fair comparison of training-inference costs for each datasets, we first assume that each training dataset is used to train separate classifiers. This setting enables us to compare the computational cost of our TRACES framework under consistent conditions between datasets. 

\begin{figure}[h] 
    \centering
    \vspace{-0.25cm}
    \begin{minipage}{0.45\linewidth}
        \includegraphics[width=\linewidth]{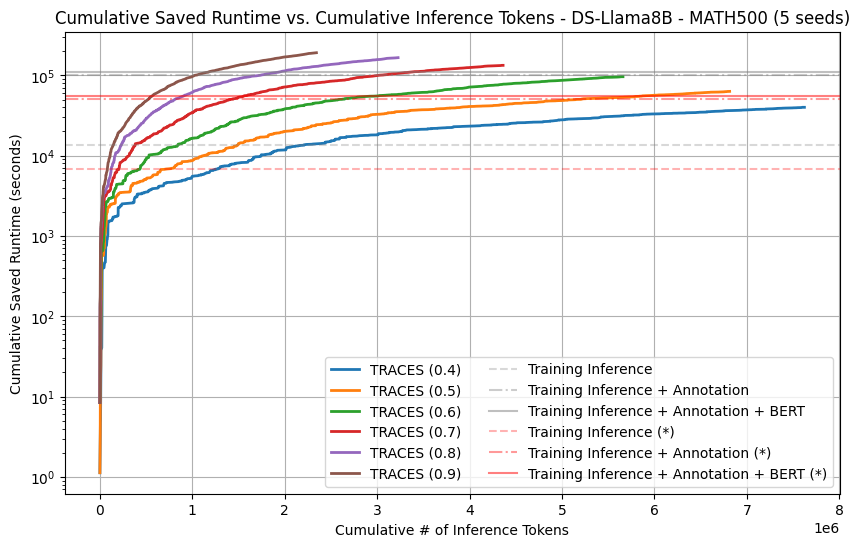}
        \subcaption{MATH500 on DS-8B}
        \label{fig:trade_off_math500_ds8b}
    \end{minipage}
    \hfill
    \begin{minipage}{0.45\linewidth}
        \includegraphics[width=\linewidth]{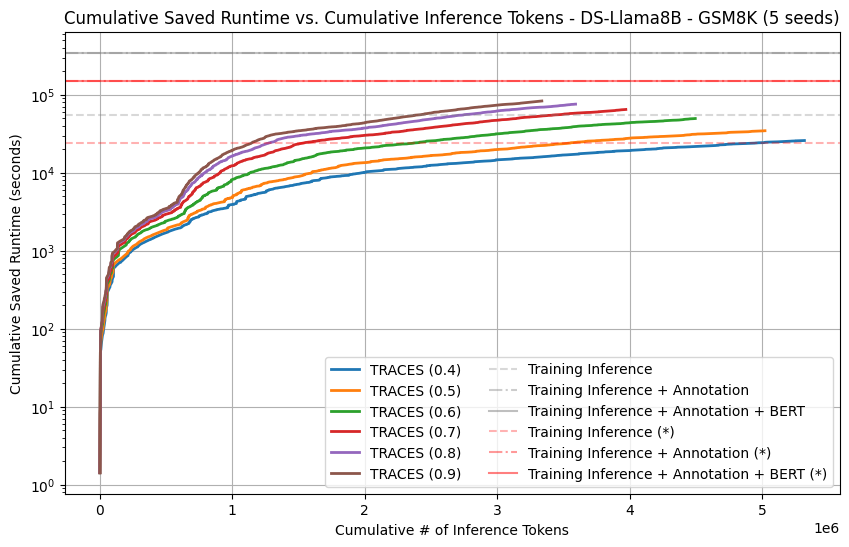}
        \subcaption{GSM8K on DS-8B}
        \label{fig:trade_off_gsm8k_ds8b}
    \end{minipage}

    \begin{minipage}{0.45\linewidth}
        \includegraphics[width=\linewidth]{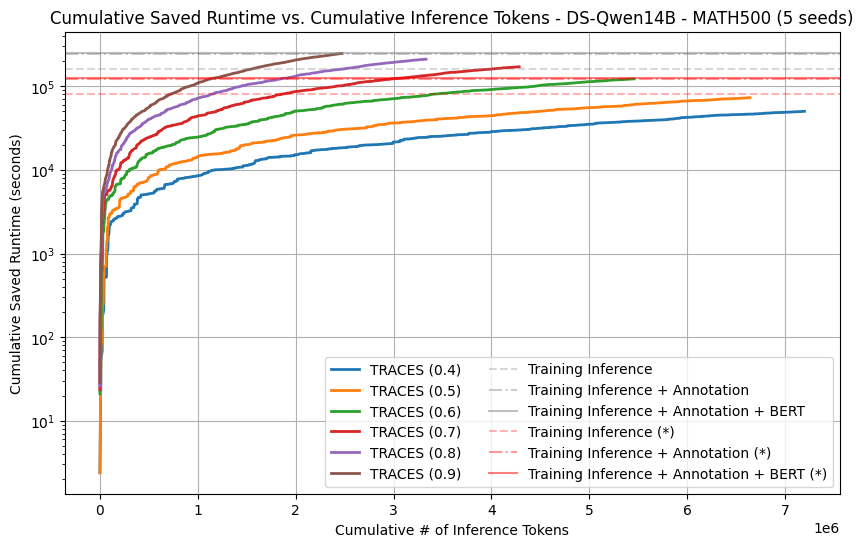}
        \subcaption{MATH500 on DS-14B}
        \label{fig:trade_off_math500_ds14b}
    \end{minipage}
    \hfill
    \begin{minipage}{0.45\linewidth}
        \includegraphics[width=\linewidth]{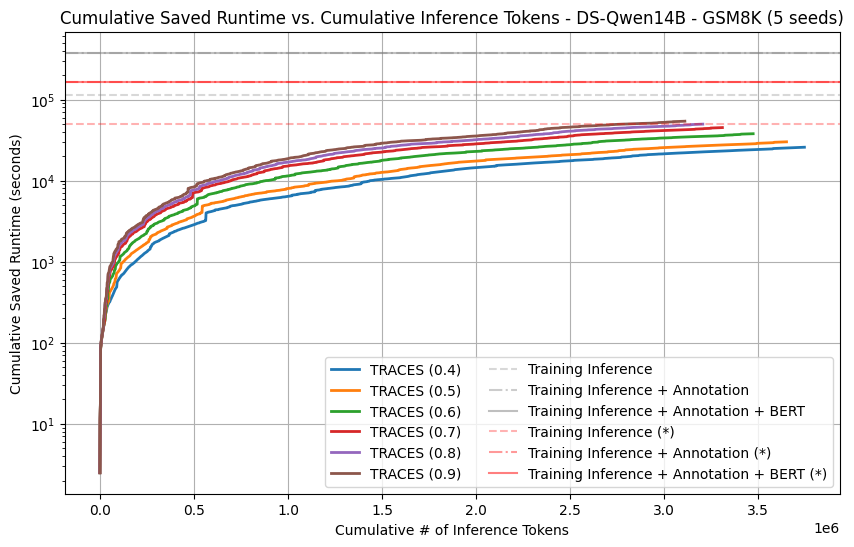}
        \subcaption{GSM8K on DS-14B}
        \label{fig:trade_off_gsm8k_ds14b}
    \end{minipage}

    \begin{minipage}{0.45\linewidth}
        \includegraphics[width=\linewidth]{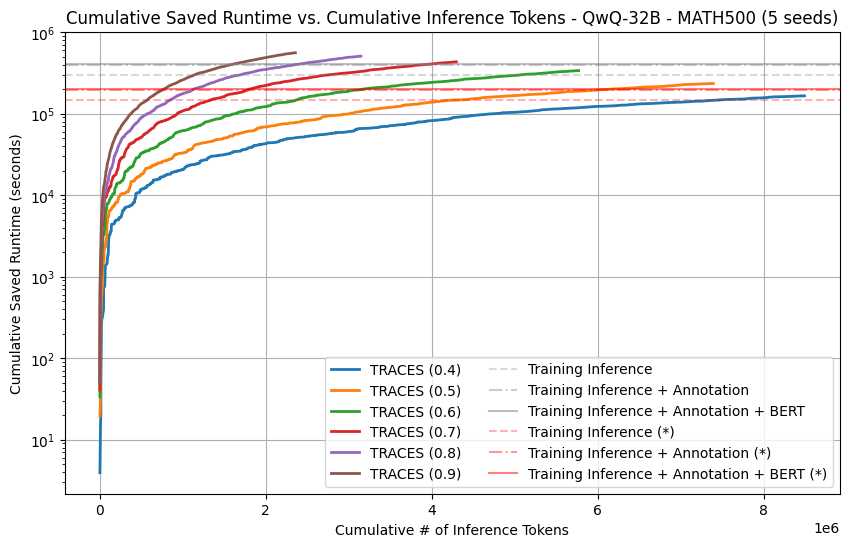}
        \subcaption{MATH500 on QwQ-32B}
        \label{fig:trade_off_math500_qwq32b}
    \end{minipage}
    \hfill
    \begin{minipage}{0.45\linewidth}
        \includegraphics[width=\linewidth]{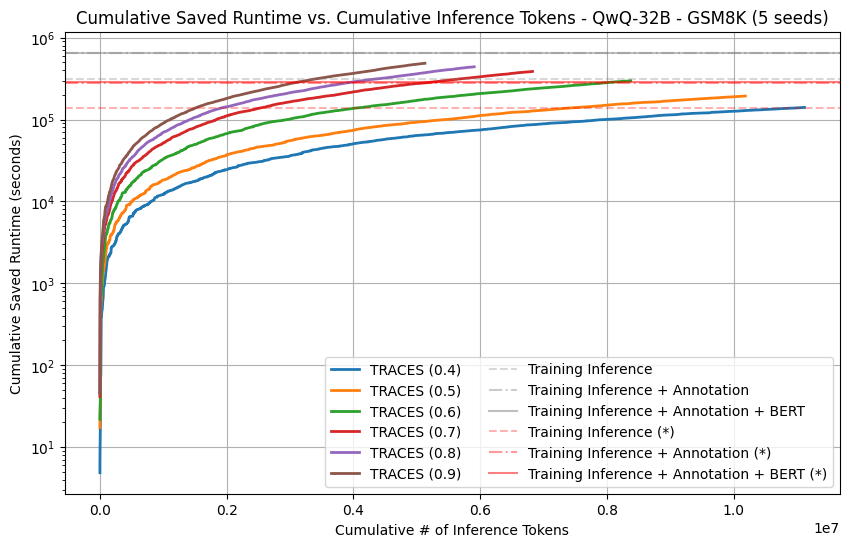}
        \subcaption{GSM8K on QwQ-32B}
        \label{fig:trade_off_gsm8k_qwq32b}
    \end{minipage}
    \vspace{-0.2cm}
    \caption{Training-Inference cost trade-off - MATH500 and GSM8K (5 seeds)}
    \label{fig:training_trade_off_math500_gsm8k}
    \vspace{-0.25cm}
\end{figure}

\textbf{MATH500 and GSM8K.} Figure \ref{fig:training_trade_off_math500_gsm8k} show the cumulative saved runtime of the TRACES configurations obtained compared to the standard inference baseline, against the cumulative number of inference tokens on the test MATH500 and GSM8K datasets -- using the five seeds, for the three selected LRMs. First, we observe that the inference of the models, as well as the annotation of the traces by \texttt{GPT-4o-mini} represent a significant amount of the cumulated runtime. In comparison, the estimated training time of Step-Taggers is of lower magnitude (e.g. for MATH500 on DS-Qwen14B - 2.6 hours of training runtime for BERT vs. 67.5 hours for the reasoning data generation + annotation, see lines in gray in Figure \ref{fig:trade_off_math500_ds14b}). 

\textbf{Realistic configuration.} The training datasets are significantly larger for both MATH500 and GSM8K, which inflates the training time. To address this, we considered smaller training datasets, equivalent to the size of the test datasets (* - see red lines). We assume that such size is enough to train accurate classifiers on the focused model and dataset. Under this assumption, we observe that most of our TRACES configurations fully recover the cost of training during inference compared to standard inference. Also, including 5 inference seeds is realistic. Indeed, the training is a one-time exercise and the model is then used for many inferences. It shows that TRACES scales as more tokens are generated during inference.

\textbf{AIME.} Similarly, our TRACES framework lead to interesting training-inference trade-off on the AIME dataset. Indeed, the TRACES ($\delta = 0.9$) configuration recovers all training cost by 100k inference token generated. We also observe that other configurations seems to scale well. Since training runtime are fixed (one time exercise), we expected consistent runtime saving when scaling the inference of our TRACES configurations. 

\textbf{GPQA and MMLU-Pro.} Same observation can be made on the GPQA dataset, where the TRACES ($\delta \geq 0.7$) configurations recover all training cost by 600k inference token generated. Interestingly, the TRACES ($\delta \geq 0.6$) configurations on the MMLU-Pro dataset recovers the training cost before 1.5M inference token generated, and leads to more than two fold training runtime saving at full test dataset inference. This is because we selected a training dataset of size being only 11\% of the test dataset size, significantly decreasing the training runtime. It demonstrates that on low resources setting, selecting a small training dataset can reduce the training cost, while still leading to high efficiency gains and maintaining the accuracy.

\begin{figure}[h] 
    \centering
    \begin{minipage}{0.45\linewidth}
        \includegraphics[width=\linewidth]{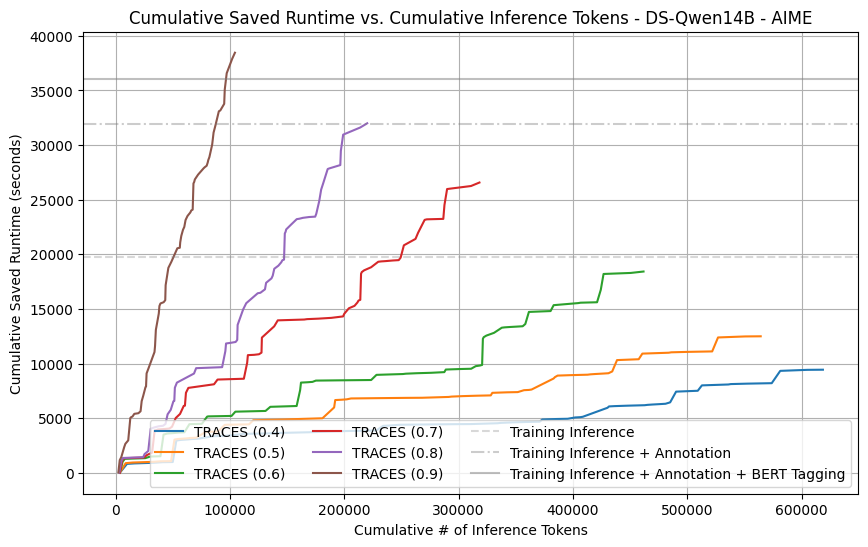}
        \subcaption{\small AIME}
        \label{fig:trade_off_aime_ds14b}
    \end{minipage}
    \\ 
    \begin{minipage}{0.45\linewidth}
        \includegraphics[width=\linewidth]{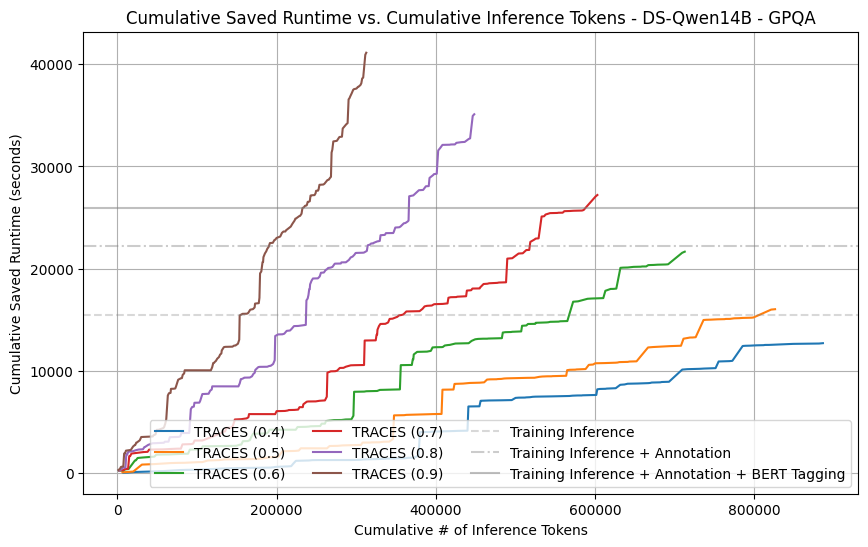}
        \subcaption{\small GPQA}
        \label{fig:trade_off_gpqa_ds14b}
    \end{minipage}
    \hfill
    \begin{minipage}{0.45\linewidth}
        \includegraphics[width=\linewidth]{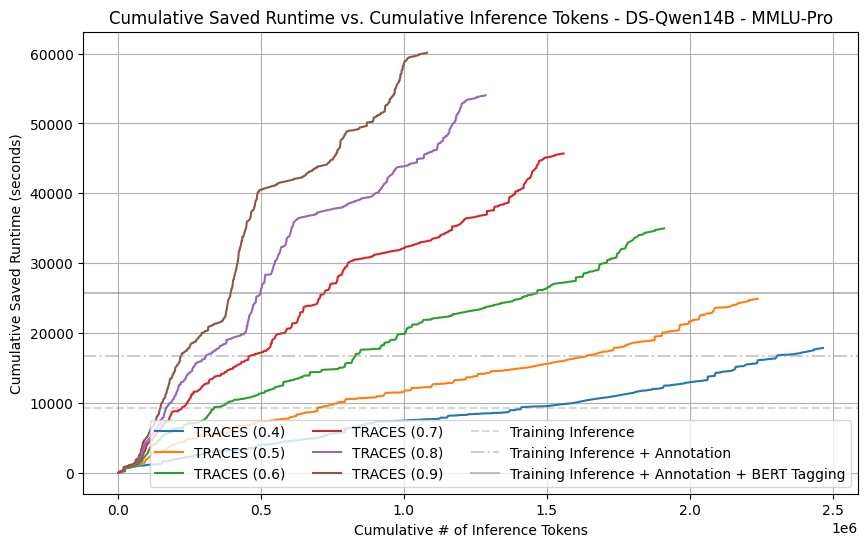}
        \subcaption{\small MMLU-Pro}
        \label{fig:trade_off_mmlu_ds14b}
    \end{minipage}
    \vspace{-0.2cm}
    \caption{Training-Inference cost trade-off - DS-Qwen14B on seed $42$}
    \label{fig:training_trade_off_other_datasets}
    \vspace{-0.3cm}
\end{figure}

\textbf{Takeaways.} Overall, we demonstrated that our TRACES framework can recover its one-time training cost during inference, and continues to deliver additional runtime saving compared to standard inference as more inference tokens are generated. The size of the training datasets mainly impact the trade-off. This is because our training procedure includes two costly and equivalent phases: the generation and annotation of CoT data. In comparison, the cost of training step-tagger modules is minimal (of smaller magnitude). 

Furthermore, we showed that the training cost can be substantially reduced while retaining the performance of our framework. More specifically, we demonstrated that our the step-tagging module can be trained once, on a single model and limited datasets, but still generalize well. This finding confirms the good robustness of our framework, and reduces significantly the training-inference cost trade-off.

\newpage

\section{Beyond efficiency: Failure Case Analysis with TRACES}

\subsection{Verification steps against problem's difficulty} \label{sec:appendix-verificaton-pb-difficulty}

To better understand how TRACES works, understanding the relation between the number of verification steps and the problem difficulty is interesting and is an important analysis. To address this, we suggest to observe the number of verification steps per complexity level of the MATH500 dataset. In the main body, we defined the ideal early-stopping (see Section \ref{sec:experimental-section}), and differentiated steps located before and after this ideal step.

Table \ref{tab:verification_analysis} presents the average number of \emph{Evaluative} step-types used by TRACES per complexity level of the MATH500 train dataset (i.e. Verification and Final Conclusion) located before and after the ideal early-stopping step, which we refer to as pre-ES and post-ES, respectively.

\begin{table}[h]
\tiny
\centering
\begin{tabular}{l cccc | ccc}
\toprule
 & \multicolumn{4}{c|}{\textbf{Correct samples}} & \multicolumn{3}{c}{\textbf{Incorrect samples}} \\
\cmidrule(lr){2-5} \cmidrule(lr){6-8}
\textbf{Level} & \textbf{N} & \textbf{Eval pre-ES} & \textbf{Eval post-ES} & \textbf{Ratio (post/pre)} & \textbf{N} & \textbf{Eval pre-ES} & \textbf{Constr pre-ES} \\
\midrule
L1 & 84  & 0.25 & 10.31 & 41.2$\times$ & 5  & 25.60 & 64.00 \\
L2 & 172 & 0.94 & 11.97 & 12.7$\times$ & 8  & 6.75  & 35.13 \\
L3 & 197 & 2.20 & 16.47 & 7.5$\times$  & 13 & 13.38 & 22.62 \\
L4 & 210 & 2.30 & 17.46 & 7.6$\times$  & 17 & 25.47 & 92.82 \\
L5 & 232 & 4.38 & 19.62 & 4.5$\times$  & 62 & 22.95 & 138.84 \\
\bottomrule
\end{tabular}%
\caption{Analysis of evaluative (verification and final conclusion) steps before and after the ideal early-stopping (IES) point, split by difficulty level and correctness. \emph{Eval pre-ES} counts verification steps occurring before the correct answer is first reached, \emph{Eval post-ES} counts those occurring after. The ratio (post/pre) quantifies how much verification is wasted relative to useful verification on correctly-solved problems.}
\label{tab:verification_analysis}
\end{table}

Looking at the correct samples only, we observe that the number of Eval pre-ES grows as the difficulty increases. It confirms that more verification steps seems to be needed to get to the correct answer for more difficult problems. As we detailed in Appendix \ref{sec:appendix-infl-fact-threshold}, TRACES can adapt to the difficulty of the question by setting the $\delta$ carefully. We showed that tasks requiring more computations (i.e. more complex tasks) needs a lower value of $\delta$ (typically in the range of [0.4,0.6]) to allow the model to verify itself more extensively. Interestingly, when looking at the incorrect samples, there is no specific correlation between the complexity and the number of evaluation steps.

\newpage

\subsection{Pattern incorrect traces}

Further, a deeper analysis on the patterns in failure cases (i.e. incorrect traces) would be interesting, as the TRACES framework could reveal interesting patterns.

In Appendix \ref{sec:appendix-ideal-early-stopping} and \ref{sec:appendix-verificaton-pb-difficulty}, we conducted extended analysis on the reasoning transition phase, but we only looked at correct reasoning traces. To better analyze the patterns in failure cases using TRACES, we introduce three metrics:

\begin{itemize}
    \item \textbf{The entropy of reasoning steps:} Given a step sequence $S = \{s_1, \dots, s_n\}$, we obtained the associated tags $\tau_i$ for each steps $s_i$, namely $T = \{\tau_1, \dots, \tau_n\}$. We then compute the entropy of the step type sequence: $E_S = -\sum{p_i log(p_i)}$, where $p_i = \frac{|1(\tau_j, \tau^*)}{|T|}$. Basically, we compute the "diversity" of the step-types for a given sequence. If the sequence is diverse, the entropy score should be higher.

    \item \textbf{The number of loops:} Given a step sequence, we recall the Constructive (Problem Re-statement and Definition Recall) and Evaluative (Verification and Final Conclusion) step-types. We then count the number of times the model moves from Evaluative to Constructive and then back to Evaluative. The higher this number is, the higher the model is looping from constructing its answer to verifying it.

    \item \textbf{Stagnation:} Given a step sequence, we now only consider the raw tags (as per the ReasonType taxonomy). We count the maximum number of successive steps of the same types in the reasoning trace. 

\end{itemize}

Table \ref{tab:incorrect_pattern_failure_analysis} reports our results on the MATH500 train dataset for DS-Llama8B, DS-Qwen14B, and QwQ-32B.

\begin{table}[h]
\tiny
\centering
\begin{tabular}{l ccccc}
\toprule
 & \multicolumn{5}{c}{\textbf{Correct samples}} \\
\cmidrule(lr){2-6} 
\textbf{Model} & \textbf{N} & \textbf{Entropy 0-50\%} & \textbf{Entropy 50-100\%} & \textbf{Loops} & \textbf{Stagnation} \\
\midrule
DS-8B & 895 & 1.61 ± 0.34 & 1.65 ± 0.33 & 4.83 ± 5.93 & 5.51 ± 4.00 \\
DS-14B & 941 & 1.59 ± 0.38 & 1.65 ± 0.33 & 4.67 ± 5.52 & 5.36 ± 2.84 \\
QwQ-32B & 967 & 1.77 ± 0.28 & 1.82 ± 0.26 & 6.40 ± 6.14 & 5.89 ± 2.97 \\

\midrule

& \multicolumn{5}{c}{\textbf{Incorrect samples}} \\
\cmidrule(lr){2-6} 
\textbf{Model} & \textbf{N} & \textbf{Entropy 0-50\%} & \textbf{Entropy 50-100\%} & \textbf{Loops} & \textbf{Stagnation} \\
\midrule
DS-8B & 105 & 1.82 ± 0.32 & 1.78 ± 0.31 & 9.34 ± 7.22 & 6.67 ± 3.74 \\
DS-14B & 59 & 1.86 ± 0.41 & 1.81 ± 0.41 & 13.81 ± 13.48 & 6.53 ± 4.05 \\
QwQ-32B & 33 & 1.99 ± 0.33 & 2.03 ± 0.26 & 15.88 ± 12.76 & 7.00 ± 3.50 \\

\bottomrule
\end{tabular}%
\caption{Analysis of stagnation and loops in correct vs. incorrect reasoning traces - MATH500 train}
\label{tab:incorrect_pattern_failure_analysis}
\end{table}

Table \ref{tab:incorrect_pattern_failure_analysis} shows the step-type entropy for early-generation (0 to 50\% to completion), and for late-generation (50\% to 100\% to completion). We observe that for correct samples, the entropy is slightly higher for late-generation (so slightly more diverse). Interestingly, the entropy of reasoning sequences is systematically higher for incorrect samples compared to correct samples. It means that the type of steps in incorrect reasoning sequences is more variable than for correct reasoning traces.

Importantly, Table \ref{tab:incorrect_pattern_failure_analysis} reports the average number of loops and stagnation. We note that across models, the two metrics are systematically higher for incorrect reasoning traces, especially for the number of loops (4.83 vs. 9.34 for DS-8B, 4.67 vs. 13.81 for DS-14B and 6.40 vs. 15.88 for QwQ-32B). Our TRACES framework seems to highlight a different pattern for incorrect traces than for correct traces.

\textbf{Takeaway.} With further analysis, our findings supports that TRACES could be leverage to detect wrong reasoning traces, which could have even more impact regarding the efficiency of the model. For instance, if a pattern, such as looping or stagnation, is detected early on, we could cut the generation earlier, or intervene in the generation to recover correctness.

\newpage

\section{Performance of TRACES} \label{sec:appendix-st-es-performance}

\subsection{MATH500 and GSM8K} \label{sec:appendix-st-es-performance-math500-gsm8k}

\begin{figure}[h]
    \centering
    \begin{subfigure}{0.32\linewidth}
        \centering
        \includegraphics[width=\linewidth]{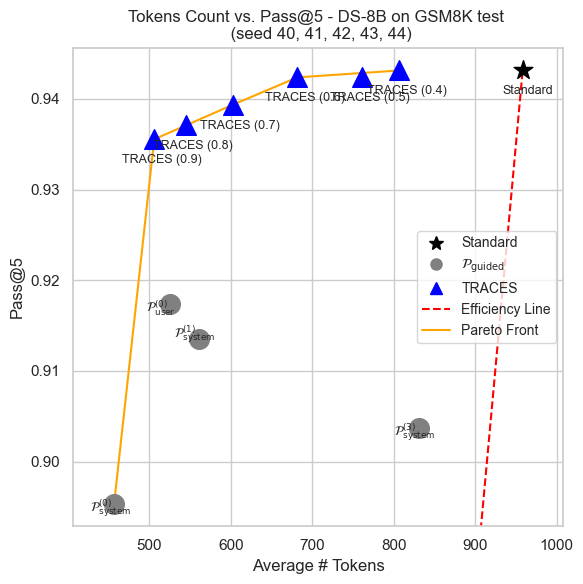}
        \caption{DS-Llama8B on GSM8K}
        \label{fig:DS8B-GSM8K-pass5} 
    \end{subfigure}
    \hfill
    \begin{subfigure}{0.32\linewidth}
        \centering
        \includegraphics[width=\linewidth]{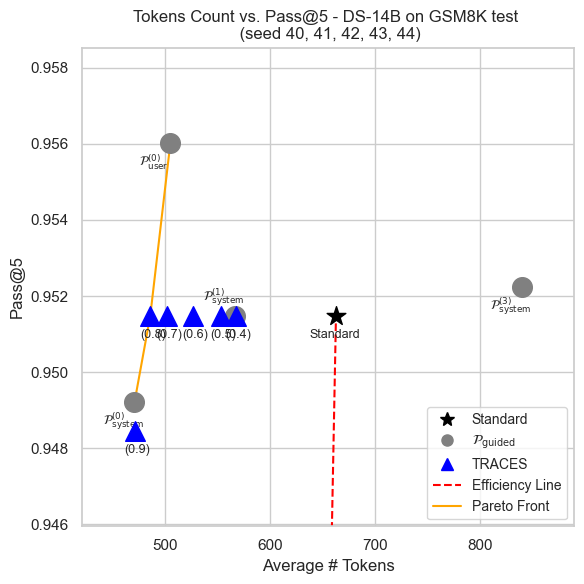}
        \caption{DS-Qwen14B on GSM8K}
        \label{fig:DS14B-GSM8K-pass5}
    \end{subfigure}
    \hfill
    \begin{subfigure}{0.32\linewidth}
        \centering
        \includegraphics[width=\linewidth]{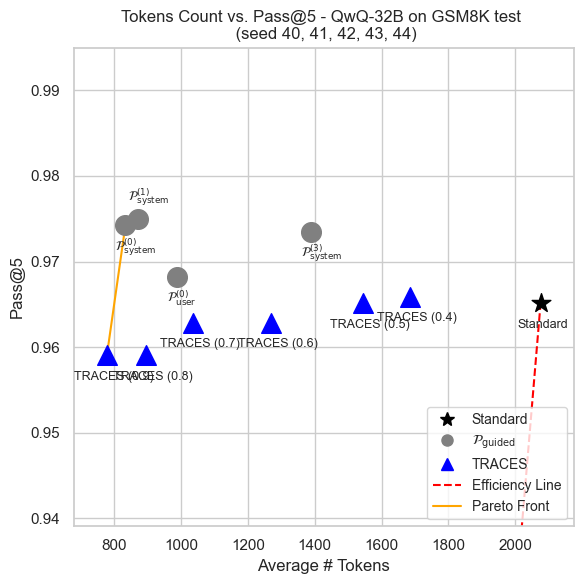}
        \caption{QwQ-32B on GSM8K}
        \label{fig:QwQ32B-GSM8K-pass5}
    \end{subfigure}
    \begin{subfigure}{0.32\linewidth}
        \centering
        \includegraphics[width=\linewidth]{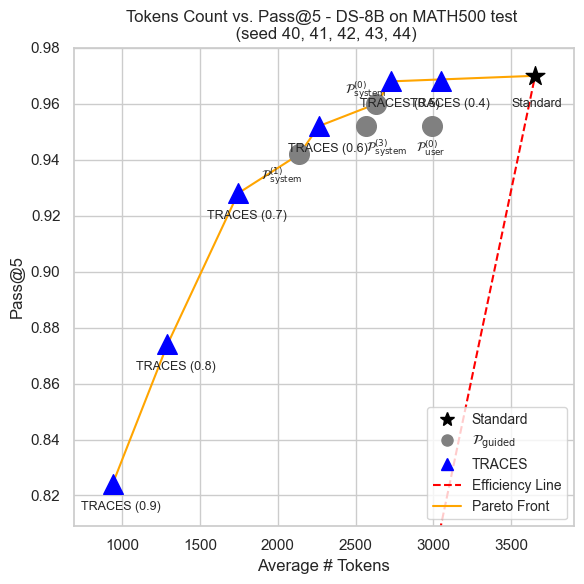}
        \caption{DS-Llama8B on MATH500}
        \label{fig:DS8B-MATH500-pass5} 
    \end{subfigure}
    \hfill
    \begin{subfigure}{0.32\linewidth}
        \centering
        \includegraphics[width=\linewidth]{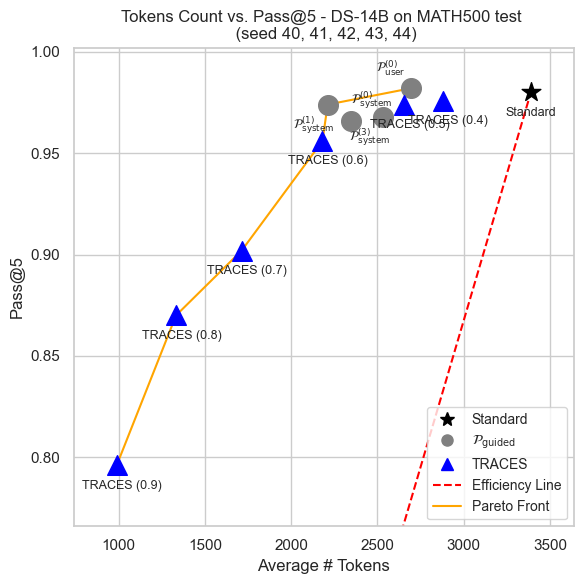}
        \caption{DS-Qwen14B on MATH500}
        \label{fig:DS14B-MATH500-pass5}
    \end{subfigure}
    \hfill
    \begin{subfigure}{0.32\linewidth}
        \centering
        \includegraphics[width=\linewidth]{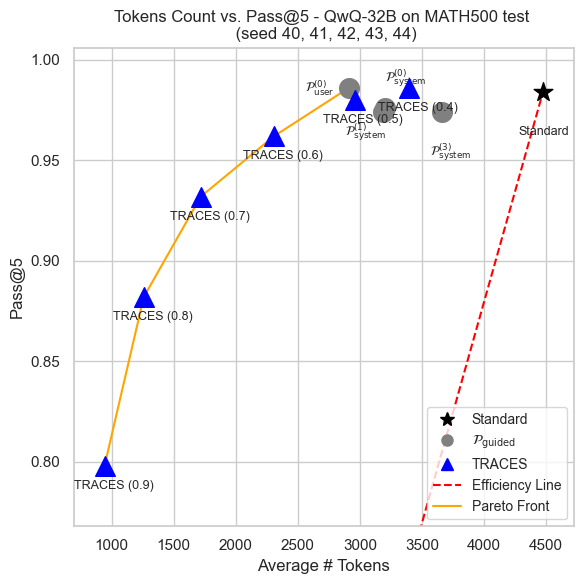}
        \caption{QwQ-32B on MATH500}
        \label{fig:QwQ32B-MATH500-pass5}
    \end{subfigure}
    \vspace{-0.2cm}
    \caption{Number of Tokens vs. Pass@$5$ - $\mathcal{P_{\text{guided}}}$ Baselines vs. TRACES criteria}
    \label{fig:st-es-performance-pass-5}
    \vspace{-0.5cm}
\end{figure}

\begin{figure}[h]
    \centering
    \begin{subfigure}{0.32\linewidth}
        \centering
        \includegraphics[width=\linewidth]{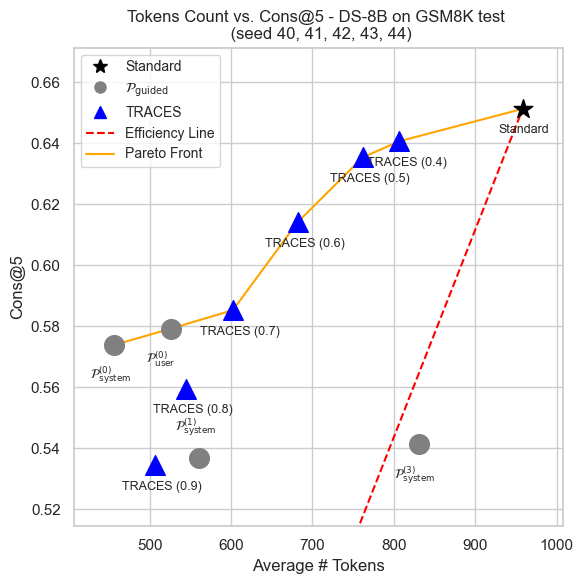}
        \caption{DS-Llama8B on GSM8K}
        \label{fig:DS8B-GSM8K-cons5} 
    \end{subfigure}
    \hfill
    \begin{subfigure}{0.32\linewidth}
        \centering
        \includegraphics[width=\linewidth]{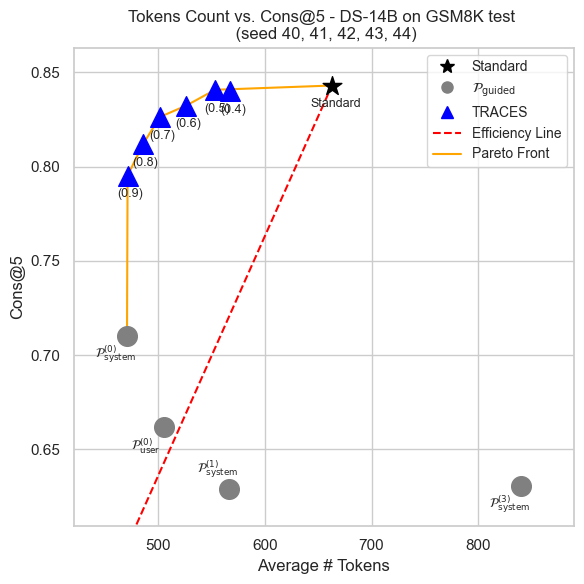}
        \caption{DS-Qwen14B on GSM8K}
        \label{fig:DS14B-GSM8K-cons5}
    \end{subfigure}
    \hfill
    \begin{subfigure}{0.32\linewidth}
        \centering
        \includegraphics[width=\linewidth]{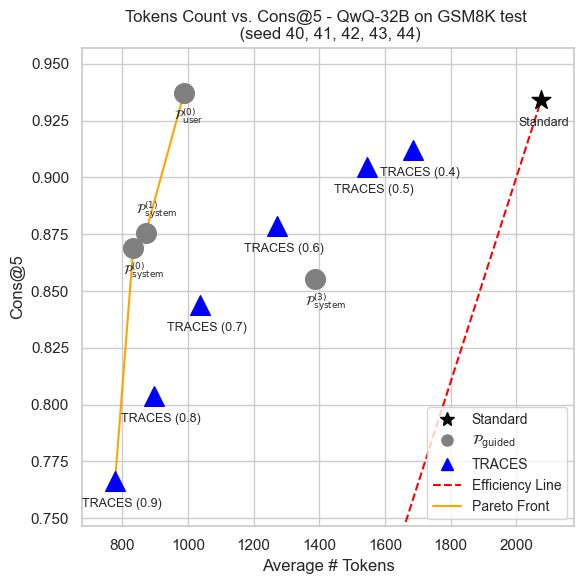}
        \caption{QwQ-32B on GSM8K}
        \label{fig:QwQ32B-GSM8K-cons5}
    \end{subfigure}
    \begin{subfigure}{0.32\linewidth}
        \centering
        \includegraphics[width=\linewidth]{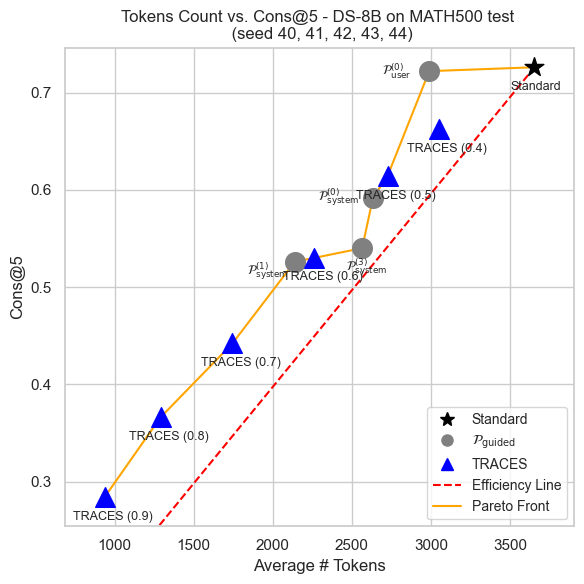}
        \caption{DS-Llama8B on MATH500}
        \label{fig:DS8B-MATH500-cons5} 
    \end{subfigure}
    \hfill
    \begin{subfigure}{0.32\linewidth}
        \centering
        \includegraphics[width=\linewidth]{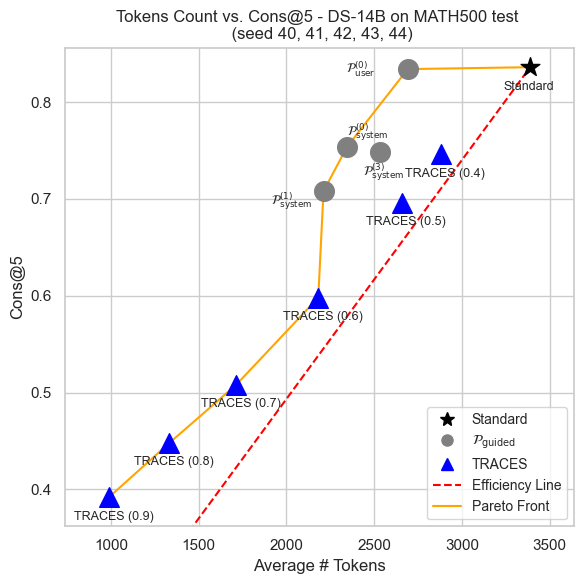}
        \caption{DS-Qwen14B on MATH500}
        \label{fig:DS14B-MATH500-cons5}
    \end{subfigure}
    \hfill
    \begin{subfigure}{0.32\linewidth}
        \centering
        \includegraphics[width=\linewidth]{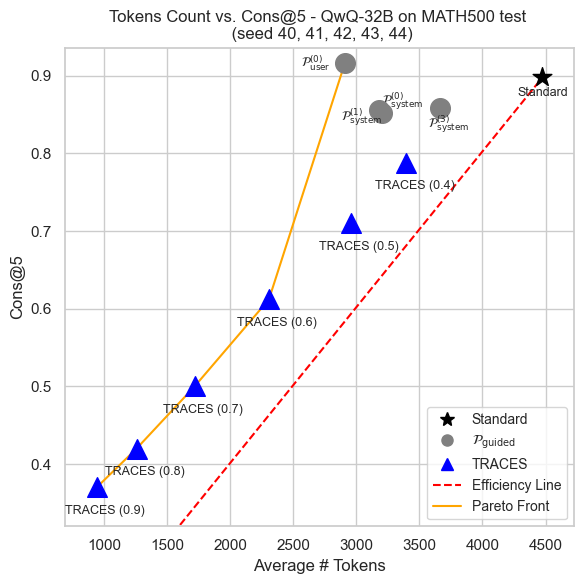}
        \caption{QwQ-32B on MATH500}
        \label{fig:QwQ32B-MATH500-cons5}
    \end{subfigure}
    \vspace{-0.2cm}
    \caption{Number of Tokens vs. Cons@$5$ - $\mathcal{P_{\text{guided}}}$ Baselines vs. TRACES criteria}
    \label{fig:st-es-performance-cons-5}
    \vspace{-0.5cm}
\end{figure}

Table \ref{tab:full_results_5_seeds} reports all the token-usage, the proportion of saved number of tokens, the Avg@5, the Pass@5 and the Cons@5 for all configurations. Results are averaged over the 5 seeds we used. We also show in Figure \ref{fig:st-es-performance-pass-5} and \ref{fig:st-es-performance-cons-5} the average token count against the Pass@$5$ and Cons@$5$, respectively, for the three LRMs on the MATH500 and GSM8K datasets.

\begin{table}[h]
\centering
\tiny
\begin{tabular}{ll
                ccccc
                ccccc}
\toprule
\multirow{2}{*}{\textbf{Model}} & \multirow{2}{*}{\textbf{Config.}} 
& \multicolumn{5}{c}{\textbf{MATH500}} 
& \multicolumn{5}{c}{\textbf{GSM8K}} \\
\cmidrule(lr){3-7} \cmidrule(lr){8-12}
& & \# Tokens & Saved & Avg@5 & Pass@5 & Cons@5 & \# Tokens & Saved & Avg@5 & Pass@5 & Cons@5 \\
\midrule

\multirow{10}{*}{DS-8B}
  & Base & 3655.0 & -- & 0.878 & 0.970 & 0.726 & 958.3 & -- & 0.829 & 0.943 & 0.651 \\
  
  \cmidrule(lr){3-7} \cmidrule(lr){8-12}
  
  & $\mathcal{P}^{(0)}_{\text{user}}$ & 2989.6 & 18.21 & 0.866 & 0.952 & 0.722 & 525.8 & 45.13 & 0.771 & 0.917 & 0.579 \\
  & $\mathcal{P}^{(0)}_{\text{system}}$ & 2634.4 & 27.92 & 0.817 & 0.960 & 0.592 & 456.9 & 52.32 & 0.763 & 0.895 & 0.574 \\
  & $\mathcal{P}^{(1)}_{\text{system}}$ & 2139.5 & 41.46 & 0.782 & 0.942 & 0.526 & 560.8 & 41.48 & 0.754 & 0.914 & 0.537 \\
  & $\mathcal{P}^{(3)}_{\text{system}}$ & 2565.3 & 29.81 & 0.789 & 0.952 & 0.540 & 830.5 & 13.34 & 0.748 & 0.904 & 0.541 \\
  
  \cmidrule(lr){3-7} \cmidrule(lr){8-12}
  
  & $\delta=0.4$ & 3049.3 & 16.57 & 0.851 & 0.968 & 0.662 & 806.8 & 15.81 & 0.825 & 0.943 & 0.641 \\
  & $\delta=0.5$ & 2725.8 & 25.42 & 0.829 & 0.968 & 0.614 & 761.6 & 20.53 & 0.823 & 0.942 & 0.635 \\
  & $\delta=0.6$ & 2264.1 & 38.05 & 0.785 & 0.952 & 0.530 & 681.8 & 28.85 & 0.815 & 0.942 & 0.614 \\
  & $\delta=0.7$ & 1745.2 & 52.25 & 0.715 & 0.928 & 0.442 & 602.3 & 37.18 & 0.804 & 0.939 & 0.585 \\
  & $\delta=0.8$ & 1290.8 & 64.68 & 0.649 & 0.874 & 0.366 & 544.8 & 43.15 & 0.792 & 0.937 & 0.559 \\
  & $\delta=0.9$ & 937.4 & 74.35 & 0.567 & 0.824 & 0.284 & 506.1 & 47.18 & 0.775 & 0.936 & 0.534 \\
  
\midrule

\multirow{10}{*}{DS-14B}
  & Base & 3388.8 & -- & 0.923 & 0.980 & 0.836 & 662.9 & -- & 0.910 & 0.952 & 0.843 \\
  
  \cmidrule(lr){3-7} \cmidrule(lr){8-12}
  
  & $\mathcal{P}^{(0)}_{\text{user}}$ & 2691.5 & 20.58 & 0.933 & 0.982 & 0.834 & 505.1 & 23.80 & 0.856 & 0.956 & 0.662 \\
  & $\mathcal{P}^{(0)}_{\text{system}}$ & 2346.2 & 30.77 & 0.886 & 0.966 & 0.754 & 470.9 & 28.96 & 0.873 & 0.949 & 0.710 \\
  & $\mathcal{P}^{(1)}_{\text{system}}$ & 2211.4 & 34.74 & 0.873 & 0.974 & 0.708 & 566.5 & 14.54 & 0.838 & 0.952 & 0.629 \\
  & $\mathcal{P}^{(3)}_{\text{system}}$ & 2535.0 & 25.19 & 0.879 & 0.968 & 0.748 & 839.6 & -26.65 & 0.841 & 0.952 & 0.631 \\
  
  \cmidrule(lr){3-7} \cmidrule(lr){8-12}
  
  & $\delta=0.4$ & 2878.4 & 15.06 & 0.892 & 0.976 & 0.746 & 567.8 & 9.92 & 0.909 & 0.951 & 0.840 \\
  & $\delta=0.5$ & 2656.1 & 21.62 & 0.865 & 0.974 & 0.696 & 553.4 & 11.43 & 0.909 & 0.951 & 0.841 \\
  & $\delta=0.6$ & 2181.4 & 35.63 & 0.811 & 0.956 & 0.598 & 526.5 & 14.23 & 0.906 & 0.951 & 0.832 \\
  & $\delta=0.7$ & 1713.3 & 49.44 & 0.735 & 0.902 & 0.508 & 501.7 & 16.82 & 0.904 & 0.951 & 0.826 \\
  & $\delta=0.8$ & 1332.9 & 60.66 & 0.675 & 0.870 & 0.448 & 485.9 & 18.47 & 0.899 & 0.951 & 0.812 \\
  & $\delta=0.9$ & 988.8 & 70.82 & 0.595 & 0.796 & 0.392 & 471.5 & 19.97 & 0.891 & 0.948 & 0.795 \\

\midrule

\multirow{10}{*}{QwQ-32B}
  & Base & 4475.3 & -- & 0.954 & 0.984 & 0.898 & 2075.7 & -- & 0.953 & 0.965 & 0.934 \\
  
  \cmidrule(lr){3-7} \cmidrule(lr){8-12}
  
  & $\mathcal{P}^{(0)}_{\text{user}}$ & 2908.8 & 35.00 & 0.955 & 0.986 & 0.916 & 988.0 & 52.40 & 0.952 & 0.968 & 0.937  \\
  & $\mathcal{P}^{(0)}_{\text{system}}$ & 3201.1 & 28.47 & 0.932 & 0.976 & 0.852 & 833.3 & 59.85 & 0.940 & 0.974 & 0.869 \\
  & $\mathcal{P}^{(1)}_{\text{system}}$ & 3182.4 & 28.89 & 0.925 & 0.974 & 0.856 & 871.2 & 58.02 & 0.943 & 0.975 & 0.876 \\
  & $\mathcal{P}^{(3)}_{\text{system}}$ & 3665.5 & 18.09 & 0.926 & 0.974 & 0.858 & 1387.3 & 33.16 & 0.935 & 0.974 & 0.855 \\
  
  \cmidrule(lr){3-7} \cmidrule(lr){8-12}
  
  & $\delta=0.4$ & 3395.6 & 24.12 & 0.913 & 0.986 & 0.788 & 1685.9 & 18.79 & 0.948 & 0.966 & 0.912 \\
  & $\delta=0.5$ & 2956.7 & 33.93 & 0.878 & 0.980 & 0.710 & 1544.7 & 25.58 & 0.946 & 0.965 & 0.904 \\
  & $\delta=0.6$ & 2307.5 & 48.44 & 0.824 & 0.962 & 0.612 & 1270.1 & 38.81 & 0.936 & 0.963 & 0.879 \\
  & $\delta=0.7$ & 1718.2 & 61.61 & 0.745 & 0.932 & 0.500 & 1035.3 & 50.12 & 0.925 & 0.963 & 0.844 \\
  & $\delta=0.8$ & 1258.4 & 71.88 & 0.668 & 0.882 & 0.420 & 895.9 & 56.84 & 0.909 & 0.959 & 0.804 \\
  & $\delta=0.9$ & 942.4 & 78.94 & 0.602 & 0.798 & 0.370 & 777.9 & 62.52 & 0.889 & 0.959 & 0.766 \\

\bottomrule
\end{tabular}
\caption{Performance of TRACES - 5 seeds (40, 41, 42, 43, 44)}
\label{tab:full_results_5_seeds}
\end{table}

\newpage

\newpage

\textbf{Variance of the results across the seeds.}

\begin{table}[h]
\centering
\tiny
\begin{tabular}{l
                cccc
                cccc}
\toprule
\multirow{2}{*}{\textbf{Datasets}} 
& \multicolumn{4}{c}{\textbf{MATH500}} 
& \multicolumn{4}{c}{\textbf{GSM8K}} \\
\cmidrule(lr){2-5} \cmidrule(lr){6-9} 
& Avg. \# Tok. & Std. \# Tok. & Avg@5 & Std@5 & Avg. \# Tok. & Std. \# Tok. & Avg@5 & Std@5 \\
\midrule

DS-Llama8B & 3655.0 & ±290.04 & 0.878 & ±0.009 & 958.3 & ±213.65 & 0.829 & ±0.014 \\

DS-Qwen14B & 3388.8 & ±164.40 & 0.923 & ±0.023 & 662.9 & ±57.26 & 0.910 & ±0.003 \\

QwQ-32B & 4475.3 & ±248.39 & 0.954 & ±0.018 & 2075.7 & ±16.08 & 0.953 & ±0.002 \\

\bottomrule
\end{tabular}
\caption{Variance of LRMs on MATH500 and GSM8K across the 5 seeds selected}
\label{tab:variance_results}
\end{table}

\subsection{Harder mathematical tasks and beyond mathematical tasks} \label{sec:appendix-harder-datasets}

\begin{table}[h]
\centering
\tiny
\begin{tabular}{l
                ccc
                ccc
                ccc}
\toprule
\multirow{2}{*}{\textbf{Config.}} 
& \multicolumn{3}{c}{\textbf{AIME 23-24}} & \multicolumn{3}{c}{\textbf{GPQA-Diamond}} & \multicolumn{3}{c}{\textbf{MMLU-Pro}} \\
\cmidrule(lr){2-4} \cmidrule(lr){5-7} \cmidrule(lr){8-10}
& \# Tokens & Saved & Pass@1 & \# Tokens & Saved & Pass@1 & \# Tokens & Saved & Pass@1 \\
\midrule

Standard & 13096.3 & - & 0.483 & 7024.7 & - & 0.603 & 2501.9 & - & 0.652 \\
  
  \cmidrule(lr){2-10} 
  
$\mathcal{P}^{(0)}_{\text{user}}$ & 12748.4 & 2.66 & 0.422 & 4801.0 & 31.66 & 0.577 & 2194.1 & 12.30 & 0.665 \\
$\mathcal{P}^{(0)}_{\text{system}}$ & 11095.9 & 15.27 & 0.500 & 4768.3 & 32.12 & 0.571 & 2292.9 & 8.35 & 0.675 \\
$\mathcal{P}^{(1)}_{\text{system}}$ & 10181.2 & 22.26 & 0.517 & 3736.2 & 46.81 & 0.539 & 1705.6 & 31.83 & 0.637 \\
$\mathcal{P}^{(3)}_{\text{system}}$ & 11635.7 & 11.15 & 0.467 & 4530.1 & 35.51 & 0.564 & 1935.2 & 22.65 & 0.613 \\
  
  \cmidrule(lr){2-10} 
  
$\delta=0.4$ & 10524.3 & 19.64 & 0.417 & 5683.2 & 19.10 & 0.589 & 1923.1 & 23.13 & 0.669 \\
$\delta=0.5$ & 9478.9 & 27.62 & 0.383 & 5297.7 & 24.58 & 0.583 & 1880.6 & 24.83 & 0.664 \\
$\delta=0.6$ & 7983.7 & 39.04 & 0.300 & 4570.1 & 34.94 & 0.603 & 1766.3 & 29.40 & 0.665 \\
$\delta=0.7$ & 5949.5 & 54.57 & 0.200 & 3866.4 & 44.95 & 0.622 & 1614.7 & 35.46 & 0.652 \\
$\delta=0.8$ & 4015.3 & 69.34 & 0.133 & 2874.6 & 59.07 & 0.583 & 1286.9 & 48.56 & 0.650 \\
$\delta=0.9$ & 1986.5 & 84.83 & 0.050 & 2004.7 & 71.46 & 0.558 & 989.8 & 60.44 & 0.634 \\

\bottomrule
\end{tabular}
\caption{Performance of TRACES - DS-Qwen14B - seed 42}
\label{tab:other_dataset}
\end{table}

\begin{table}[h]
\centering
\tiny
\begin{tabular}{ll
                ccc
                ccc}
\toprule
\multirow{2}{*}{\textbf{Model}}  &
\multirow{2}{*}{\textbf{Config.}} 
& \multicolumn{3}{c}{\textbf{BeyondAIME}} & \multicolumn{3}{c}{\textbf{IMOAnswer}} \\
\cmidrule(lr){3-5} \cmidrule(lr){6-8} 
& & \# Tokens & Saved (\%) & Pass@1 & \# Tokens & Saved (\%) & Pass@1 \\
\midrule

  
  
  
  


\multirow{12}{*}{Qwen3-8B}
  & Standard & 17972.8 & - & 0.380 & 17601.8 & - & 0.328 \\
  
  \cmidrule(lr){2-8} 
  
& Basel. $\mathcal{P}^{(0)}_{\text{user}}$ & 11296.6 & 37.15 & 0.300 & 9905.8 & 43.72 & 0.233 \\
& Basel. $\mathcal{P}^{(0)}_{\text{system}}$ & 6186.4 & 65.58 & 0.200 & 6356.9 & 63.88 & 0.190 \\
& Basel. $\mathcal{P}^{(1)}_{\text{system}}$ & 2181.6 & 87.86 & 0.120 & 4091.8 & 76.75 & 0.135 \\
& Basel. $\mathcal{P}^{(3)}_{\text{system}}$ & 2306.0 & 87.17 & 0.140 & 4164.7 & 76.34 & 0.165 \\
  
  \cmidrule(lr){2-8} 
  
& TRACES (0.1) & 17429.3 & 3.02 & 0.360 & 16643.2 & 5.44 & 0.325 \\
& TRACES (0.2) & 17040.3 & 5.19 & 0.350 & 16048.6 & 8.82 & 0.320 \\
& TRACES (0.3) & 16104.1 & 10.39 & 0.330 & 15189.0 & 13.71 & 0.310 \\
& TRACES (0.4) & 14082.4 & 21.65 & 0.240 & 13297.9 & 24.45 & 0.300 \\
& TRACES (0.5) & 11774.8 & 34.49 & 0.120 & 11269.0 & 35.97 & 0.280 \\
& TRACES (0.6) & 8847.1 & 50.77 & 0.070 & 8885.7 & 49.52 & 0.250 \\

\midrule

\multirow{12}{*}{GPT-OSS-120B}
  & Standard & 5781.4 & - & 0.420 & 6697.3 & - & 0.365 \\
  
  \cmidrule(lr){2-8} 
  
& Basel. $\mathcal{P}^{(0)}_{\text{user}}$ & 4075.4 & 29.50 & 0.450 & 3521.6 & 47.41 & 0.370 \\
& Basel. $\mathcal{P}^{(0)}_{\text{system}}$ & 4843.1 & 16.22 & 0.390 & 5022.2 & 25.01 & 0.365 \\
& Basel. $\mathcal{P}^{(1)}_{\text{system}}$ & 4481.7 & 22.48 & 0.310 & 4627.9 & 30.89 & 0.365 \\
& Basel. $\mathcal{P}^{(3)}_{\text{system}}$ & 4564.8 & 21.04 & 0.310 & 4873.6 & 27.23 & 0.363 \\
  
  \cmidrule(lr){2-8} 
  
& TRACES (0.4) & 5121.9 & 11.40 & 0.380 & 5606.1 & 16.29 & 0.338 \\
& TRACES (0.5) & 4623.5 & 20.03 & 0.340 & 5117.7 & 23.59 & 0.333 \\
& TRACES (0.6) & 3572.6 & 38.20 & 0.280 & 4216.9 & 37.03 & 0.310 \\
& TRACES (0.7) & 2806.8 & 51.45 & 0.220 & 3193.3 & 52.32 & 0.270 \\
& TRACES (0.8) & 2179.6 & 62.29 & 0.170 & 2265.5 & 66.17 & 0.223 \\
& TRACES (0.9) & 1611.0 & 72.13 & 0.130 & 1574.4 & 76.49 & 0.168 \\

\bottomrule
\end{tabular}
\caption{Performance of TRACES - BeyondAIME and IMO AnswerBench - seed 42}
\label{tab:TRACES-other_datasets}
\end{table}

\newpage

\subsection{Additional models} \label{sec:appendix-additional-models}

\begin{table}[h]
\centering
\tiny
\begin{tabular}{ll
                ccc
                ccc}
\toprule
\multirow{2}{*}{\textbf{Model}}  &
\multirow{2}{*}{\textbf{Config.}} 
& \multicolumn{3}{c}{\textbf{MATH500}} & \multicolumn{3}{c}{\textbf{GPQA-Diamond}} \\
\cmidrule(lr){3-5} \cmidrule(lr){6-8} 
& & \# Tokens & Saved (\%) & Pass@1 & \# Tokens & Saved (\%) & Pass@1 \\
\midrule

\multirow{12}{*}{Qwen-3-8B}
  & Standard & 3933.3 & - & 0.882 & 9821.6 & - & 0.545 \\
  
  \cmidrule(lr){2-8} 
  
& Basel. $\mathcal{P}^{(0)}_{\text{user}}$ & 2300.2 & 41.52 & 0.890 & 3543.8 & 63.92 & 0.510 \\
& Basel. $\mathcal{P}^{(0)}_{\text{system}}$ & 1247.2 & 68.29 & 0.822 & 5366.4 & 45.36 & 0.368 \\
& Basel. $\mathcal{P}^{(1)}_{\text{system}}$ & 1306.9 & 66.79 & 0.762 & 4845.5 & 50.67 & 0.424 \\
& Basel. $\mathcal{P}^{(3)}_{\text{system}}$ & 1365.9 & 65.29 & 0.768 & 5332.5 & 45.71 & 0.379 \\
  
  \cmidrule(lr){2-8} 
  
& TRACES (0.4) & 3208.9 & 18.43 & 0.894 & 9043.4 & 7.93 & 0.596 \\
& TRACES (0.5) & 2866.9 & 27.13 & 0.880 & 8578.7 & 12.66 & 0.581 \\
& TRACES (0.6) & 2302.5 & 41.47 & 0.858 & 7128.7 & 27.42 & 0.540 \\
& TRACES (0.7) & 1756.4 & 55.36 & 0.802 & 6084.7 & 38.05 & 0.510 \\
& TRACES (0.8) & 1376.5 & 65.01 & 0.752 & 5204.7 & 47.01 & 0.469 \\
& TRACES (0.9) & 1058.3 & 73.09 & 0.670 & 3700.7 & 62.32 & 0.399 \\

\midrule

\multirow{12}{*}{Qwen3-30B-A3B-Thinking-2507}
  & Standard & 5666.1 & - & 0.993 & 7435.3 & - & 0.707 \\
  
  \cmidrule(lr){2-8} 
  
& Basel. $\mathcal{P}^{(0)}_{\text{user}}$ & 3557.6 & 37.21 & 0.998 & 5710.7 & 23.19 & 0.707 \\
& Basel. $\mathcal{P}^{(0)}_{\text{system}}$ & 4026.4 & 28.94 & 0.987 & 5277.8 & 29.02 & 0.348 \\
& Basel. $\mathcal{P}^{(1)}_{\text{system}}$ & 3652.7 & 35.53 & 0.981 & 5462.7 & 26.53 & 0.232 \\
& Basel. $\mathcal{P}^{(3)}_{\text{system}}$ & 3564.1 & 37.09 & 0.974 & 5266.2 & 29.17 & 0.278 \\
  
  \cmidrule(lr){2-8} 
  
& TRACES (0.4) & 3003.9 & 46.98 & 0.854 & 6645.1 & 10.62 & 0.707 \\
& TRACES (0.5) & 2496.6 & 55.93 & 0.828 & 6199.9 & 16.61 & 0.687 \\
& TRACES (0.6) & 1864.5 & 67.09 & 0.776 & 5293.4 & 28.81 & 0.641 \\
& TRACES (0.7) & 1456.3 & 74.29 & 0.716 & 4183.8 & 43.73 & 0.596 \\
& TRACES (0.8) & 1303.4 & 76.99 & 0.700 & 3472.2 & 53.30 & 0.555 \\
& TRACES (0.9) & 1184.5 & 79.09 & 0.684 & 2250.9 & 69.73 & 0.489 \\

\midrule

\multirow{12}{*}{GPT-OSS-120B}
  & Standard & 1147.9 & - & 0.940 & 2249.4 & - & 0.561 \\
  
  \cmidrule(lr){2-8} 
  
& Basel. $\mathcal{P}^{(0)}_{\text{user}}$ & 697.3 & 39.25 & 0.958 & 1329.3 & 40.90 & 0.535 \\
& Basel. $\mathcal{P}^{(0)}_{\text{system}}$ & 977.2 & 14.87 & 0.914 & 1577.7 & 29.86 & 0.556 \\
& Basel. $\mathcal{P}^{(1)}_{\text{system}}$ & 884.6 & 22.94 & 0.896 & 1410.7 & 37.29 & 0.540 \\
& Basel. $\mathcal{P}^{(3)}_{\text{system}}$ & 909.6 & 20.76 & 0.904 & 1534.6 & 31.78 & 0.530 \\
  
  \cmidrule(lr){2-8} 
  
& TRACES (0.4) & 1040.5 & 9.36 & 0.940 & 2014.5 & 10.44 & 0.606 \\
& TRACES (0.5) & 985.8 & 14.12 & 0.930 & 1889.3 & 16.01 & 0.591 \\
& TRACES (0.6) & 882.6 & 23.11 & 0.906 & 1535.3 & 31.75 & 0.556 \\
& TRACES (0.7) & 769.3 & 32.98 & 0.876 & 1362.1 & 39.45 & 0.535 \\
& TRACES (0.8) & 708.8 & 38.25 & 0.852 & 1282.2 & 42.99 & 0.520 \\
& TRACES (0.9) & 638.5 & 44.38 & 0.812 & 1180.9 & 47.50 & 0.495 \\

\bottomrule
\end{tabular}
\caption{Performance of TRACES - Qwen-3-8B and DS-Qwen3-8B - seed 42}
\label{tab:TRACES-other_models}
\end{table}

\end{document}